\documentclass[letterpaper]{article} 
\usepackage[preprint]{aaai2027}  
\usepackage[hyphens]{url}  
\usepackage{graphicx} 
\usepackage{natbib}  
\usepackage{caption} 
\usepackage{algorithm}
\usepackage{algorithmic}
\usepackage{amsfonts}       
\usepackage{amsmath}
\usepackage{newfloat}
\usepackage{listings}
\DeclareCaptionStyle{ruled}{labelfont=normalfont,labelsep=colon,strut=off} 
\floatstyle{ruled}
\newfloat{listing}{tb}{lst}{}
\floatname{listing}{Listing}

\usepackage{booktabs}
\usepackage{multirow}
\title{Observation-Grounded Self-Predictive Reinforcement Learning for \\ Visual Continuous Control}
\author {
    Xinwei Liu\textsuperscript{\rm 1},
    Junyuan Liang\textsuperscript{\rm 1}\corresponding,
    Jianting Zhang\textsuperscript{\rm 2},
    Wuhui Chen\textsuperscript{\rm 1}
}
\affiliations {
    \textsuperscript{\rm 1}Sun Yat-sen University\\
    \textsuperscript{\rm 2}Purdue University\\
    liuxw73@mail2.sysu.edu.cn,
    liangjy53@mail2.sysu.edu.cn, 
    zhan4674@purdue.edu, chenwuh@mail.sysu.edu.cn
}

\begin{document}

\maketitle

\begin{abstract}
Sample-efficient policy learning from pixels is a long-standing challenge in reinforcement learning (RL). 
Recent dynamics-based representation learning methods have significantly improved the sample efficiency of model-free visual RL by learning dynamics-aware representations through auxiliary prediction performed either in latent space (self-prediction) or observation space (observation prediction).
However, state-of-the-art methods from both categories still struggle on challenging visual control tasks when training data is limited. 
We posit that relying on either predictive objective alone may be insufficient.  
Self-prediction encourages latent representations to be temporally predictive over multiple future steps, but does not explicitly require latent transitions to align with observation-level dynamics. 
In contrast, observation prediction grounds learned representations in observation-level dynamics, but does not directly regularize the temporal predictability of latent representations over extended horizons.
In this paper, we propose Observation-Grounded Self-Predictive Representations (OG-SPR), a model-free visual RL algorithm for continuous control that learns representations that are both temporally predictive in latent space and grounded in observation-level dynamics. Building on an actor-critic framework, OG-SPR incorporates two core auxiliary objectives: multi-step latent self-prediction and next-observation prediction. In addition, OG-SPR uses short-term value prediction as an auxiliary task to further stabilize value learning.
We empirically show that directly imposing latent self-prediction on the shared representation may over-constrain it and does not necessarily improve performance. 
To address this issue, OG-SPR introduces two lightweight adapters for latent self-prediction, allowing the shared representation to benefit from temporally predictive signals without being forced to directly satisfy the self-prediction objective.
Experiments on 28 visual control tasks from the DeepMind Control Suite show that OG-SPR improves aggregate performance over state-of-the-art self-predictive and observation-predictive RL methods, with particularly pronounced gains in challenging domains such as dog and humanoid.
\end{abstract}
\section{Introduction}
\begin{figure}[t]
\centering
\includegraphics[width=0.97\columnwidth]{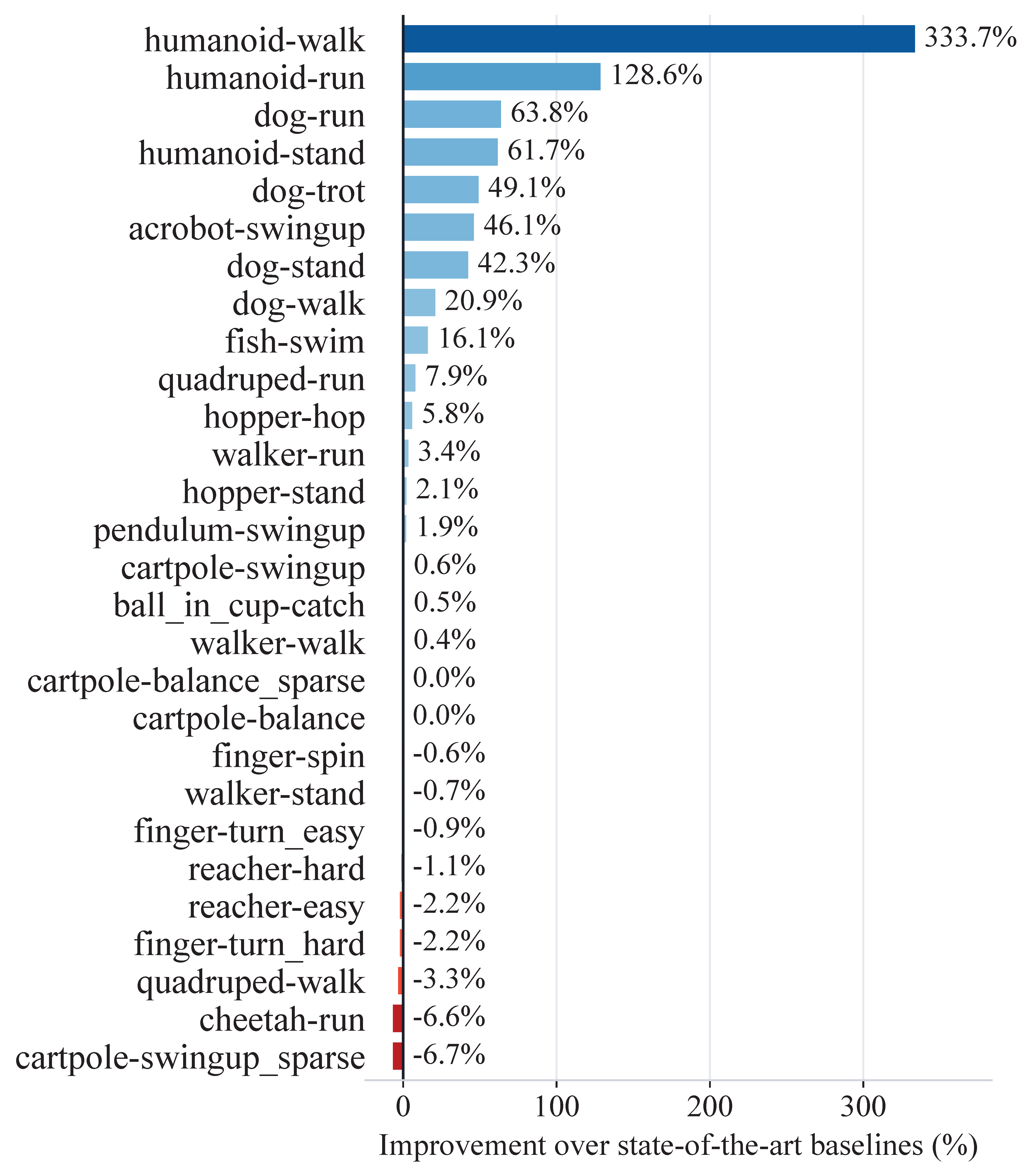} 
\caption{Performance improvements of OG-SPR over state-of-the-art baselines on the DeepMind Control Suite at 500k environment steps. 
For each task, the baseline score is defined as the stronger result between the recent self-predictive method MR.Q~\cite{DBLP:conf/iclr/FujimotoD0TR25} and observation-predictive method AnonMethod~\cite{anonymous2026preliminary}.}
\label{improvement}
\end{figure}
Learning policies from pixel inputs is an important problem in reinforcement learning (RL). While prior work has shown the potential of mastering visual continuous control by model-free methods, such as DrQ \cite{DBLP:conf/iclr/YaratsKF21} and DrQ-v2 \cite{DBLP:conf/iclr/YaratsFLP22}, sample efficiency remains a major challenge.
Recently, dynamics-based representation learning has made significant progress in data-efficient visual RL \cite{DBLP:conf/iclr/SchwarzerAGHCB21, DBLP:conf/icml/SchwarzerOCBAC23, DBLP:conf/iclr/FujimotoD0TR25, DBLP:journals/corr/abs-2506-05418, anonymous2026preliminary}. These methods learn dynamics-aware representations for model-free RL algorithms, without using a learned dynamics model for planning or value estimation, as done in model-based RL methods \cite{DBLP:conf/icml/HansenSW22, DBLP:conf/iclr/00010024, hafner2025dreamerv3}. Two representative lines of work are self-predictive methods and observation-predictive methods. 
Self-predictive approaches \cite{DBLP:conf/nips/ZhengWSMZXDH23, DBLP:journals/corr/abs-2406-02696, DBLP:conf/iclr/FujimotoD0TR25}, inspired by self-supervised learning \cite{DBLP:conf/nips/GrillSATRBDPGAP20}, learn representations by predicting latent embeddings of future observations over multiple time steps. 
In contrast, observation-predictive methods \cite{DBLP:conf/icml/GeladaKBNB19, anonymous2026preliminary} train representations by predicting next observations.

Both lines of work have recently achieved performance matching or surpassing state-of-the-art model-based methods, such as DreamerV3 \cite{hafner2025dreamerv3} and TD-MPC2 \cite{DBLP:conf/iclr/00010024}, on diverse tasks \cite{DBLP:conf/iclr/FujimotoD0TR25}. 
This suggests that learning meaningful representations is a key enabler for data-efficient visual RL. 
Despite this progress, the success of these methods remains uneven across tasks. 
Strong recent methods from both lines still struggle in challenging visual control domains under limited data budgets, such as 500k environment steps \cite{DBLP:conf/iclr/FujimotoD0TR25, anonymous2026preliminary}.
We posit that relying on either latent self-prediction or next-observation prediction alone may be insufficient for data-efficient visual continuous control. 
Self-prediction encourages latent representations to be temporally predictive over multiple future steps, but does not explicitly require latent transitions to align with observation-level dynamics. In contrast, observation prediction grounds learned representations in observation-level dynamics through future observation reconstruction, but does not directly enforce long-horizon temporal predictability in latent space.
These two objectives therefore impose complementary inductive biases, motivating the following question:

\emph{Can a model-free visual RL agent become more data-efficient in continuous control by learning representations that are both temporally predictive in latent space and grounded in observation-level dynamics?}

In this paper, we provide a positive answer to this question by exploring how to leverage the complementary strengths of latent self-prediction and observation prediction.
Specifically, we present Observation-Grounded Self-Predictive Representations (OG-SPR), a model-free RL algorithm that learns representations satisfying both objectives.
Building on an off-policy actor-critic framework \cite{sutton1998reinforcement}, OG-SPR learns state-action representations through two dynamics-oriented auxiliary tasks: multi-step latent self-prediction and next-observation prediction. 
In addition to these dynamics-oriented objectives, OG-SPR also includes short-term value prediction as an auxiliary objective to stabilize value learning \cite{anonymous2026preliminary}.
However, as shown in our ablation study, naively combining these auxiliary objectives does not effectively improve performance. 
We hypothesize that directly imposing the latent self-prediction objective on the shared representation may over-constrain it.
To mitigate this issue, OG-SPR introduces two lightweight adapters for latent self-prediction, with one placed after the observation encoder and the other after the state-action encoder.
These adapters form an adapter-mediated self-prediction branch, so that the shared representation can receive temporally predictive learning signals without being forced to directly satisfy the self-prediction objective.

We evaluate OG-SPR on 28 visual control tasks from the DeepMind Control Suite (DMControl) \cite{DBLP:journals/corr/abs-1801-00690} under limited data budgets. 
Experimental results show that OG-SPR achieves better average performance than state-of-the-art self-predictive and observation-predictive RL methods \cite{DBLP:conf/iclr/FujimotoD0TR25, anonymous2026preliminary}, with particularly strong gains in challenging domains such as \textit{dog} and \textit{humanoid}, as shown in Figure~\ref{improvement}.
OG-SPR represents an initial step toward data-efficient RL for complex visual continuous control tasks by integrating self-predictive and observation-predictive methods.
We hope this work can inspire further research on their intersection to improve the data efficiency of model-free visual RL.

\section{Related Work}
\subsection{Representation Learning for Data-Efficient RL}
Representation learning is widely used to improve the sample efficiency of RL. Prior work has explored a wide range of representation learning approaches for RL \cite{DBLP:journals/tmlr/EchchahedC25}. Metric-based methods \cite{DBLP:conf/aaai/LiaoZ023, DBLP:journals/corr/abs-2507-18519} shape the embedding space by enforcing task-relevant similarities among state representations. 
Data-augmentation methods \cite{DBLP:conf/iclr/YaratsFLP22, DBLP:conf/iclr/MaL0L0000T24} encourage the learned representations to be invariant to irrelevant visual changes. Contrastive learning methods \cite{DBLP:conf/nips/EysenbachZLS22, DBLP:conf/iclr/LiuTE25} learn representations by contrasting positive pairs against negative pairs. 
Our work focuses on dynamics-based representation learning, which learns dynamics-aware representations for model-free RL through dynamics prediction \cite{DBLP:conf/icml/GeladaKBNB19, DBLP:conf/iclr/FujimotoD0TR25}.

\subsection{Dynamics-Based Representation Learning}
Recent dynamics-based methods have achieved performance competitive with or better than model-based methods, such as DreamerV3 \cite{hafner2025dreamerv3} and TD-MPC2 \cite{DBLP:conf/iclr/00010024}, on diverse tasks with less algorithmic and computational complexity \cite{DBLP:conf/iclr/FujimotoD0TR25, anonymous2026preliminary}. 
This branch of research can be broadly divided into self-predictive and observation-predictive methods. Self-predictive methods learn representations by predicting latent embeddings of future observations. 
TD7 \cite{DBLP:conf/nips/FujimotoCSGPM23} learns state-action representations in the latent dynamics space for low-dimensional continuous control tasks. Building on TD7, MR.Q \cite{DBLP:conf/iclr/FujimotoD0TR25} is a state-of-the-art self-predictive method that augments latent self-prediction with additional predictive objectives, such as reward and termination prediction. Observation-predictive methods also train representations through dynamics modeling, but ground the prediction objective in observation space. 
OFENet \cite{DBLP:conf/icml/OtaOJMN20} is a representative observation-predictive approach for low-dimensional settings. 
AnonMethod \cite{anonymous2026preliminary} is a recent observation-predictive method. It addresses a bottleneck of observation-predictive representation learning in low-dimensional settings, where reconstruction losses tend to be dominated by observation dimensions with large value ranges. To mitigate this issue, AnonMethod performs both representation learning and RL in a normalized observation space, making observation-predictive learning applicable across diverse domains.
However, existing dynamics-based methods typically focus on prediction either in latent space or in observation space, and these two directions have largely evolved as separate lines of work. 
The potential of training representations simultaneously in latent and observation spaces for model-free RL remains underexplored. OG-SPR takes an initial step toward addressing this gap.

\subsection{Model-Based Reinforcement Learning}
Model-based methods, such as the Dreamer series \cite{DBLP:conf/iclr/HafnerLB020, DBLP:conf/iclr/HafnerL0B21, hafner2025dreamerv3}, also learn latent dynamics with observation prediction. Our work differs from these methods in both learning objectives and how the learned dynamics are used. 
In terms of learning objectives, OG-SPR overlaps with these methods only in observation prediction, while the remaining objectives are distinct. 
In terms of usage, OG-SPR does not maintain a learned dynamics model for RL training. Instead, it learns representations through dynamics modeling in both latent and observation spaces for model-free RL.

\section{Preliminaries}
\subsection{Reinforcement Learning}
Reinforcement learning (RL) addresses the problem of sequential decision making, usually formulated as a Markov Decision Process (MDP). 
An MDP can be represented by a tuple $(\mathcal{S}, \mathcal{A}, P, r, \gamma)$, 
where $\mathcal{S}$ and $\mathcal{A}$ denote the state and action spaces, respectively; $P(s_{t+1}|s_t,a_t)$ denotes the transition probability of the next state \begin{math} s_{t+1} \end{math} given the current state \begin{math} s_t \end{math} and action $a_t$; $r:\mathcal{S} \times \mathcal{A} \rightarrow \mathbb{R}$ is the reward function; $\gamma \in [0, 1)$ is the discount factor. The objective of RL is to learn a policy $\pi:\mathcal{S} \rightarrow \mathcal{A}$ that maximizes the discounted cumulative reward $\sum_{t=0}^{\infty} \gamma^{t} r_{t}$. Actor-critic methods \cite{sutton1998reinforcement} typically learn an action-value function $Q^{\pi}(s,a)= \mathbb{E}_{\pi}[\sum_{t=0}^{\infty} \gamma^{t} r_{t}|s_0=s,a_0=a]$ which models the expected return, from an initial state \begin{math}s\end{math} and action \begin{math}a\end{math}. 

In low-dimensional settings, the observation an agent receives at time step \begin{math}t\end{math}, denoted as \begin{math}o_t\end{math}, is usually treated as the state \begin{math}s_t\end{math} (\begin{math}s_t := o_t \end{math}). In complex settings (e.g., visual RL), \begin{math}o_t\end{math} does not necessarily satisfy the Markov property and therefore may not be directly treated as \begin{math}s_t\end{math}. This induces a partially observable MDP (POMDP). Following common practice \cite{DBLP:journals/corr/MnihKSGAWR13, DBLP:conf/iclr/YaratsFLP22}, we approximate the Markov state by stacking three consecutive prior observations into a state \begin{math}s_t := [o_{t-2}, o_{t-1}, o_t] \end{math}.

\section{Methodology}
\begin{figure*}[t]
\centering
\includegraphics[width=0.80\textwidth]{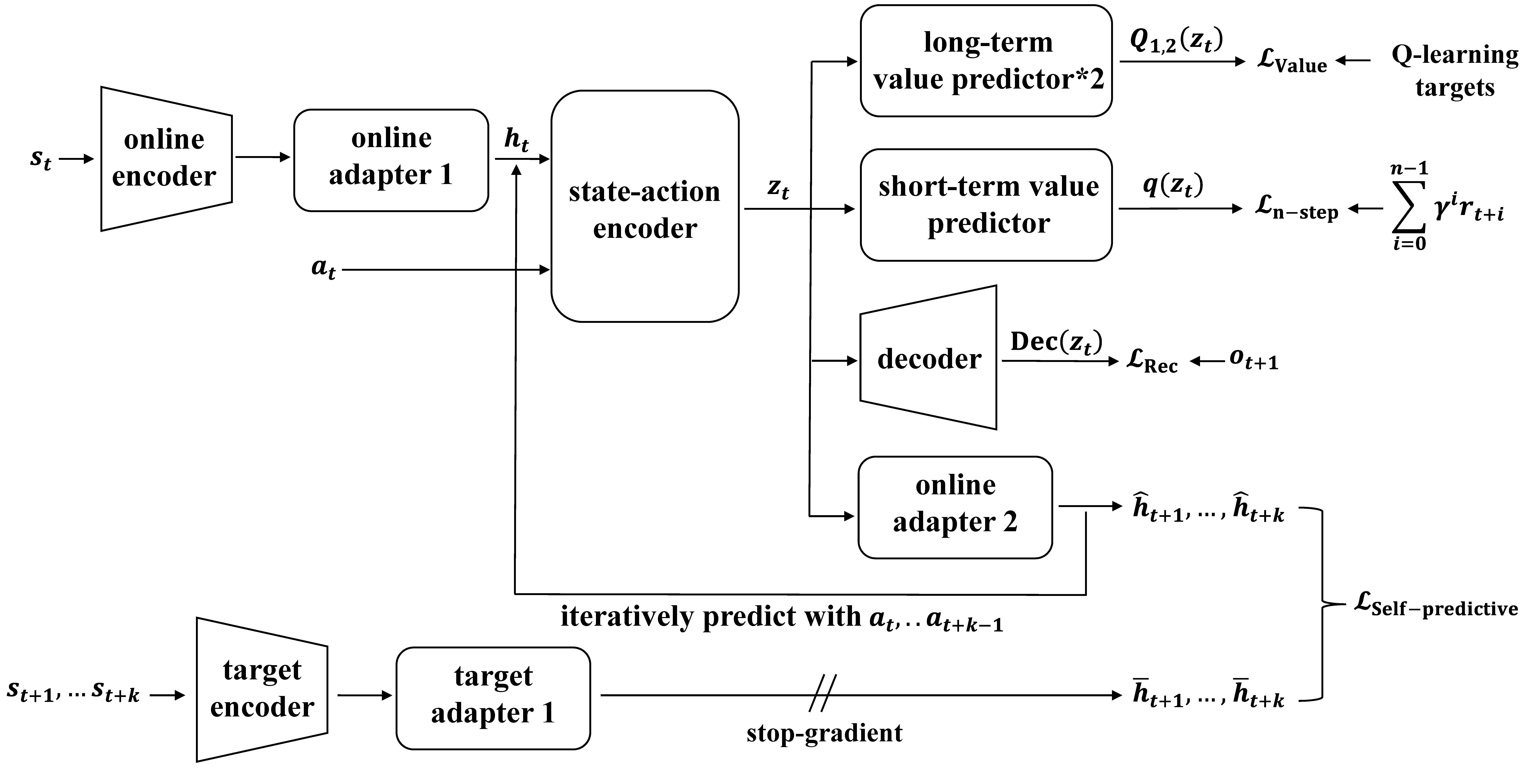} \\
(a) Value network with two long-term value prediction heads and three auxiliary prediction heads. \\
\includegraphics[width=0.48\textwidth]{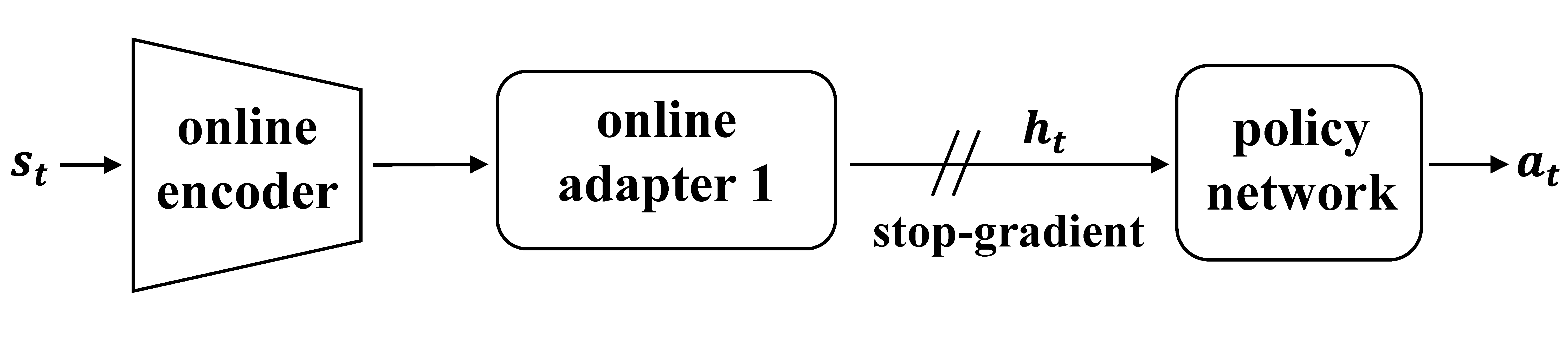} \\
(b) Policy network with detached inputs. \\
\caption{An illustration of the full OG-SPR method.}
\label{framework}
\end{figure*}

As shown in Figure~\ref{framework}, OG-SPR is built on an actor-critic framework with value and policy networks. 
An image-based state $s$ is first passed through an online encoder $f$ followed by Adapter 1 $u_1$, producing a low-dimensional latent state representation $h$. The latent state is then combined with the action $a$ and fed into a state-action encoder $g$ to obtain a state-action representation $z$:
\begin{equation}
\label{eq:1}
\begin{aligned}
h = u_1\left(f(s)\right), \quad\quad z = g(h, a).
\end{aligned}
\end{equation}
Adapter 1 maps encoder features into the latent state space on which policy learning, value learning, and latent self-prediction are built.
Similar to AnonMethod, the policy network takes $h$ with gradients stopped, while the state-action representation $z$ is used for long-term value learning and three auxiliary tasks.

\subsection{Value Learning}
The value target is constructed based on TD3 \cite{DBLP:conf/icml/FujimotoHM18}. 
Specifically, we train two separate long-term value predictors $Q_1$ and $Q_2$, with identical architectures. 
Both predictors take the same state-action representations as inputs. 
To stabilize bootstrapping, we maintain target copies of the networks used for target value estimation.
Target networks and their outputs are denoted by adding a prime to their online counterparts (e.g., $f'$ and $h'$).
Target parameters are not updated via gradient descent, but periodically synchronized with their online counterparts \cite{DBLP:conf/iclr/FujimotoD0TR25}. The smoothed target action $a^{\pi'}$ is produced by the target policy network $\pi'$ with clipped Gaussian noise:
\begin{equation}
\label{eq:2}
\begin{gathered}
a^{\pi'}= \operatorname{clip}(a',-1, 1), \\
a'=\pi'(h') + \operatorname{clip}(\epsilon, -c, c), \quad\epsilon \sim \mathcal{N}(0, \sigma^2),
\end{gathered}
\end{equation}
where $h' = u_1'\left(f'(s)\right)$. 
We use $n$-step returns for value learning \cite{DBLP:conf/iclr/YaratsFLP22}. Given a sampled transition sequence $\tau = (s_t, a_t, r_{t:t+n-1}, s_{t+n})$, the value target $y$ is:
\begin{equation}
\label{eq:3}
\begin{gathered}
y = \sum_{i=0}^{n-1} \gamma^i r_{t+i} + \gamma^n  \min_{j \in \{1,2\}}{Q'_{j}(z'_{t+n})}, \\  z'_{t+n} = g'\left( h'_{t+n},a_{t+n}^{\pi'} \right),
\end{gathered}
\end{equation}
where $h'_{t+n} = u_1'(f'(s_{t+n}))$, and $a_{t+n}^{\pi'}$ is computed based on $s_{t+n}$ via Equation~\ref{eq:2}. 
We use the Huber loss instead of mean squared error (MSE) for value learning to eliminate bias from prioritized sampling \cite{DBLP:conf/nips/FujimotoMP20, DBLP:conf/iclr/FujimotoD0TR25}:
\begin{equation}
\label{eq:4}
\begin{aligned}
\mathcal{L}_\text{Value} = \operatorname{Huber}\left(Q_{1}(z_t), y\right) + \operatorname{Huber}\left(Q_{2}(z_t), y\right).
\end{aligned}
\end{equation}

\subsection{Policy Learning}
The policy network $\pi$ is optimized using the deterministic policy gradient \cite{DBLP:conf/icml/SilverLHDWR14}. 
An L2 regularization term is added to the pre-activation policy outputs $\tilde{a}^{\pi}$, which helps avoid local minima when rewards and value estimates are sparse \cite{DBLP:journals/nature/SchrittwieserAH20, DBLP:conf/iclr/FujimotoD0TR25}:
\begin{equation}
\label{eq:5}
\begin{gathered}
a^{{\pi}} = \tanh(\tilde{a}^{\pi}), \\
\mathcal{L}_\text{Policy} = - \frac{1}{2} \sum_{i \in\{1,2\}}{Q_{i}(z)} + \lambda_\text{pre-activ} \|\tilde{a}^{\pi}\|_2^2, 
\end{gathered}
\end{equation}
where $z = g\left(u_1(f(s)) , a^\pi\right)$. $a^\pi$ is computed from $s$ using the online encoder, Adapter 1, and policy network without policy smoothing noise.

\subsection{Auxiliary Tasks} 
\paragraph{Latent Self-Prediction.} Starting from the current latent state $h_t$, 
future latent states are predicted recursively by applying the state-action encoder $g$ and 
Adapter 2 $u_2$ with the corresponding actions from the sampled transition sequence:
\begin{equation}
\label{eq:8}
\begin{aligned}
\hat{h}_{t+1}&= u_2(g({h}_{t}, a_{t})), \\
\hat{h}_{t+i+1}&= u_2(g(\hat{h}_{t+i}, a_{t+i})), \quad i=1,\ldots,K-1. 
\end{aligned}
\end{equation}
Adapter 2 maps state-action representations back to the latent space defined by Adapter 1. Both adapters use the same output activation function to maintain consistent output spaces. By inserting this adapter between the state-action representation and the latent prediction target, this design prevents the self-prediction objective from being directly imposed on the shared state-action representation, thereby alleviating the constraint on it. The targets for latent self-prediction are computed using target networks \cite{DBLP:conf/iclr/SchwarzerAGHCB21, DBLP:conf/iclr/FujimotoD0TR25}. Given the target latent embeddings $({h'}_{t+1},\ldots, {h'}_{t+K})$ produced by the target encoder $f'$ and target Adapter 1 $u_1'$, the self-prediction loss is computed using the mean squared error (MSE):
\begin{equation}
\label{eq:9}
\mathcal{L}_\text{Self-predictive}= \sum_{k=1}^{K} \operatorname{MSE}\left(\hat{h}_{t+k}, {h'}_{t+k}\right).
\end{equation}

\paragraph{Next-Observation Prediction.} The second auxiliary task predicts the next observation from the state-action representation. 
Specifically, a decoder $\operatorname{Dec}$ takes $z_t$ as input and reconstructs 
the next observation $o_{t+1}$ using the MSE loss:
\begin{equation}
\label{eq:7}
\begin{aligned}
\mathcal{L}_\text{Rec} = 
\left\|\operatorname{Dec}(z_t)-o_{t+1}\right\|_2^2.
\end{aligned}
\end{equation}

\paragraph{Short-Term Value Prediction.} We adopt short-term value prediction as an auxiliary stabilizing objective, which provides lower-variance training signals than bootstrapped value targets and stabilizes representation learning when value estimates are noisy \cite{anonymous2026preliminary}. A predictor $q$ takes state-action representations $z$ as input and predicts the cumulative discounted reward over the next $n$ steps. The auxiliary loss is computed as the cross-entropy (CE) loss between the predicted logits and a two-hot encoding of the target:  
\begin{equation}
\label{eq:6}
\begin{gathered}
 R_t=\sum_{j=0}^{n-1} \gamma^j r_{t+j}, \\
\mathcal{L}_\text{n-step}
= \operatorname{CE}\left(q(z_t), \operatorname{TwoHot}(R_t)\right),
\end{gathered}
\end{equation}
where $R_t$ is the non-bootstrapped term in the value target in Equation~\ref{eq:3}. We use a non-uniform return support for the two-hot encoding, with bin locations generated by the $\operatorname{symexp}$ transform, $\operatorname{symexp}(x)=\operatorname{sign}(x)(\exp(|x|)-1)$ \cite{hafner2025dreamerv3}.

During training, gradients from the policy loss are not propagated to the online encoder and Adapter 1, whereas all components of the value network are jointly optimized using the combined loss $\mathcal{L}$:
\begin{equation}
\label{eq:10}
\begin{aligned}
\mathcal{L} = \mathcal{L}_\text{Value} + \lambda_\text{Rec}\mathcal{L}_\text{Rec} + \lambda_\text{n-step}\mathcal{L}_\text{n-step} \\ +  \lambda_\text{Self-predictive}\mathcal{L}_\text{Self-predictive},
\end{aligned}
\end{equation}
where $\lambda_\text{Rec}$, $\lambda_\text{n-step}$ and $\lambda_\text{Self-predictive}$ denote the auxiliary loss weights for next-observation prediction, short-term value prediction, and latent self-prediction, respectively. The pseudocode of OG-SPR is provided in the appendix.

\section{Experiments}
In this section, we first compare OG-SPR with baselines in terms of performance and computational efficiency on continuous control tasks. We then conduct ablation studies to analyze the key components of OG-SPR, followed by a representation analysis to examine how self-prediction and observation prediction affect the learned state and state-action representations. Finally, we provide an additional evaluation on discrete-action domains to examine the behavior of OG-SPR beyond continuous-control settings.

\subsection{Experimental Setup}
\paragraph{Environments.} We primarily evaluate OG-SPR on the DMControl benchmark \cite{DBLP:journals/corr/abs-1801-00690}, a collection of continuous control tasks built on the MuJoCo simulator \cite{DBLP:conf/iros/TodorovET12}. The maximum episode score for each task is 1000. We consider 28 tasks used in prior work \cite{DBLP:conf/iclr/FujimotoD0TR25,anonymous2026preliminary}. Agents are trained for 500k environment steps, equivalent to 1M frames in the original environment under an action repeat of 2. The input state is constructed by stacking three previous frames, which are resized to 84 $\times$ 84 pixels in RGB format. 

\paragraph{Baselines.} We compare OG-SPR with representative data-efficient RL methods, including both model-based and model-free approaches: (1) TD-MPC2 \cite{DBLP:conf/iclr/00010024}, a strong model-based method for continuous control tasks, (2) DreamerV3 \cite{hafner2025dreamerv3}, a general-purpose model-based RL algorithm that performs well across diverse domains, (3) MR.Q \cite{DBLP:conf/iclr/FujimotoD0TR25}, a state-of-the-art self-predictive method that achieves performance competitive with model-based RL approaches, DreamerV3 and TD-MPC2, (4) AnonMethod \cite{anonymous2026preliminary}, a recent observation-predictive method achieving performance competitive with or better than MR.Q, (5) DrQ-v2 \cite{DBLP:conf/iclr/YaratsFLP22}, a simple yet strong model-free RL baseline for visual continuous control tasks based on data augmentation.

\paragraph{Implementation Details.} We follow MR.Q in constructing the actor-critic backbone of OG-SPR. Each adapter is implemented as a linear layer. We apply random shift augmentation to pixel observations \cite{DBLP:conf/iclr/YaratsFLP22}, but use different augmentation strategies for the observation-predictive and self-predictive objectives. For next-observation prediction, we apply the same sampled spatial shift to both the input observation and the prediction target to preserve spatial alignment. For latent self-prediction, we apply independently sampled shifts to future target observations. The prediction horizon for latent self-prediction is set to 5, following prior self-predictive methods \cite{DBLP:conf/iclr/SchwarzerAGHCB21, DBLP:conf/iclr/FujimotoD0TR25}. The auxiliary loss weights are kept fixed across tasks. 
We adopt the same default settings as AnonMethod for the observation-predictive and short-term value objectives, setting $\lambda_\text{Rec}=0.1$ and $\lambda_\text{n-step}=1.0$.
To determine $\lambda_\text{Self-predictive}$, we start from the default weight used in SPR \cite{DBLP:conf/iclr/SchwarzerAGHCB21}, 2.0, although SPR uses a cosine-similarity loss while OG-SPR uses MSE. 
We then select $\lambda_\text{Self-predictive}$ using a small subset containing only \textit{quadruped-run} and \textit{dog-run}, searching over candidate values spaced by 1.0. The resulting value, 5.0, is used for all 28 tasks without per-task tuning. Additional details are provided in the appendix.

\paragraph{Evaluation Protocol.} All experiments are run for 5 seeds. For each seed, we evaluate the agent every 5k environment steps over 10 episodes and report the average episode return as the evaluation score. Results for all baselines are obtained by re-running their official implementations, except for DrQ-v2. For DrQ-v2, we use the results reported in MR.Q.

Most prior work reports results on DMControl using raw, unnormalized scores \cite{DBLP:conf/iclr/YaratsKF21, DBLP:conf/iclr/YaratsFLP22, DBLP:conf/nips/ZhengWSMZXDH23, DBLP:conf/iclr/00010024, hafner2025dreamerv3, DBLP:conf/iclr/FujimotoD0TR25}. However, aggregate statistics over raw scores can be dominated by relatively easy tasks with high returns, such as \textit{cartpole-balance} and \textit{reacher-easy}. This may obscure improvements on more challenging tasks, such as \textit{dog-run} and \textit{humanoid-walk}, where raw scores are typically much lower. 
To provide a complementary view of aggregate performance, we additionally report DrQv2-normalized scores, which measure relative improvement over DrQ-v2, a strong model-free baseline for visual continuous control. The metric is computed in a way similar to the human-normalized score \cite{DBLP:conf/icml/WangSHHLF16} commonly used for Atari games \cite{DBLP:journals/jair/BellemareNVB13}:
{\small 
\begin{equation*}
\label{eq:14}
\begin{aligned}
\operatorname{DrQv2-Normalized}(x) = \frac{x-\text{random score}}{{\text{DrQ-v2 score} - \text{random score}}}.
\end{aligned}
\end{equation*}}
This normalization reduces the influence of tasks where DrQ-v2 already performs well and highlights improvements on tasks where DrQ-v2 remains weak.
Benchmark-level aggregate performance is reported using the mean and interquartile mean (IQM) over task-level scores, where each task-level score is averaged across seeds.

\subsection{Main Results}
As shown in Table~\ref{main_result}, OG-SPR achieves the best aggregate performance on both raw scores and DrQv2-normalized scores. 
\begin{table*}[t]
    \small
    \setlength{\tabcolsep}{1mm}
    \centering
    \begin{tabular}{lcccccc}
        \toprule
        \textbf{Tasks} & \textbf{DrQ-v2} & \textbf{TD-MPC2} &
        \textbf{DreamerV3} & \textbf{MR.Q} & \textbf{AnonMethod} & \textbf{OG-SPR (ours)} \\
        \midrule
        \multicolumn{7}{l}{\textit{Raw Scores of Selected Challenging Tasks}} \\
        \midrule

        dog-run & 
        10 {{[9, 12]}} & 
        8 {{[5, 11]}} & 
        29 {{[23, 36]}} &   
        51 {{[45, 56]}} & 
        55 {{[46, 69]}} &
        \textbf{90} {{[64, 115]}} \\
        
        dog-stand & 
        43 {{[37, 49]}} & 
        161 {{[147, 175]}} & 
        118 {{[78, 157]}} &  
        251 {{[233, 269]}} & 
        239 {{[214, 263]}}  &
        \textbf{357} {{[267, 448]}} \\
        
        dog-trot & 
        14 {{[11, 18]}} & 
        16 {{[12, 18]}} & 
        44 {{[34, 54]}} &  
        70 {{[61, 80]}} & 
        61 {{[59, 63]}} &
        \textbf{105} {{[79, 128]}} \\

        dog-walk & 
        22 {{[18, 29]}} & 
        14 {{[12, 17]}} & 
        42 {{[31, 52]}} &
        90 {{[79, 101]}} & 
        81 {{[72, 91]}} &
        \textbf{108} {{[102, 118]}} \\

        humanoid-run & 
        1 {{[1, 1]}} & 
        1 {{[1, 1]}} & 
        1 {{[1, 2]}} &  
        1 {{[1, 2]}} & 
        1 {{[1, 1]}} &
        \textbf{3} {{[2, 4]}} \\

        humanoid-run (2M) & 
        2 {{[2, 3]}} & 
        1 {{[1, 1]}} & 
        4 {{[1, 9]}} &  
        27 {{[4, 50]}} & 
        1 {{[1, 1]}} &
        \textbf{100} {{[94, 106]}} \\
        
        humanoid-stand & 
        6 {{[6, 7]}} & 
        6 {{[6, 7]}} & 
        7 {{[3, 10]}} &  
        8 {{[7, 8]}} & 
        6 {{[5, 7]}}  &
        \textbf{12} {{[8, 20]}} \\

        humanoid-stand (2M) & 
        9 {{[8, 11]}} & 
        7 {{[6, 7]}} & 
        11 {{[7, 17]}} &  
        79 {{[8, 200]}} & 
        31 {{[8, 55]}}  &
        \textbf{339} {{[284, 401]}} \\
        
        humanoid-walk & 
        2 {{[2, 2]}} & 
        2 {{[1, 2]}} & 
        2 {{[1, 3]}} &  
        3 {{[2, 5]}} & 
        2 {{[2, 3]}} &
        \textbf{15} {{[7, 22]}} \\

        humanoid-walk (2M) & 
        19 {{[2, 52]}} &
        2 {{[2, 2]}} & 
        2 {{[1, 3]}} &  
        48 {{[3, 93]}} & 
        54 {{[3, 153]}} &
        \textbf{316} {{[299, 333]}} \\

        \midrule
        \multicolumn{7}{l}{\textit{Aggregate Raw Results of 28 Tasks}} \\
        \midrule
        Mean & 510 [497, 523] & 485 [467, 505] & 536 [507, 563] & 599 [591, 606] & 607 [602, 612]  & \textbf{626} [620, 632] \\
        IQM & 545 [519, 564] & 489 [453, 524] & 571 [516, 622] & 685 [670, 699] & 702 [692, 711] & \textbf{730} [718, 741]  \\
        \midrule
        \multicolumn{7}{l}{\textit{Aggregate DrQv2-Normalized Results of 28 Tasks}} \\
        \midrule
         Mean & 1.00 & 1.35 [1.27, 1.42] & 1.94 [1.64, 2.24] & 2.84 [2.72, 2.96] & 2.65 [2.51, 2.80]  & \textbf{5.05} [4.40, 5.71]  \\
        IQM & 1.00 & 0.97 [0.92, 1.01] & 1.21 [1.14, 1.32] & 1.29 [1.24, 1.36] & 1.25 [1.20, 1.31]  & \textbf{1.61} [1.50, 1.76] \\
        \bottomrule
    \end{tabular}
    \caption{
    Results on selected challenging DMControl tasks and aggregate results over all 28 tasks at default 500k environment steps. 
    \textbf{2M} denotes results evaluated at 2M environment steps. See the appendix for humanoid-domain results evaluated at 1M and 1.5M environment steps. 
    Brackets denote 95\% bootstrap confidence intervals. Bold numbers indicate the best performance.
    }
    \label{main_result}
\end{table*}
Notably, OG-SPR substantially outperforms both MR.Q and AnonMethod in terms of DrQv2-normalized scores. Its mean DrQ-v2-normalized score is approximately $1.8\times$ that of MR.Q and $1.9\times$ that of AnonMethod. 
Figure~\ref{improvement} shows the percentage improvement of OG-SPR over the stronger of MR.Q and AnonMethod on each task, computed using raw scores. OG-SPR brings particularly large relative gains in challenging visual control domains, achieving \textbf{20.9--63.8\%} and \textbf{61.7--333.7\%} improvements in the \textit{dog} and \textit{humanoid} domains, respectively. 
Under the default 500k-step budget, all methods obtain relatively low scores in the \textit{humanoid} domain, making it difficult to fully reveal performance differences in this challenging domain. 
We therefore additionally evaluate the humanoid tasks with a relaxed training budget of 2M environment steps in Table~\ref{main_result}. Under this setting, OG-SPR still achieves substantial gains over all baselines.
We also provide aggregate learning curves computed using raw scores and DrQv2-normalized scores in Figures~\ref{main_results}. Full per-task results are provided in the appendix.

\begin{figure}[t]
\centering
\includegraphics[width=0.49\columnwidth]{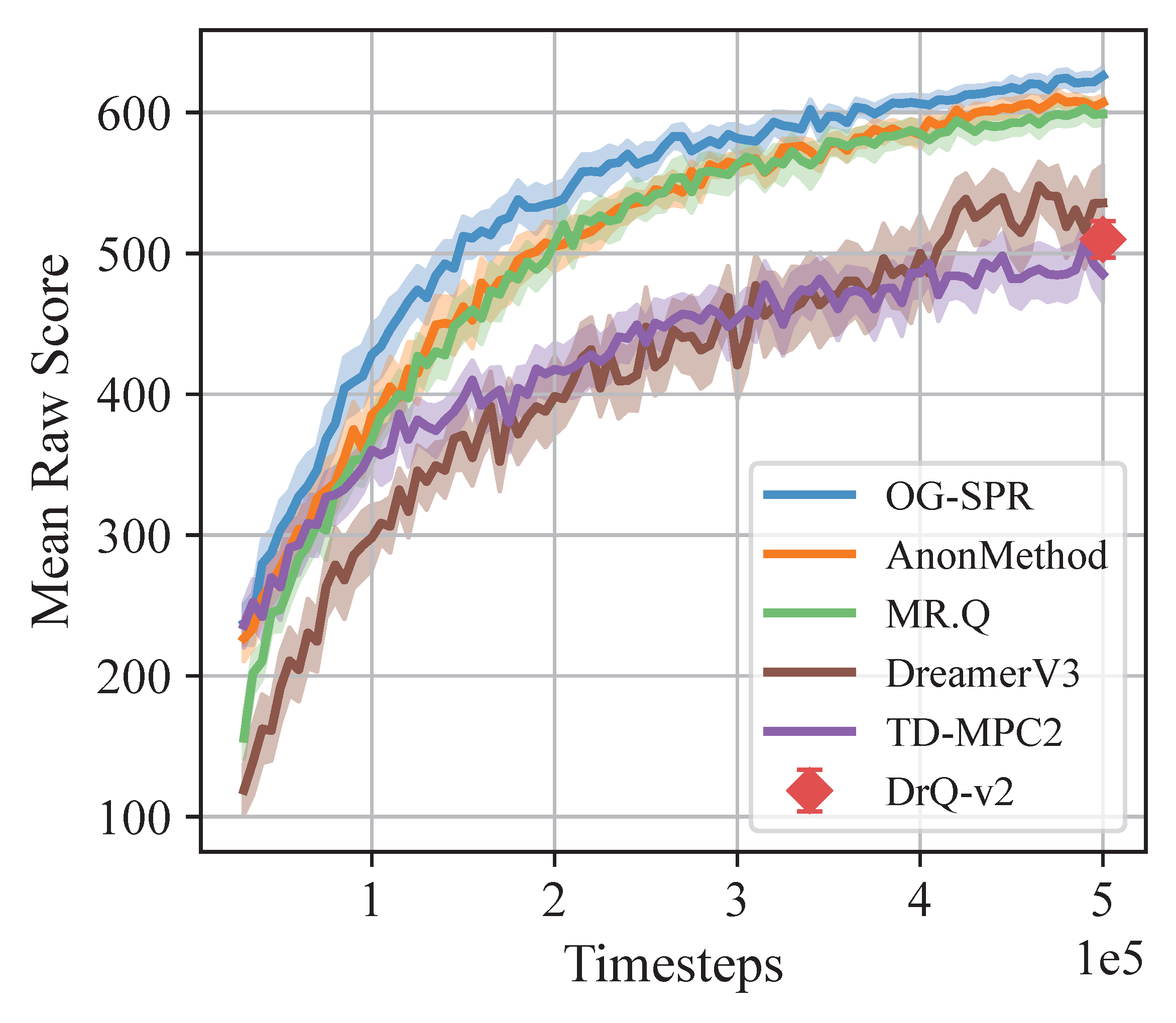} 
\includegraphics[width=0.49\columnwidth]{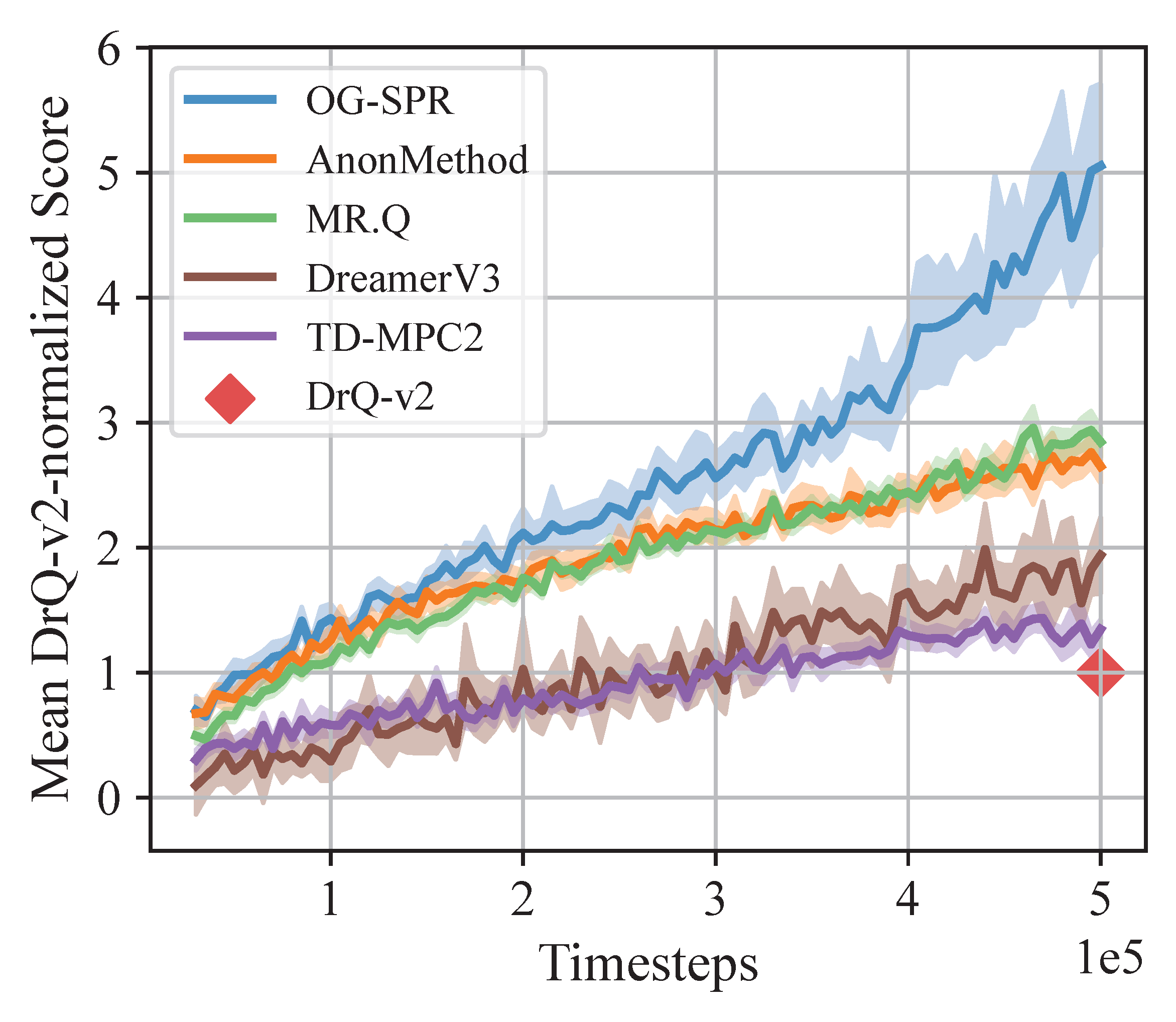} 
\caption{Aggregate learning curves on DMControl, computed as the mean over 28 tasks using raw scores and DrQ-v2-normalized scores. Shaded area captures a 95\% bootstrap confidence interval.}
\label{main_results}
\end{figure}

Since OG-SPR incorporates both observation-predictive and self-predictive objectives, it incurs higher training cost than AnonMethod and MR.Q.
OG-SPR requires 18.3$\%$ more training time than MR.Q, and trains substantially faster than DreamerV3 (see the appendix) when all methods are implemented in PyTorch \cite{DBLP:conf/nips/PaszkeGMLBCKLGA19}.

\subsection{Ablation Study}
We conduct ablation studies for two purposes: 
\begin{enumerate}
    \item To evaluate the effectiveness of learning representations that are both temporally predictive in latent space and grounded in observation-level dynamics.
    \item To examine the importance of the adapter design.
\end{enumerate}

The aggregate ablation results are summarized in Table~\ref{ab_results}. 
We denote by \textbf{OG-SPR - OP} the variant that removes the observation-predictive objective $\mathcal{L}_\text{Rec}$. \textbf{OG-SPR - SP} indicates the variant that removes the self-predictive objective $\mathcal{L}_\text{Self-predictive}$. \textbf{OG-SPR - SVP} denotes removing the short-term value prediction loss $\mathcal{L}_\text{n-step}$. 
Overall, removing any auxiliary objective degrades the performance of OG-SPR. The performance drop is especially pronounced when removing the self-predictive objective.

To study the adapter design, we compare OG-SPR with three variants. 
In \textbf{Shared Adapter}, the two adapters share parameters, reducing the flexibility of the mappings used for recursive self-prediction. 
In \textbf{No Adapter}, both adapters are removed, and the self-prediction objective is directly imposed on the shared representation. 
These two variants progressively reduce the flexibility of the adapter-mediated self-prediction branch.
Both \textbf{Shared Adapter} and \textbf{No Adapter} underperform complete OG-SPR, 
indicating that reducing or removing adapter flexibility hurts performance. 
We also consider \textbf{Adapter 2 Only}, where Adapter 1 is removed, which is the design used in MR.Q. 
Compared with the full OG-SPR, \textbf{Adapter 2 Only} performs worse because removing Adapter 1 makes the state encoder output space inconsistent with the output space of Adapter 2, as the former uses an ELU activation whereas the latter uses no activation function. 
These results support the importance of using two separate adapters with matched output spaces for latent self-prediction. 

\begin{table}[t]
    \setlength{\tabcolsep}{1mm}
    \centering
    \begin{tabular}{lcc c cc}
        \toprule
        \multirow{2}{*}{\textbf{Methods}}  & \multicolumn{2}{c}{\textbf{Raw}} & & \multicolumn{2}{c}{\textbf{DrQv2-Normalized}} \\
        \cmidrule(lr){2-3} \cmidrule(lr){5-6}
        & Mean  & IQM && Mean & IQM \\
        \midrule

        OG-SPR (complete)
        & \textbf{626}
        & \textbf{730}
        &
        & \textbf{5.05}
        & {1.61} \\
        \midrule
        OG-SPR - OP
        & 615
        & 712
        &
        & 4.29
        & \textbf{1.63} \\
        
        OG-SPR - SP
        & 608
        & 704
        &
        & 2.62
        & 1.26 \\
    
        OG-SPR - SVP
        & 613
        & 706
        &
        & 4.60
        & {1.59} \\
        \midrule
        {Shared Adapter}
        & 611
        & 702
        &
        & 4.38
        & 1.48 \\

        {No Adapter}
        & 597
        & 676
        &
        & 3.73
        & 1.39 \\

        {Adapter 2 Only}
        & 603
        & 691
        &
        & 3.54
        & 1.41 \\

        \bottomrule
    \end{tabular}
    \caption{
    Aggregate ablation results on the DMControl benchmark. The mean and IQM are reported for both raw scores and DrQv2-Normalized scores. Bold numbers indicate the best performance.
    }
    \label{ab_results}
\end{table}

\subsection{Analyzing State and State-Action Representations}
To understand how self-prediction and observation prediction shape representations, we analyze the state and state-action representations learned by OG-SPR and its two ablated variants, \textbf{OG-SPR - OP} and \textbf{OG-SPR - SP}.
We freeze the learned representation networks and train lightweight linear predictors to recover the proprioceptive states corresponding to the visual observations.
We consider two prediction tasks. The first predicts the current proprioceptive state from the latent state representation, with the prediction loss denoted by $\mathcal{L}_\text{state}$. The second predicts the next proprioceptive state from the state-action representation, with the loss denoted by $\mathcal{L}_\text{next-state}$.
Details and results are provided in the appendix.

The two ablated variants exhibit distinct strengths. The observation-predictive variant \textbf{OG-SPR - SP} generally achieves lower $\mathcal{L}_\text{next-state}$ than the self-predictive variant \textbf{OG-SPR - OP}, suggesting that observation prediction helps state-action representations capture more accurate dynamics information. 
In contrast, the self-predictive variant tends to achieve lower $\mathcal{L}_\text{state}$, suggesting that self-prediction learns state representations that are more informative for recovering the underlying proprioceptive state.
These observations are consistent with the different inductive biases of the two predictive objectives. 
Multi-step latent self-prediction primarily regularizes the state representation to be temporally predictive over horizons, without explicitly constraining latent transitions to align with observation-level dynamics. 
In contrast, next-observation prediction learns the state-action representation through next-observation reconstruction, encouraging it to capture observation-grounded dynamics information, while not directly imposing a multi-step predictability constraint on the latent state representation.

OG-SPR inherits the characteristics of both objectives. Its $\mathcal{L}_\text{state}$ is close to that of \textbf{OG-SPR - OP}, while its $\mathcal{L}_\text{next-state}$ often lies between those of \textbf{OG-SPR - OP} and \textbf{OG-SPR - SP}. Overall, OG-SPR strikes a balance between extracting useful information from the current pixel input and capturing accurate dynamics information.

\subsection{Evaluation on Discrete-Action Tasks}
Although our main focus is visual continuous control, we also evaluate OG-SPR on Atari100k \cite{DBLP:conf/iclr/KaiserBMOCCEFKL20}, a benchmark for data-efficient discrete-action visual RL, to examine how the proposed design behaves when adapted to a discrete-action domain. 
We adopt a lightweight discrete-action adaptation following MR.Q and AnonMethod, by modifying the output activation of the policy network.
The implementation details and hyperparameters are provided in the appendix. 
We report aggregated human-normalized scores, the standard evaluation metric for the Atari benchmark. Similar to the DrQv2-normalized score, we also consider SPR-normalized scores, where each game score is normalized by the corresponding score of SPR \cite{DBLP:conf/iclr/SchwarzerAGHCB21}, a strong self-predictive baseline for this benchmark.
As shown in Table~\ref{atari_results}, OG-SPR remains competitive on Atari100k, ranking first or second on three of the four aggregate metrics. 
Together with the main DMControl results, these findings suggest that OG-SPR is particularly effective in continuous-control domains while remaining competitive in discrete-action settings.

One possible explanation for the less pronounced improvements on Atari100k is that adapting continuous-control algorithms to discrete-action domains may require nontrivial algorithmic modifications for cross-domain adaptation.
The Dreamer family provides a representative example. DreamerV2 \cite{DBLP:conf/iclr/HafnerL0B21} introduced discrete latent representations when extending Dreamer \cite{DBLP:conf/iclr/HafnerLB020} to Atari, while DreamerV3 \cite{hafner2025dreamerv3} further introduced a set of cross-domain stabilization techniques for stable learning across diverse domains. Our Atari experiments instead use a minimal discrete-action adaptation, allowing us to examine whether OG-SPR transfers to discrete-action domains without substantial cross-domain modifications. 
We leave the development of robust cross-domain mechanisms for model-free visual RL to future work.

\begin{table}[t]
\centering
\begin{tabular}{lcccc}
\toprule
\textbf{Metrics }& \textbf{SPR} & \textbf{MR.Q} & \textbf{AnonMethod} & \textbf{OG-SPR} \\
\midrule
\multicolumn{5}{l}{\textit{Human-Normalized Results}} \\
\midrule
Mean & 0.66 & \underline{0.90} & \textbf{0.93}  & 0.81  \\
IQM & \textbf{0.43} & 0.40 & {0.40}  & \underline{0.42}  \\
\midrule
\multicolumn{5}{l}{\textit{SPR-Normalized Results}} \\
\midrule
Mean & 1.00 & 1.12 & \textbf{1.25}  & \underline{1.24}  \\
IQM  & 1.00 & 0.99 & \underline{1.10}  & \textbf{1.11} \\
\bottomrule
\end{tabular}
\caption{
Aggregate results on Atari100k. We report the mean and IQM under both human-normalized and SPR-normalized scores. Bold and underline indicate the best and second-best results, respectively.
}
\label{atari_results}
\end{table}

\section{Conclusion and Limitations}
In this paper, we presented Observation-Grounded Self-Predictive Representations (OG-SPR), a model-free visual RL algorithm that improves data efficiency by integrating multi-step latent self-prediction, next-observation prediction, and short-term value prediction within an actor-critic framework. To alleviate the direct constraint imposed by latent self-prediction on the shared representation, OG-SPR introduces two lightweight adapters that form an adapter-mediated self-prediction branch. Experiments on DMControl show that OG-SPR improves sample efficiency and achieves new state-of-the-art aggregate performance on the benchmark, with particularly pronounced gains in challenging domains. Despite the considerable room for improvement in these challenging domains under limited data budgets, OG-SPR provides an initial step toward data-efficient RL for complex visual continuous control tasks.

Several limitations remain. Although OG-SPR can be adapted to discrete-action domains, its gains are less consistent than those observed in visual continuous control.
Further work is needed to make the method robust across different action spaces. Besides, OG-SPR uses fixed auxiliary loss weights across tasks, which may not be optimal for all tasks. Future work could explore adaptive weighting strategies to better coordinate multiple auxiliary objectives. 

{
\small
\bibliography{aaai2027}}

@inproceedings{DBLP:conf/iclr/YaratsFLP22,
  author       = {Denis Yarats and
                  Rob Fergus and
                  Alessandro Lazaric and
                  Lerrel Pinto},
  title        = {Mastering Visual Continuous Control: Improved Data-Augmented Reinforcement
                  Learning},
  booktitle    = {The Tenth International Conference on Learning Representations, {ICLR}
                  2022, Virtual Event, April 25-29, 2022},
  publisher    = {OpenReview.net},
  year         = {2022},
  url          = {https://openreview.net/forum?id=\_SJ-\_yyes8},
  bibsource    = {dblp computer science bibliography, https://dblp.org}
}

@inproceedings{DBLP:conf/iclr/YaratsKF21,
  author       = {Denis Yarats and
                  Ilya Kostrikov and
                  Rob Fergus},
  title        = {Image Augmentation Is All You Need: Regularizing Deep Reinforcement
                  Learning from Pixels},
  booktitle    = {9th International Conference on Learning Representations, {ICLR} 2021,
                  Virtual Event, Austria, May 3-7, 2021},
  publisher    = {OpenReview.net},
  year         = {2021},
  url          = {https://openreview.net/forum?id=GY6-6sTvGaf},
  bibsource    = {dblp computer science bibliography, https://dblp.org}
}

@inproceedings{DBLP:conf/iclr/FujimotoD0TR25,
  author       = {Scott Fujimoto and
                  Pierluca D'Oro and
                  Amy Zhang and
                  Yuandong Tian and
                  Michael Rabbat},
  title        = {Towards General-Purpose Model-Free Reinforcement Learning},
  booktitle    = {The Thirteenth International Conference on Learning Representations,
                  {ICLR} 2025, Singapore, April 24-28, 2025},
  publisher    = {OpenReview.net},
  year         = {2025},
  url          = {https://openreview.net/forum?id=R1hIXdST22},
  bibsource    = {dblp computer science bibliography, https://dblp.org}
}

@inproceedings{DBLP:conf/iclr/SchwarzerAGHCB21,
  author       = {Max Schwarzer and
                  Ankesh Anand and
                  Rishab Goel and
                  R. Devon Hjelm and
                  Aaron C. Courville and
                  Philip Bachman},
  title        = {Data-Efficient Reinforcement Learning with Self-Predictive Representations},
  booktitle    = {9th International Conference on Learning Representations, {ICLR} 2021,
                  Virtual Event, Austria, May 3-7, 2021},
  publisher    = {OpenReview.net},
  year         = {2021},
  url          = {https://openreview.net/forum?id=uCQfPZwRaUu},
  bibsource    = {dblp computer science bibliography, https://dblp.org}
}

@inproceedings{DBLP:conf/icml/SchwarzerOCBAC23,
  author       = {Max Schwarzer and
                  Johan S. Obando{-}Ceron and
                  Aaron C. Courville and
                  Marc G. Bellemare and
                  Rishabh Agarwal and
                  Pablo Samuel Castro},
  editor       = {Andreas Krause and
                  Emma Brunskill and
                  Kyunghyun Cho and
                  Barbara Engelhardt and
                  Sivan Sabato and
                  Jonathan Scarlett},
  title        = {Bigger, Better, Faster: Human-level Atari with human-level efficiency},
  booktitle    = {International Conference on Machine Learning, {ICML} 2023, 23-29 July
                  2023, Honolulu, Hawaii, {USA}},
  series       = {Proceedings of Machine Learning Research},
  volume       = {202},
  pages        = {30365--30380},
  publisher    = {{PMLR}},
  year         = {2023},
  url          = {https://proceedings.mlr.press/v202/schwarzer23a.html},
  bibsource    = {dblp computer science bibliography, https://dblp.org}
}

@article{DBLP:journals/corr/abs-2506-05418,
  author       = {Kyungsoo Kim and
                  Jeongsoo Ha and
                  Yusung Kim},
  title        = {Self-Predictive Dynamics for Generalization of Vision-based Reinforcement
                  Learning},
  journal      = {CoRR},
  volume       = {abs/2506.05418},
  year         = {2025},
  url          = {https://doi.org/10.48550/arXiv.2506.05418},
  doi          = {10.48550/ARXIV.2506.05418},
  eprinttype   = {arXiv},
  eprint       = {2506.05418},
  bibsource    = {dblp computer science bibliography, https://dblp.org}
}

@inproceedings{DBLP:conf/nips/GrillSATRBDPGAP20,
  author       = {Jean{-}Bastien Grill and
                  Florian Strub and
                  Florent Altch{\'{e}} and
                  Corentin Tallec and
                  Pierre H. Richemond and
                  Elena Buchatskaya and
                  Carl Doersch and
                  Bernardo {\'{A}}vila Pires and
                  Zhaohan Guo and
                  Mohammad Gheshlaghi Azar and
                  Bilal Piot and
                  Koray Kavukcuoglu and
                  R{\'{e}}mi Munos and
                  Michal Valko},
  editor       = {Hugo Larochelle and
                  Marc'Aurelio Ranzato and
                  Raia Hadsell and
                  Maria{-}Florina Balcan and
                  Hsuan{-}Tien Lin},
  title        = {Bootstrap Your Own Latent - {A} New Approach to Self-Supervised Learning},
  booktitle    = {Advances in Neural Information Processing Systems 33: Annual Conference
                  on Neural Information Processing Systems 2020, NeurIPS 2020, December
                  6-12, 2020, virtual},
  year         = {2020},
  url          = {https://proceedings.neurips.cc/paper/2020/hash/f3ada80d5c4ee70142b17b8192b2958e-Abstract.html},
  bibsource    = {dblp computer science bibliography, https://dblp.org}
}

@inproceedings{DBLP:conf/icml/GeladaKBNB19,
  author       = {Carles Gelada and
                  Saurabh Kumar and
                  Jacob Buckman and
                  Ofir Nachum and
                  Marc G. Bellemare},
  editor       = {Kamalika Chaudhuri and
                  Ruslan Salakhutdinov},
  title        = {DeepMDP: Learning Continuous Latent Space Models for Representation
                  Learning},
  booktitle    = {Proceedings of the 36th International Conference on Machine Learning,
                  {ICML} 2019, 9-15 June 2019, Long Beach, California, {USA}},
  series       = {Proceedings of Machine Learning Research},
  volume       = {97},
  pages        = {2170--2179},
  publisher    = {{PMLR}},
  year         = {2019},
  url          = {http://proceedings.mlr.press/v97/gelada19a.html},
  bibsource    = {dblp computer science bibliography, https://dblp.org}
}

@article{hafner2025dreamerv3,
  title={Mastering diverse control tasks through world models},
  author={Hafner, Danijar and Pasukonis, Jurgis and Ba, Jimmy and Lillicrap, Timothy},
  journal={Nature},
  pages={1--7},
  year={2025},
  publisher={Nature Publishing Group}
}

@inproceedings{DBLP:conf/iclr/00010024,
  author       = {Nicklas Hansen and
                  Hao Su and
                  Xiaolong Wang},
  title        = {{TD-MPC2:} Scalable, Robust World Models for Continuous Control},
  booktitle    = {The Twelfth International Conference on Learning Representations,
                  {ICLR} 2024, Vienna, Austria, May 7-11, 2024},
  publisher    = {OpenReview.net},
  year         = {2024},
  url          = {https://openreview.net/forum?id=Oxh5CstDJU},
  bibsource    = {dblp computer science bibliography, https://dblp.org}
}

@book{sutton1998reinforcement,
  title={Reinforcement learning: An introduction},
  author={Sutton, Richard S and Barto, Andrew G and others},
  volume={1},
  year={1998},
  publisher={MIT press Cambridge}
}

@article{DBLP:journals/corr/abs-1801-00690,
  author       = {Yuval Tassa and
                  Yotam Doron and
                  Alistair Muldal and
                  Tom Erez and
                  Yazhe Li and
                  Diego de Las Casas and
                  David Budden and
                  Abbas Abdolmaleki and
                  Josh Merel and
                  Andrew Lefrancq and
                  Timothy P. Lillicrap and
                  Martin A. Riedmiller},
  title        = {DeepMind Control Suite},
  journal      = {CoRR},
  volume       = {abs/1801.00690},
  year         = {2018},
  url          = {http://arxiv.org/abs/1801.00690},
  eprinttype   = {arXiv},
  eprint       = {1801.00690},
  bibsource    = {dblp computer science bibliography, https://dblp.org}
}

@inproceedings{DBLP:conf/iclr/MaL0L0000T24,
  author       = {Guozheng Ma and
                  Lu Li and
                  Sen Zhang and
                  Zixuan Liu and
                  Zhen Wang and
                  Yixin Chen and
                  Li Shen and
                  Xueqian Wang and
                  Dacheng Tao},
  title        = {Revisiting Plasticity in Visual Reinforcement Learning: Data, Modules
                  and Training Stages},
  booktitle    = {The Twelfth International Conference on Learning Representations,
                  {ICLR} 2024, Vienna, Austria, May 7-11, 2024},
  publisher    = {OpenReview.net},
  year         = {2024},
  url          = {https://openreview.net/forum?id=0aR1s9YxoL},
  bibsource    = {dblp computer science bibliography, https://dblp.org}
}

@inproceedings{DBLP:conf/nips/EysenbachZLS22,
  author       = {Benjamin Eysenbach and
                  Tianjun Zhang and
                  Sergey Levine and
                  Ruslan Salakhutdinov},
  editor       = {Sanmi Koyejo and
                  S. Mohamed and
                  A. Agarwal and
                  Danielle Belgrave and
                  K. Cho and
                  A. Oh},
  title        = {Contrastive Learning as Goal-Conditioned Reinforcement Learning},
  booktitle    = {Advances in Neural Information Processing Systems 35: Annual Conference
                  on Neural Information Processing Systems 2022, NeurIPS 2022, New Orleans,
                  LA, USA, November 28 - December 9, 2022},
  year         = {2022},
  url          = {http://papers.nips.cc/paper\_files/paper/2022/hash/e7663e974c4ee7a2b475a4775201ce1f-Abstract-Conference.html},
  bibsource    = {dblp computer science bibliography, https://dblp.org}
}

@inproceedings{DBLP:conf/iclr/LiuTE25,
  author       = {Grace Liu and
                  Michael Tang and
                  Benjamin Eysenbach},
  title        = {A Single Goal is All You Need: Skills and Exploration Emerge from
                  Contrastive {RL} without Rewards, Demonstrations, or Subgoals},
  booktitle    = {The Thirteenth International Conference on Learning Representations,
                  {ICLR} 2025, Singapore, April 24-28, 2025},
  publisher    = {OpenReview.net},
  year         = {2025},
  url          = {https://openreview.net/forum?id=xCkgX4Xfu0},
  bibsource    = {dblp computer science bibliography, https://dblp.org}
}

@inproceedings{DBLP:conf/aaai/LiaoZ023,
  author       = {Weijian Liao and
                  Zongzhang Zhang and
                  Yang Yu},
  editor       = {Brian Williams and
                  Yiling Chen and
                  Jennifer Neville},
  title        = {Policy-Independent Behavioral Metric-Based Representation for Deep
                  Reinforcement Learning},
  booktitle    = {Thirty-Seventh {AAAI} Conference on Artificial Intelligence, {AAAI}
                  2023, Thirty-Fifth Conference on Innovative Applications of Artificial
                  Intelligence, {IAAI} 2023, Thirteenth Symposium on Educational Advances
                  in Artificial Intelligence, {EAAI} 2023, Washington, DC, USA, February
                  7-14, 2023},
  pages        = {8746--8754},
  publisher    = {{AAAI} Press},
  year         = {2023},
  url          = {https://doi.org/10.1609/aaai.v37i7.26052},
  doi          = {10.1609/AAAI.V37I7.26052},
  bibsource    = {dblp computer science bibliography, https://dblp.org}
}

@article{DBLP:journals/corr/abs-2507-18519,
  author       = {Leiji Zhang and
                  Zeyu Wang and
                  Xin Li and
                  Yao{-}Hui Li},
  title        = {Revisiting Bisimulation Metric for Robust Representations in Reinforcement
                  Learning},
  journal      = {CoRR},
  volume       = {abs/2507.18519},
  year         = {2025},
  url          = {https://doi.org/10.48550/arXiv.2507.18519},
  doi          = {10.48550/ARXIV.2507.18519},
  eprinttype   = {arXiv},
  eprint       = {2507.18519},
  bibsource    = {dblp computer science bibliography, https://dblp.org}
}

@inproceedings{DBLP:conf/nips/FujimotoCSGPM23,
  author       = {Scott Fujimoto and
                  Wei{-}Di Chang and
                  Edward J. Smith and
                  Shixiang Gu and
                  Doina Precup and
                  David Meger},
  editor       = {Alice Oh and
                  Tristan Naumann and
                  Amir Globerson and
                  Kate Saenko and
                  Moritz Hardt and
                  Sergey Levine},
  title        = {For {SALE:} State-Action Representation Learning for Deep Reinforcement
                  Learning},
  booktitle    = {Advances in Neural Information Processing Systems 36: Annual Conference
                  on Neural Information Processing Systems 2023, NeurIPS 2023, New Orleans,
                  LA, USA, December 10 - 16, 2023},
  year         = {2023},
  url          = {http://papers.nips.cc/paper\_files/paper/2023/hash/c20ac0df6c213db6d3a930fe9c7296c8-Abstract-Conference.html},
  bibsource    = {dblp computer science bibliography, https://dblp.org}
}

@inproceedings{DBLP:conf/icml/OtaOJMN20,
  author       = {Kei Ota and
                  Tomoaki Oiki and
                  Devesh K. Jha and
                  Toshisada Mariyama and
                  Daniel Nikovski},
  title        = {Can Increasing Input Dimensionality Improve Deep Reinforcement Learning?},
  booktitle    = {Proceedings of the 37th International Conference on Machine Learning,
                  {ICML} 2020, 13-18 July 2020, Virtual Event},
  series       = {Proceedings of Machine Learning Research},
  volume       = {119},
  pages        = {7424--7433},
  publisher    = {{PMLR}},
  year         = {2020},
  url          = {http://proceedings.mlr.press/v119/ota20a.html},
  bibsource    = {dblp computer science bibliography, https://dblp.org}
}

@inproceedings{DBLP:conf/iclr/HafnerLB020,
  author       = {Danijar Hafner and
                  Timothy P. Lillicrap and
                  Jimmy Ba and
                  Mohammad Norouzi},
  title        = {Dream to Control: Learning Behaviors by Latent Imagination},
  booktitle    = {8th International Conference on Learning Representations, {ICLR} 2020,
                  Addis Ababa, Ethiopia, April 26-30, 2020},
  publisher    = {OpenReview.net},
  year         = {2020},
  url          = {https://openreview.net/forum?id=S1lOTC4tDS},
  bibsource    = {dblp computer science bibliography, https://dblp.org}
}

@inproceedings{DBLP:conf/iclr/HafnerL0B21,
  author       = {Danijar Hafner and
                  Timothy P. Lillicrap and
                  Mohammad Norouzi and
                  Jimmy Ba},
  title        = {Mastering Atari with Discrete World Models},
  booktitle    = {9th International Conference on Learning Representations, {ICLR} 2021,
                  Virtual Event, Austria, May 3-7, 2021},
  publisher    = {OpenReview.net},
  year         = {2021},
  url          = {https://openreview.net/forum?id=0oabwyZbOu},
  bibsource    = {dblp computer science bibliography, https://dblp.org}
}

@inproceedings{DBLP:conf/icml/FujimotoHM18,
  author       = {Scott Fujimoto and
                  Herke van Hoof and
                  David Meger},
  editor       = {Jennifer G. Dy and
                  Andreas Krause},
  title        = {Addressing Function Approximation Error in Actor-Critic Methods},
  booktitle    = {Proceedings of the 35th International Conference on Machine Learning,
                  {ICML} 2018, Stockholmsm{\"{a}}ssan, Stockholm, Sweden, July
                  10-15, 2018},
  series       = {Proceedings of Machine Learning Research},
  volume       = {80},
  pages        = {1582--1591},
  publisher    = {{PMLR}},
  year         = {2018},
  url          = {http://proceedings.mlr.press/v80/fujimoto18a.html},
  bibsource    = {dblp computer science bibliography, https://dblp.org}
}

@inproceedings{DBLP:conf/iros/TodorovET12,
  author       = {Emanuel Todorov and
                  Tom Erez and
                  Yuval Tassa},
  title        = {MuJoCo: {A} physics engine for model-based control},
  booktitle    = {2012 {IEEE/RSJ} International Conference on Intelligent Robots and
                  Systems, {IROS} 2012, Vilamoura, Algarve, Portugal, October 7-12,
                  2012},
  pages        = {5026--5033},
  publisher    = {{IEEE}},
  year         = {2012},
  url          = {https://doi.org/10.1109/IROS.2012.6386109},
  doi          = {10.1109/IROS.2012.6386109},
  bibsource    = {dblp computer science bibliography, https://dblp.org}
}

@inproceedings{DBLP:conf/nips/PaszkeGMLBCKLGA19,
  author       = {Adam Paszke and
                  Sam Gross and
                  Francisco Massa and
                  Adam Lerer and
                  James Bradbury and
                  Gregory Chanan and
                  Trevor Killeen and
                  Zeming Lin and
                  Natalia Gimelshein and
                  Luca Antiga and
                  Alban Desmaison and
                  Andreas K{\"{o}}pf and
                  Edward Z. Yang and
                  Zachary DeVito and
                  Martin Raison and
                  Alykhan Tejani and
                  Sasank Chilamkurthy and
                  Benoit Steiner and
                  Lu Fang and
                  Junjie Bai and
                  Soumith Chintala},
  editor       = {Hanna M. Wallach and
                  Hugo Larochelle and
                  Alina Beygelzimer and
                  Florence d'Alch{\'{e}}{-}Buc and
                  Emily B. Fox and
                  Roman Garnett},
  title        = {PyTorch: An Imperative Style, High-Performance Deep Learning Library},
  booktitle    = {Advances in Neural Information Processing Systems 32: Annual Conference
                  on Neural Information Processing Systems 2019, NeurIPS 2019, December
                  8-14, 2019, Vancouver, BC, Canada},
  pages        = {8024--8035},
  year         = {2019},
  url          = {https://proceedings.neurips.cc/paper/2019/hash/bdbca288fee7f92f2bfa9f7012727740-Abstract.html},
  bibsource    = {dblp computer science bibliography, https://dblp.org}
}

@inproceedings{DBLP:conf/iclr/KaiserBMOCCEFKL20,
  author       = {Lukasz Kaiser and
                  Mohammad Babaeizadeh and
                  Piotr Milos and
                  Blazej Osinski and
                  Roy H. Campbell and
                  Konrad Czechowski and
                  Dumitru Erhan and
                  Chelsea Finn and
                  Piotr Kozakowski and
                  Sergey Levine and
                  Afroz Mohiuddin and
                  Ryan Sepassi and
                  George Tucker and
                  Henryk Michalewski},
  title        = {Model Based Reinforcement Learning for Atari},
  booktitle    = {8th International Conference on Learning Representations, {ICLR} 2020,
                  Addis Ababa, Ethiopia, April 26-30, 2020},
  publisher    = {OpenReview.net},
  year         = {2020},
  url          = {https://openreview.net/forum?id=S1xCPJHtDB},
  bibsource    = {dblp computer science bibliography, https://dblp.org}
}

@article{DBLP:journals/jair/BellemareNVB13,
  author       = {Marc G. Bellemare and
                  Yavar Naddaf and
                  Joel Veness and
                  Michael Bowling},
  title        = {The Arcade Learning Environment: An Evaluation Platform for General
                  Agents},
  journal      = {J. Artif. Intell. Res.},
  volume       = {47},
  pages        = {253--279},
  year         = {2013},
  url          = {https://doi.org/10.1613/jair.3912},
  doi          = {10.1613/JAIR.3912},
  bibsource    = {dblp computer science bibliography, https://dblp.org}
}

@article{DBLP:journals/corr/abs-2406-02696,
  author       = {Aidan Scannell and
                  Kalle Kujanp{\"{a}}{\"{a}} and
                  Yi Zhao and
                  Mohammadreza Nakhaei and
                  Arno Solin and
                  Joni Pajarinen},
  title        = {iQRL - Implicitly Quantized Representations for Sample-efficient Reinforcement
                  Learning},
  journal      = {CoRR},
  volume       = {abs/2406.02696},
  year         = {2024},
  url          = {https://doi.org/10.48550/arXiv.2406.02696},
  doi          = {10.48550/ARXIV.2406.02696},
  eprinttype   = {arXiv},
  eprint       = {2406.02696},
  bibsource    = {dblp computer science bibliography, https://dblp.org}
}

@inproceedings{DBLP:conf/nips/ZhengWSMZXDH23,
  author       = {Ruijie Zheng and
                  Xiyao Wang and
                  Yanchao Sun and
                  Shuang Ma and
                  Jieyu Zhao and
                  Huazhe Xu and
                  Hal Daum{\'{e}} III and
                  Furong Huang},
  editor       = {Alice Oh and
                  Tristan Naumann and
                  Amir Globerson and
                  Kate Saenko and
                  Moritz Hardt and
                  Sergey Levine},
  title        = {{TACO:} Temporal Latent Action-Driven Contrastive Loss for Visual
                  Reinforcement Learning},
  booktitle    = {Advances in Neural Information Processing Systems 36: Annual Conference
                  on Neural Information Processing Systems 2023, NeurIPS 2023, New Orleans,
                  LA, USA, December 10 - 16, 2023},
  year         = {2023},
  url          = {http://papers.nips.cc/paper\_files/paper/2023/hash/96d00450ed65531ffe2996daed487536-Abstract-Conference.html},
  bibsource    = {dblp computer science bibliography, https://dblp.org}
}

@inproceedings{DBLP:conf/icml/SilverLHDWR14,
  author       = {David Silver and
                  Guy Lever and
                  Nicolas Heess and
                  Thomas Degris and
                  Daan Wierstra and
                  Martin A. Riedmiller},
  title        = {Deterministic Policy Gradient Algorithms},
  booktitle    = {Proceedings of the 31th International Conference on Machine Learning,
                  {ICML} 2014, Beijing, China, 21-26 June 2014},
  series       = {{JMLR} Workshop and Conference Proceedings},
  volume       = {32},
  pages        = {387--395},
  publisher    = {JMLR.org},
  year         = {2014},
  url          = {http://proceedings.mlr.press/v32/silver14.html},
  bibsource    = {dblp computer science bibliography, https://dblp.org}
}

@article{DBLP:journals/nature/SchrittwieserAH20,
  author       = {Julian Schrittwieser and
                  Ioannis Antonoglou and
                  Thomas Hubert and
                  Karen Simonyan and
                  Laurent Sifre and
                  Simon Schmitt and
                  Arthur Guez and
                  Edward Lockhart and
                  Demis Hassabis and
                  Thore Graepel and
                  Timothy P. Lillicrap and
                  David Silver},
  title        = {Mastering Atari, Go, chess and shogi by planning with a learned model},
  journal      = {Nat.},
  volume       = {588},
  number       = {7839},
  pages        = {604--609},
  year         = {2020},
  url          = {https://doi.org/10.1038/s41586-020-03051-4},
  doi          = {10.1038/S41586-020-03051-4},
  bibsource    = {dblp computer science bibliography, https://dblp.org}
}

@inproceedings{DBLP:conf/icml/WangSHHLF16,
  author       = {Ziyu Wang and
                  Tom Schaul and
                  Matteo Hessel and
                  Hado van Hasselt and
                  Marc Lanctot and
                  Nando de Freitas},
  editor       = {Maria{-}Florina Balcan and
                  Kilian Q. Weinberger},
  title        = {Dueling Network Architectures for Deep Reinforcement Learning},
  booktitle    = {Proceedings of the 33nd International Conference on Machine Learning,
                  {ICML} 2016, New York City, NY, USA, June 19-24, 2016},
  series       = {{JMLR} Workshop and Conference Proceedings},
  volume       = {48},
  pages        = {1995--2003},
  publisher    = {JMLR.org},
  year         = {2016},
  url          = {http://proceedings.mlr.press/v48/wangf16.html},
  bibsource    = {dblp computer science bibliography, https://dblp.org}
}

@inproceedings{DBLP:conf/icml/HansenSW22,
  author       = {Nicklas Hansen and
                  Hao Su and
                  Xiaolong Wang},
  editor       = {Kamalika Chaudhuri and
                  Stefanie Jegelka and
                  Le Song and
                  Csaba Szepesv{\'{a}}ri and
                  Gang Niu and
                  Sivan Sabato},
  title        = {Temporal Difference Learning for Model Predictive Control},
  booktitle    = {International Conference on Machine Learning, {ICML} 2022, 17-23 July
                  2022, Baltimore, Maryland, {USA}},
  series       = {Proceedings of Machine Learning Research},
  volume       = {162},
  pages        = {8387--8406},
  publisher    = {{PMLR}},
  year         = {2022},
  url          = {https://proceedings.mlr.press/v162/hansen22a.html},
  bibsource    = {dblp computer science bibliography, https://dblp.org}
}

@article{DBLP:journals/corr/MnihKSGAWR13,
  author       = {Volodymyr Mnih and
                  Koray Kavukcuoglu and
                  David Silver and
                  Alex Graves and
                  Ioannis Antonoglou and
                  Daan Wierstra and
                  Martin A. Riedmiller},
  title        = {Playing Atari with Deep Reinforcement Learning},
  journal      = {CoRR},
  volume       = {abs/1312.5602},
  year         = {2013},
  url          = {http://arxiv.org/abs/1312.5602},
  eprinttype   = {arXiv},
  eprint       = {1312.5602},
  bibsource    = {dblp computer science bibliography, https://dblp.org}
}

@article{DBLP:journals/tmlr/EchchahedC25,
  author       = {Ayoub Echchahed and
                  Pablo Samuel Castro},
  title        = {A Survey of State Representation Learning for Deep Reinforcement Learning},
  journal      = {Trans. Mach. Learn. Res.},
  volume       = {2025},
  year         = {2025},
  url          = {https://openreview.net/forum?id=gOk34vUHtz},
  bibsource    = {dblp computer science bibliography, https://dblp.org}
}

@inproceedings{DBLP:conf/nips/FujimotoMP20,
  author       = {Scott Fujimoto and
                  David Meger and
                  Doina Precup},
  editor       = {Hugo Larochelle and
                  Marc'Aurelio Ranzato and
                  Raia Hadsell and
                  Maria{-}Florina Balcan and
                  Hsuan{-}Tien Lin},
  title        = {An Equivalence between Loss Functions and Non-Uniform Sampling in
                  Experience Replay},
  booktitle    = {Advances in Neural Information Processing Systems 33: Annual Conference
                  on Neural Information Processing Systems 2020, NeurIPS 2020, December
                  6-12, 2020, virtual},
  year         = {2020},
  url          = {https://proceedings.neurips.cc/paper/2020/hash/a3bf6e4db673b6449c2f7d13ee6ec9c0-Abstract.html},
  bibsource    = {dblp computer science bibliography, https://dblp.org}
}

@inproceedings{DBLP:conf/iclr/LoshchilovH19,
  author       = {Ilya Loshchilov and
                  Frank Hutter},
  title        = {Decoupled Weight Decay Regularization},
  booktitle    = {7th International Conference on Learning Representations, {ICLR} 2019,
                  New Orleans, LA, USA, May 6-9, 2019},
  publisher    = {OpenReview.net},
  year         = {2019},
  url          = {https://openreview.net/forum?id=Bkg6RiCqY7},
  bibsource    = {dblp computer science bibliography, https://dblp.org}
}

@article{DBLP:journals/nature/MnihKSRVBGRFOPB15,
  author       = {Volodymyr Mnih and
                  Koray Kavukcuoglu and
                  David Silver and
                  Andrei A. Rusu and
                  Joel Veness and
                  Marc G. Bellemare and
                  Alex Graves and
                  Martin A. Riedmiller and
                  Andreas Fidjeland and
                  Georg Ostrovski and
                  Stig Petersen and
                  Charles Beattie and
                  Amir Sadik and
                  Ioannis Antonoglou and
                  Helen King and
                  Dharshan Kumaran and
                  Daan Wierstra and
                  Shane Legg and
                  Demis Hassabis},
  title        = {Human-level control through deep reinforcement learning},
  journal      = {Nat.},
  volume       = {518},
  number       = {7540},
  pages        = {529--533},
  year         = {2015},
  url          = {https://doi.org/10.1038/nature14236},
  doi          = {10.1038/NATURE14236},
  bibsource    = {dblp computer science bibliography, https://dblp.org}
}

@inproceedings{DBLP:conf/aaai/HesselMHSODHPAS18,
  author       = {Matteo Hessel and
                  Joseph Modayil and
                  Hado van Hasselt and
                  Tom Schaul and
                  Georg Ostrovski and
                  Will Dabney and
                  Dan Horgan and
                  Bilal Piot and
                  Mohammad Gheshlaghi Azar and
                  David Silver},
  editor       = {Sheila A. McIlraith and
                  Kilian Q. Weinberger},
  title        = {Rainbow: Combining Improvements in Deep Reinforcement Learning},
  booktitle    = {Proceedings of the Thirty-Second {AAAI} Conference on Artificial Intelligence,
                  (AAAI-18), the 30th innovative Applications of Artificial Intelligence
                  (IAAI-18), and the 8th {AAAI} Symposium on Educational Advances in
                  Artificial Intelligence (EAAI-18), New Orleans, Louisiana, USA, February
                  2-7, 2018},
  pages        = {3215--3222},
  publisher    = {{AAAI} Press},
  year         = {2018},
  url          = {https://doi.org/10.1609/aaai.v32i1.11796},
  doi          = {10.1609/AAAI.V32I1.11796},
  bibsource    = {dblp computer science bibliography, https://dblp.org}
}

@misc{jax2018github,
  author = {James Bradbury and Roy Frostig and Peter Hawkins and Matthew James Johnson and Yash Katariya and Chris Leary and Dougal Maclaurin and George Necula and Adam Paszke and Jake VanderPlas and Skye Wanderman-Milne and Qiao Zhang},
  title = {{JAX}: composable transformations of {Python}+{NumPy} programs},
  url = {https://github.com/jax-ml/jax},
  year = {2018},
}

@misc{anonymous2026preliminary,
  author = {Anonymous},
  title  = {Title Withheld for Anonymous Review},
  year   = {2026},
  note   = {Citation suppressed to preserve double-blind review}
}

\end{document}


\twocolumn[
\begin{center}
\section*{Supplementary Material}
\end{center}
]

\appendix

\section{Pseudocode of OG-SPR}
Algorithm~\ref{algorithm} summarizes the training procedure of OG-SPR.

\begin{algorithm}[tb]
\caption{Training Procedure of OG-SPR}
\label{algorithm}
Let $\theta_V$ denote the parameters of the online value network, including the encoder, adapters, state-action encoder, value predictors, and auxiliary prediction heads (i.e., the decoder and the short-term value predictor).

Let $\theta_\pi$ denote the parameters of the online policy network.

Let $\theta'_V$ and $\theta'_\pi$ denote the target parameters used for bootstrapped value estimation and latent self-prediction targets.

Let $T_\text{target}$ denote the update frequency of target networks.

\begin{algorithmic}[1]
\STATE Initialize replay buffer $\mathcal{B}$ and current training step $t$
\WHILE{Training}
\STATE Collect experience $(s, a, r, s')$ and add to buffer $\mathcal{B}$
\STATE Sample a minibatch of sequences $({s}_i, a_i, r_{i:i+n-1}, o_{i+1}, {s}_{i+n}) \sim \mathcal{B}$
\STATE Compute latent states $h_i$ and state-action representations $z_i$ 
\STATE Compute $\mathcal{L}_\text{Value}$, $\mathcal{L}_\text{n-step}$ and $\mathcal{L}_\text{Rec}$ based on the same $z_i$
\STATE Compute $\mathcal{L}_\text{Policy}$ using the same detached $h_i$ 
\STATE Sample another minibatch of sequences $({s}_{i:i+K}, a_{i:i+K-1}) \sim \mathcal{B}$
\STATE Compute $\mathcal{L}_\text{Self-predictive}$ 
\STATE Compute the combined loss $\mathcal{L}$ 
\STATE Update $\theta_V$ by minimizing $\mathcal{L}$
\STATE Update $\theta_\pi$ by minimizing $\mathcal{L}_\text{Policy}$
\STATE $t \gets t+1$
\IF{$t \bmod T_\text{target} = 0$}
\STATE $\{\theta'_V, \theta'_\pi\} \gets \{\theta_V, \theta_\pi\}$
\ENDIF
\ENDWHILE
\end{algorithmic}
\end{algorithm}
\section{Overview of AnonMethod}
\label{app:anonmethod_overview}

AnonMethod~\cite{anonymous2026preliminary} is an observation-predictive model-free RL method that learns representations by predicting dynamics in observation space.
It learns representations through next-observation prediction in a normalized observation space, in contrast to self-predictive methods that predict future latent embeddings.

In AnonMethod, observations are first normalized and then encoded into latent state representations.
This latent state is combined with the action to produce a state-action representation, which is fed to three predictors for value learning, short-term value prediction, and next-observation prediction.
The policy network takes the latent state as input with gradients stopped.

AnonMethod introduces two auxiliary tasks.
The first is short-term value prediction, which predicts the discounted reward accumulated over a short horizon and provides a stabilizing training signal for representation learning.
The second is next-observation prediction, where a decoder reconstructs the next normalized observation from the state-action representation.
This objective encourages the learned representation to capture observation-grounded dynamics information. For reproducibility, we include the implementation used for AnonMethod in the supplementary material.

\section{Additional Details for Experimental Setup}

\subsection{Computing Infrastructure}
All experiments were conducted on a server running Ubuntu 24.04.4 LTS (Noble Numbat) on an x86\_64 architecture. The server was equipped with two Intel Xeon Platinum 8368Q CPUs operating at 2.60 GHz, providing 76 physical CPU cores in total, 251 GiB of system memory, and four NVIDIA GeForce RTX 4090 GPUs. The software environment consisted of Python 3.9.23 and PyTorch 2.6.0.

\subsection{Randomness Control}
For each method or variant on each task, we conduct five independent runs using the random seeds $\{42, 99, 123, 520, 668\}$. We control the random number generators used by Python, NumPy, and PyTorch and enable deterministic behavior for supported cuDNN operations.

\section{Implementation Details}
\subsection{Network Architecture}
We follow MR.Q in constructing the actor-critic backbone of OG-SPR.
The observation encoder consists of four convolutional (Conv) layers, each with 32 output channels, a kernel size of 3, strides (2, 2, 2, 1) across the four layers, and ELU activations. The convolutional layers are followed by a linear layer applied to the flattened output, layer normalization (LayerNorm), and a final ELU activation.
The state-action encoder is a three-layer MLP, with LayerNorm and ELU applied after each of the first two layers. 
Each long-term value predictor is a four-layer MLP, with LayerNorm and ELU applied after each of the first three layers. 
Similar to AnonMethod, the decoder consists of a linear layer that projects the input vector to a feature map, followed by four transposed convolutional (TransConv) layers. 
Spectral normalization is applied to the first layer of the decoder. 
The short-term value predictor is a two-layer MLP, where the first layer is followed by LayerNorm and an ELU activation.
The policy network is a three-layer MLP with LayerNorm followed by ReLU after the first two layers. 
The source code for OG-SPR is provided in the supplementary material.

\subsection{Other Design Choices}
For design choices that are orthogonal to our main representation-learning contribution, we align OG-SPR with dynamics-based baselines MR.Q and AnonMethod. 
This controlled setup helps attribute performance differences more reliably to 
the key design differences introduced by OG-SPR. 
Specifically, we use the LAP \cite{DBLP:conf/nips/FujimotoMP20} replay buffer to perform prioritized sampling during training. All networks are trained with the AdamW optimizer \cite{DBLP:conf/iclr/LoshchilovH19}. 

\subsection{Modifications for Atari100k}
Atari100k \cite{DBLP:journals/jair/BellemareNVB13, DBLP:conf/iclr/KaiserBMOCCEFKL20} is a data-efficiency benchmark consisting of 26 Atari games for discrete control. Agents are allowed 100k steps of environment interaction with an action repeat of 4, corresponding to 400k frames and approximately two hours of game time. We use the no-sticky-action setting. 
The input state is constructed by stacking the most recent four frames, which are gray-scaled, resized to 84 $\times$ 84 pixels, and set to the max between the 3rd and 4th frame \cite{DBLP:journals/nature/MnihKSRVBGRFOPB15}. 
When aggregating scores, we compute human-normalized scores using the human scores reported in \cite{DBLP:conf/icml/WangSHHLF16}:
{\small
\begin{equation}
\label{eq:14}
\begin{aligned}
\operatorname{Human-Normalized}(x) = \frac{x-\text{random score}}{\text{Human score} - \text{random score}}.
\end{aligned}
\end{equation}}

To adapt OG-SPR to discrete-action domains, we modify the policy network using discrete-action adaptations adopted by MR.Q and AnonMethod. Specifically, the action is computed as follows:
\begin{equation}
\label{eq:1}
\begin{gathered}
a^{\pi'}= \operatorname{argmax}(a'), \\
a'=\pi'(h') + \operatorname{clip}(\epsilon, -c, c), \quad\epsilon \sim \mathcal{N}(0, \sigma^2).
\end{gathered}
\end{equation}
Discrete actions are represented by a one-hot encoding, with the Gaussian noise added to each dimension. The activation function applied after the pre-activation policy outputs is Gumbel-Softmax in discrete-action settings. 

For the Atari100k experiments, we keep the observation-prediction and short-term value prediction weights fixed at $\lambda_\text{Rec}=1.0$ and $\lambda_\text{n-step}=1.0$, which are the default settings in AnonMethod.
To determine $\lambda_\text{Self-predictive}$, we use the default self-prediction loss weight of SPR, 2.0, as the starting point and perform a grid search with a step size of 1.0 on a small subset of games containing only \textit{Alien} and \textit{Amidar}. The selected value is $\lambda_\text{Self-predictive}=2.0$, which is then fixed for all games.
While OG-SPR performs one gradient update for both the value network and the policy network per environment step in DMControl, we perform two gradient updates for both networks per environment step on Atari100k following SPR \cite{DBLP:conf/iclr/SchwarzerAGHCB21}.
OG-SPR clips rewards to $\{-1, 0, +1\}$ according to their signs, which is a preprocessing method commonly used for Atari games \cite{DBLP:journals/nature/MnihKSRVBGRFOPB15, DBLP:conf/aaai/HesselMHSODHPAS18, DBLP:conf/iclr/SchwarzerAGHCB21}.

\subsection{Observation Prediction Targets} The observation for pixel-based benchmarks is composed of several previous frames (three for DMControl and four for Atari100k). This raises a question of how to construct learning targets for the next observation prediction. A naive approach is to construct a \textit{complete} form of the observation. However, such a target contains redundant information from the current observation and may hinder the model from efficiently learning dynamics-aware representations. We thus remove the overlapping frames between the next and the current observations when constructing targets.

\subsection{Hyperparameters}
Table~\ref{tab:hyperparameters} summarizes the default hyperparameters of OG-SPR.

\section{Computational Efficiency Comparison}

Table~\ref{runtime} compares the training wall-clock time of OG-SPR with baselines on the \textit{cheetah-run} task.
For this runtime comparison only, we use a smaller DreamerV3 configuration instead of its default 10M-parameter configuration, making its model size comparable to those of the other methods.
Since the official DreamerV3 implementation is based on JAX~\cite{jax2018github}, we use a PyTorch implementation of DreamerV3\footnote{\url{https://github.com/NM512/dreamerv3-torch/}} for a framework-consistent comparison with the other PyTorch-based methods.
All methods are evaluated on a single NVIDIA GeForce RTX 4090 GPU.

\begin{table}[htb]
    \centering
    \begin{tabular}{lcc}
        \toprule
        \textbf{Methods} & \textbf{Params (M)} & \textbf{Wall-Clock Time (h)}  \\
        \midrule
        DreamerV3 & 5.3 & 32.7    \\
        OG-SPR (ours) & 5.4 & 14.2  \\
        MR.Q & 4.2 & 12.0  \\
        TD-MPC2 & 5.4 & 9.4 \\
        AnonMethod & 4.9 & 7.7  \\
        DrQ-v2 & 4.4 & 5.7  \\
        \bottomrule
    \end{tabular}
    \caption{
    Comparison of training wall-clock time for 500k environment steps on the \textit{cheetah-run} task.
    }
    \label{runtime}
\end{table}

\begin{table*}[!hb]
\small
\centering
\begin{tabular}{lll}
\toprule
\textbf{Components} & \textbf{Hyperparameter} & \textbf{Value} \\
\midrule

\multirow{4}{*}{Value Learning} 
    & n-step returns & 3 \\
    & Auxiliary loss weights $(\lambda_\text{Rec}, \lambda_\text{n-step}, \lambda_\text{Self-predictive})$ & \makecell[l]{DMControl: (0.1, 1.0, 5.0) \\  Atari100k: (1.0, 1.0, 2.0)} \\
    & Discount factor $\gamma$ & 0.99 \\

\midrule

\multirow{2}{*}{TD3} 
    & Target policy noise $\sigma$ & $ \mathcal{N}(0,0.2^2)$ \\
    & Target policy noise clipping $c$ & (-0.3, 0.3) \\
\midrule

\multirow{2}{*}{LAP} 
    & Probability smoothing $\alpha$ & 0.4 \\
    & Minimum priority & 1 \\
\midrule

\multirow{1}{*}{Auxiliary Tasks} 
    & Horizon of latent self-prediction $K$ & 5 \\
\midrule

\multirow{7}{*}{Optimization} 
    & Optimizer & AdamW \\
    & Learning rate & 3e-4 \\
    & Weight decay & 1e-4 \\
    & Mini-batch size & 256 \\
    & Target update frequency & 250 \\
    & Gradient updates per training step (DMControl) &  value network: 1 , policy network: 1 \\
    & Gradient updates per training step (Atari100k) &  value network: 2 , policy network: 2 \\
\midrule

\multirow{3}{*}{Exploration} 
    & Initial random exploration time steps & \makecell[l]{DMControl: 10k \\ Atari100k: 2k} \\
    & Exploration noise & \makecell[l]{DMControl: $\mathcal{N}(0,0.1^2)$ \\ Atari100k: $\mathcal{N}(0,0.2^2)$}  \\
\midrule

\multirow{4}{*}{Observation Encoder} 
    & Structure & Conv + MLP \\
    & State dimension & 512 \\
    & Activation function & ELU \\
    & Gradient clip norm & 20 \\
\midrule

\multirow{4}{*}{State-Action Encoder} 
    & Hidden dimension & 580 \\
    & State-action representation dimension & 512 \\
    & Activation function & ELU \\
    & Gradient clip norm & 20 \\
\midrule

\multirow{3}{*}{Long-term value predictor} 
    & Hidden dimension & 512 \\
    & Activation function & ELU \\
    & Gradient clip norm & 20 \\
\midrule

\multirow{5}{*}{Short-term value predictor} 
    & Hidden dimension & 512 \\
    & Reward bins & 65 \\
    & Activation function & ELU \\
    & Gradient clip norm & 20 \\
    & Reward range & $[-10, 10]$ (effective: $[-22\text{k}, 22\text{k}]$) \\
\midrule

\multirow{6}{*}{Decoder} 
    & Structure & MLP + TransConv \\
    & Latent channels & 128 \\
    & Output channels & \makecell[l]{DMControl: 3 \\ Atari100k: 1} \\
    & Activation function & ReLU \\
    & Gradient clip norm & 20 \\
\midrule

\multirow{3}{*}{Policy Network} 
    & Hidden dimension & 512 \\
    & Activation function & ReLU \\
    & Gumbel-Softmax $\tau$ (Atari100k) & 10 \\
\bottomrule
\end{tabular}
\caption{Default hyperparameters of OG-SPR.}
\label{tab:hyperparameters}
\end{table*}

\section{Additional Representation Analysis}
This section provides additional details and complete results for the linear probing analysis introduced in the main text. 
The goal of this analysis is to examine how latent self-prediction and observation prediction affect the information encoded in the learned latent state and state-action representations. 
We conduct this analysis on four DMControl tasks: \textit{dog-trot}, \textit{humanoid-walk}, \textit{quadruped-walk}, and \textit{cheetah-run}.
The first three tasks are selected as representative cases where OG-SPR outperforms both ablated variants, \textbf{OG-SPR - SP} and \textbf{OG-SPR - OP}, which remove self-prediction (SP) and observation prediction (OP), respectively.
The fourth task, \textit{cheetah-run}, is selected as a case where OG-SPR does not achieve the best performance among the compared variants.

For each method, selected task, and final checkpoint, we freeze the learned representation modules, including the visual encoder $f$, the online Adapter 1 $u_1$, and the state-action encoder $g$. We then train lightweight linear predictors on top of the frozen representations to evaluate how much proprioceptive information is encoded in the learned latent state and state-action representations.
Let $s_t$ denote the visual input state and $x_t$ denote the corresponding proprioceptive state at time step $t$.
We normalize the proprioceptive states using the online normalization method proposed in AnonMethod \cite{anonymous2026preliminary} and use the normalized states $\bar{x}$ as prediction targets. 
This normalization reduces the influence of proprioceptive dimensions with larger numerical scales on the aggregated prediction loss.

We consider two prediction tasks. 
The first predicts the current proprioceptive state from the latent state representation, with the prediction loss denoted by $\mathcal{L}_\text{state}$. 
The second predicts the next proprioceptive state from the state-action representation, with the loss denoted by $\mathcal{L}_\text{next-state}$.
For the current-state prediction, we train a linear predictor $p_\text{state}$ on top of the latent state representation:
\begin{equation}
h_t = u_1(f(s_t)),
\end{equation}
and minimize the prediction loss:
\begin{equation}
\mathcal{L}_\text{state}
= \left\|p_\text{state}(h_t) - \bar{x}_t \right\|_2^2.
\end{equation}
For the next-state prediction, we train another linear predictor $p_\text{next}$ on top of the state-action representation:
\begin{equation}
z_t = g(h_t, a_t),
\end{equation}
and minimize the following loss:
\begin{equation}
\mathcal{L}_\text{next-state}
= \left\|p_\text{next}(z_t) - \bar{x}_{t+1} \right\|_2^2.
\end{equation}
The representation networks are kept fixed throughout training, and only the linear predictors are updated.
Thus, lower prediction losses indicate that the corresponding proprioceptive information is more easily recoverable from the frozen representations by a simple linear model.

\begin{table*}[t]
\centering
\begin{tabular}{lcccc}
\toprule
\textbf{Methods } & $\mathcal{L}_\text{state}$ & $\mathcal{L}_\text{next-state}$ & $\mathcal{L}_\text{state}+\mathcal{L}_\text{next-state}$ & \textbf{Scores} \\
\midrule
\multicolumn{5}{l}{\textit{dog-trot}} \\
\midrule
OG-SPR & 0.802 [0.768, 0.834] & {1.042} [1.023, 1.057]  & {1.845} [1.815, 1.891] & {105} [79, 128] \\
OG-SPR - OP  & 0.803 [0.768, 0.844] & {1.053} [1.019, 1.075]  & {1.856} [1.815, 1.919] & 79 [70, 88] \\
OG-SPR - SP  & 0.984 [0.947, 1.019] & {0.908} [0.872, 0.930]  & {1.892} [1.858, 1.949] & 62 [59, 66] \\
\midrule
\multicolumn{5}{l}{{\textit{humanoid-walk}}} \\
\midrule
OG-SPR & 0.666 [0.507, 0.765] & 1.184 [1.151, 1.229]  & {1.850} [1.679, 1.955] & {15} [7, 22] \\
OG-SPR - OP  & 0.666 [0.504, 0.767] & 1.193 [1.146, 1.256]  & 1.860 [1.760, 1.913] & 12 [7, 17] \\
OG-SPR - SP  & 1.119 [1.024, 1.260] & 1.142 [1.089, 1.225]  & 2.261 [2.136, 2.485] & 2 [2, 3] \\
\midrule
\multicolumn{5}{l}{{\textit{quadruped-walk}}} \\
\midrule
OG-SPR & 0.565 [0.544, 0.602] & 1.185 [1.135, 1.276] & {1.750} [1.679, 1.825] & 837 [805, 884] \\
OG-SPR - OP & 0.563 [0.543, 0.597] & 1.189 [1.139, 1.250] & 1.752 [1.721, 1.800] & 679 [458, 900] \\
OG-SPR - SP & 1.848 [1.575, 2.125] & 1.088 [1.054, 1.155] & 2.936 [2.629, 3.181] & 795 [752, 849] \\
\midrule
\multicolumn{5}{l}{{\textit{cheetah-run}}} \\
\midrule
OG-SPR & 0.0350 [0.0292, 0.0413] & 1.600 [1.540, 1.663] & 1.635 [1.567, 1.700] & 780 [772, 789] \\
OG-SPR - OP & 0.0399 [0.0298, 0.0508] & 1.541 [1.464, 1.614] & 1.581 [1.495, 1.654] & 684 [558, 760] \\
OG-SPR - SP & 3.729 [3.382, 4.032] & 1.587 [1.525, 1.645] & 5.316 [4.998, 5.638] & 854 [823, 881] \\
\bottomrule
\end{tabular}
\caption{
Linear prediction results on four selected DMControl tasks. Brackets denote 95\% bootstrap confidence intervals.
}
\label{probing}
\end{table*}

\subsection{Training Details}
All linear predictors are trained with the Adam optimizer using a learning rate of $3 \times 10^{-4}$.
We apply gradient clipping with a maximum norm of 20.
For each prediction task, the linear predictor is trained for 100k gradient updates with data collected by running a fully trained policy.

Since different policies may induce different state-action data distributions, we further control for the data-collection policy.
For each method, selected task, final checkpoint, and prediction task, we train three linear predictors, each using data collected by executing the trained actor from one of the three methods: OG-SPR, \textbf{OG-SPR - SP}, or \textbf{OG-SPR - OP}.
For example, in the next-state prediction experiment on \textit{dog-trot} with a random seed of 42, we train linear predictors using data collected by the OG-SPR actor, the \textbf{OG-SPR - SP} actor, and the \textbf{OG-SPR - OP} actor, respectively.
For each fully trained run, we average the resulting prediction losses $\mathcal{L}_\text{state}$ and $\mathcal{L}_\text{next-state}$ over the three data-collection policies, and then report the average across seeds.
This protocol reduces the potential advantage from distribution matching. If each method's linear predictor were trained and evaluated only on data collected by its own policy, its prediction loss could be lower than it would be under a shared data distribution, simply because the data used for linear prediction matches that method's induced state-action distribution. Averaging over data collected by all three policies therefore provides a fairer comparison of representation quality.
Table~\ref{probing} summarizes the linear prediction results.

\section{Complete Results}
Tables~\ref{tab:full_dmc},~\ref{tab:ab_results}, and~\ref{main_result_atari} report the complete per-task results corresponding to the aggregate results in the main results, ablation study and Atari100k benchmark, respectively. 

\begin{table*}[t]
    \setlength{\tabcolsep}{1mm}
    \small
    \centering
    \begin{tabular}{lcccccc}
        \toprule
        \textbf{Task} & \textbf{DrQ-v2} & \textbf{TD-MPC2} &
        \textbf{DreamerV3} & \textbf{MR.Q} & \textbf{AnonMethod} & \textbf{OG-SPR} \\
        \midrule
        acrobot-swingup & 
        168 {{[127, 219]}} & 
        177 {{[131, 223]}} & 
        260 {{[162, 337]}} & 
        303 {{[266, 340]}} &
        282 {{[207, 356]}} &
        443 {{[397, 494]}} \\

        ball\_in\_cup-catch & 
        909 {{[821, 973]}} & 
        746 {{[494, 942]}} & 
        968 {{[965, 971]}} & 
        974 {{[970, 977]}} &
        975 {{[973, 979]}} &
        980 {{[977, 983]}} \\
        
        cartpole-balance & 
        993 {{[990, 996]}} & 
        883 {{[665, 995]}} & 
        992 {{[992, 992]}} & 
        998 {{[997, 999]}} &
        997 {{[996, 999]}} &
        999 {{[998, 999]}} \\
        
        \makecell[l]{cartpole-\\balance\_sparse} & 
        962 {{[887, 1000]}} & 
        1000 {{[1000, 1000]}} & 
        1000 {{[1000, 1000]}} & 
        1000 {{[1000, 1000]}} &
        1000 {{[1000, 1000]}} &
        1000 {{[1000, 1000]}} \\
        
        cartpole-swingup & 
        864 {{[854, 873]}} & 
        790 {{[676, 851]}} & 
        851 {{[847, 857]}} &  
        865 {{[859, 872]}} & 
        873 {{[867, 878]}} &
        878 {{[871, 883]}} \\

        \makecell[l]{cartpole-\\swingup\_sparse} & 
        774 {{[741, 805]}} & 
        158 {{[0, 475]}} & 
        285 {{[0, 570]}} & 
        777 {{[734, 818]}} & 
        840 {{[833, 845]}} &
        784 {{[769, 814]}} \\
        
        cheetah-run & 
        728 {{[701, 753]}} & 
        548 {{[452, 612]}} & 
        733 {{[694, 772]}} & 
        759 {{[752, 768]}} & 
        835 {{[814, 867]}} &
        780 {{[772, 789]}} \\
        
        dog-run & 
        10 {{[9, 12]}} & 
        8 {{[5, 11]}} & 
        29 {{[23, 36]}} &   
        51 {{[45, 56]}} & 
        55 {{[46, 69]}} &
        90 {{[64, 115]}} \\
        
        dog-stand & 
        43 {{[37, 49]}} & 
        161 {{[147, 175]}} & 
        118 {{[78, 157]}} &  
        251 {{[233, 269]}} & 
        239 {{[214, 263]}}  &
        357 {{[267, 448]}} \\
        
        dog-trot & 
        14 {{[11, 18]}} & 
        16 {{[12, 18]}} & 
        44 {{[34, 54]}} &  
        70 {{[61, 80]}} & 
        61 {{[59, 63]}} &
        105 {{[79, 128]}} \\

        dog-walk & 
        22 {{[18, 29]}} & 
        14 {{[12, 17]}} & 
        42 {{[31, 52]}} &
        90 {{[79, 101]}} & 
        81 {{[72, 91]}} &
        108 {{[102, 118]}} \\
        
        finger-spin & 
        860 {{[787, 922]}} & 
        987 {{[985, 989]}} & 
        843 {{[704, 981]}} & 
        835 {{[682, 982]}} & 
        987 {{[985, 989]}} &
        981 {{[973, 986]}} \\
        
        finger-turn\_easy & 
        503 {{[399, 615]}} & 
        757 {{[647, 883]}} & 
        732 {{[245, 981]}} &  
        955 {{[918, 975]}} & 
        934 {{[889, 975]}} &
        947 {{[907, 976]}} \\
        
        finger-turn\_hard & 
        223 {{[121, 340]}} & 
        816 {{[770, 862]}} & 
        560 {{[177, 943]}} &  
        948 {{[910, 970]}} & 
        927 {{[878, 976]}}  &
        928 {{[853, 972]}} \\
        
        fish-swim & 
        84 {{[65, 107]}} & 
        67 {{[52, 85]}} & 
        126 {{[61, 223]}} &  
        74 {{[67, 82]}} & 
        68 {{[66, 70]}} &
        86 {{[58, 115]}} \\

        hopper-hop & 
        224 {{[170, 278]}} & 
        234 {{[220, 247]}} & 
        294 {{[289, 299]}} & 
        236 {{[183, 289]}} & 
        276 {{[256, 295]}} &
        293 {{[263, 325]}} \\
        
        hopper-stand & 
        917 {{[903, 931]}} & 
        755 {{[624, 886]}} & 
        873 {{[850, 896]}} & 
        925 {{[919, 930]}} & 
        899 {{[848, 930]}} &
        945 {{[935, 955]}} \\
        
        humanoid-run & 
        1 {{[1, 1]}} & 
        1 {{[1, 1]}} & 
        1 {{[1, 2]}} &  
        1 {{[1, 2]}} & 
        1 {{[1, 1]}} &
        3 {{[2, 4]}} \\

        \makecell[l]{humanoid-\\run (1.0M)} & 
        1 {{[1, 2]}} & 
        1 {{[1, 1]}} & 
        1 {{[1, 2]}} &  
        1 {{[1, 2]}} & 
        1 {{[1, 2]}} &
        13 {{[5, 25]}} \\

        \makecell[l]{humanoid-\\run (1.5M)} &  
        2 {{[2, 3]}} & 
        1 {{[1, 1]}} & 
        1 {{[1, 2]}} &  
        3 {{[2, 5]}} & 
        1 {{[1, 2]}} &
        67 {{[60, 71]}} \\

        \makecell[l]{humanoid-\\run (2.0M)} &  
        2 {{[2, 3]}} & 
        1 {{[1, 1]}} & 
        4 {{[1, 9]}} &  
        27 {{[4, 50]}} & 
        1 {{[1, 1]}} &
        100 {{[94, 106]}} \\
        
        humanoid-stand & 
        6 {{[6, 7]}} & 
        6 {{[6, 7]}} & 
        7 {{[3, 10]}} &  
        8 {{[7, 8]}} & 
        6 {{[5, 7]}}  &
        12 {{[8, 20]}} \\

        \makecell[l]{humanoid-\\stand (1.0M)} & 
        8 {{[7, 9]}} & 
        7 {{[7, 7]}} & 
        8 {{[6, 10]}} &  
        7 {{[7, 8]}} & 
        9 {{[7, 11]}} &
        100 {{[50, 151]}} \\

        \makecell[l]{humanoid-\\stand (1.5M)} &  
        8 {{[8, 9]}} & 
        7 {{[6, 7]}} & 
        6 {{[3, 8]}} &  
        14 {{[9, 20]}} & 
        8 {{[7, 9]}} &
        248 {{[195, 300]}} \\

        \makecell[l]{humanoid-\\stand (2.0M)} &  
        9 {{[8, 11]}} & 
        7 {{[6, 7]}} & 
        11 {{[7, 17]}} &  
        79 {{[8, 200]}} & 
        31 {{[8, 55]}} &
        339 {{[284, 401]}} \\
        
        humanoid-walk & 
        2 {{[2, 2]}} & 
        2 {{[1, 2]}} & 
        2 {{[1, 3]}} &  
        3 {{[2, 5]}} & 
        2 {{[2, 3]}} &
        15 {{[7, 22]}} \\

        \makecell[l]{humanoid-\\walk (1.0M)} & 
        3 {{[2, 3]}} & 
        2 {{[2, 2]}} & 
        2 {{[1, 3]}} &  
        3 {{[2, 3]}} & 
        3 {{[2, 4]}} &
        86 {{[44, 143]}} \\

        \makecell[l]{humanoid-\\walk (1.5M)} &  
        5 {{[2, 10]}} & 
        2 {{[2, 2]}} & 
        2 {{[1, 2]}} &  
        5 {{[2, 10]}} & 
        9 {{[2, 23]}} &
        226 {{[220, 232]}} \\

        \makecell[l]{humanoid-\\walk (2.0M)} &  
        19 {{[2, 52]}} & 
        2 {{[2, 2]}} & 
        2 {{[1, 3]}} &  
        48 {{[3, 93]}} & 
        54 {{[3, 153]}} &
        316 {{[299, 333]}} \\

        pendulum-swingup & 
        838 {{[813, 861]}} & 
        735 {{[506, 875]}} & 
        845 {{[717, 936]}} & 
        840 {{[830, 853]}} &
        836 {{[819, 855]}} &
        856 {{[830, 877]}} \\
        
        quadruped-run & 
        459 {{[412, 507]}} & 
        280 {{[158, 369]}} & 
        475 {{[433, 498]}} & 
        503 {{[474, 518]}} & 
        460 {{[458, 463]}} &
        543 {{[490, 599]}} \\
        
        quadruped-walk & 
        750 {{[699, 796]}} & 
        338 {{[282, 393]}} & 
        512 {{[293, 709]}} &  
        865 {{[799, 921]}} & 
        839 {{[817, 857]}} &
        837 {{[805, 884]}} \\
        
        reacher-easy & 
        938 {{[903, 973]}} & 
        961 {{[928, 979]}} & 
        778 {{[392, 978]}} &  
        978 {{[974, 981]}} & 
        935 {{[894, 976]}}  &
        956 {{[909, 983]}} \\
        
        reacher-hard & 
        705 {{[580, 831]}} & 
        644 {{[490, 772]}} & 
        965 {{[954, 978]}} &  
        953 {{[913, 974]}} & 
        956 {{[915, 979]}} &
        946 {{[899, 971]}} \\

        walker-run & 
        546 {{[475, 612]}} & 
        710 {{[681, 738]}} & 
        717 {{[664, 748]}} &  
        562 {{[502, 619]}} & 
        683 {{[669, 705]}} &
        707 {{[678, 736]}} \\
        
        walker-stand & 
        980 {{[977, 984]}} & 
        914 {{[852, 957]}} & 
        982 {{[977, 987]}} &  
        985 {{[984, 986]}} & 
        988 {{[987, 990]}}  &
        982 {{[977, 985]}} \\
        
        walker-walk & 
        766 {{[489, 957]}} & 
        897 {{[832, 925]}} & 
        968 {{[957, 981]}} &  
        964 {{[959, 969]}} & 
        963 {{[958, 968]}} &
        968 {{[964, 971]}} \\

        \bottomrule
    \end{tabular}
    \caption{
    Final raw episode returns on DMControl after 500k environment steps (default), corresponding to 1M frames under action repeat of 2. \textbf{1.0M}, \textbf{1.5M}, and \textbf{2.0M} denote results evaluated at 1.0M, 1.5M, and 2.0M environment steps, respectively.
    The {[bracketed values]} represent a 95\% bootstrap confidence interval. 
    }
    \label{tab:full_dmc}
\end{table*}

\begin{figure*}[t]
\centering
\includegraphics[width=0.246\textwidth]{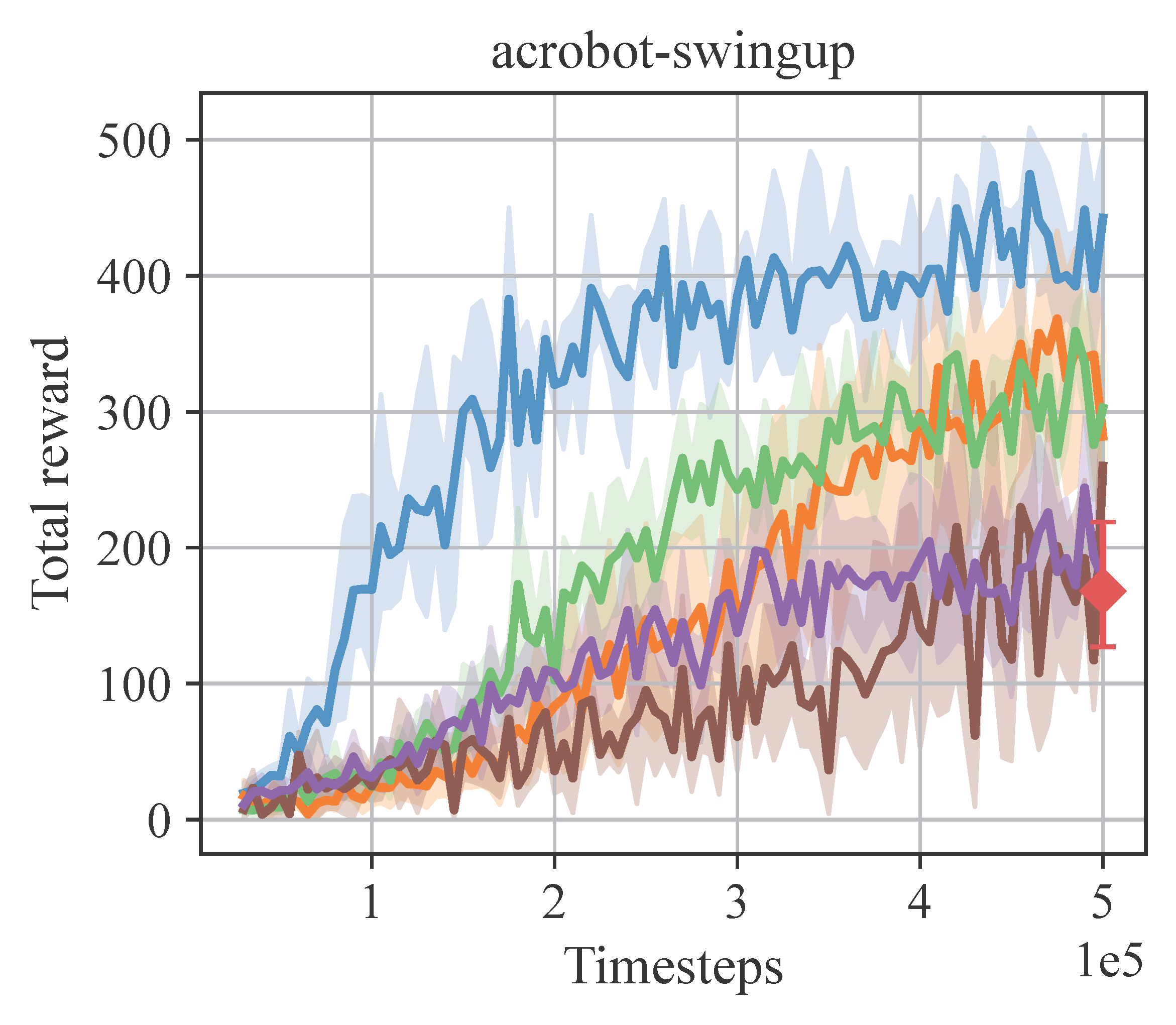}
\includegraphics[width=0.246\textwidth]{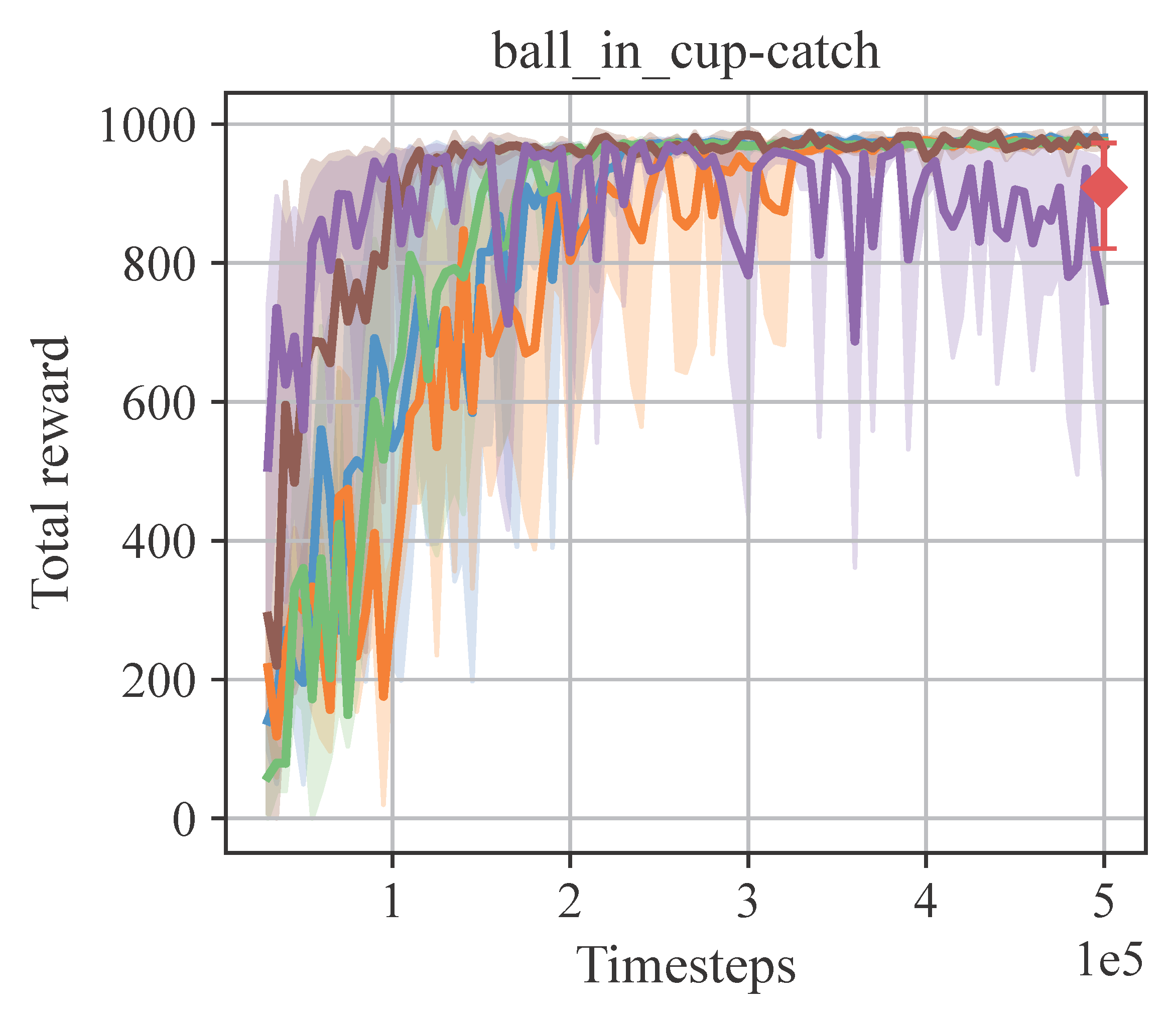}
\includegraphics[width=0.246\textwidth]{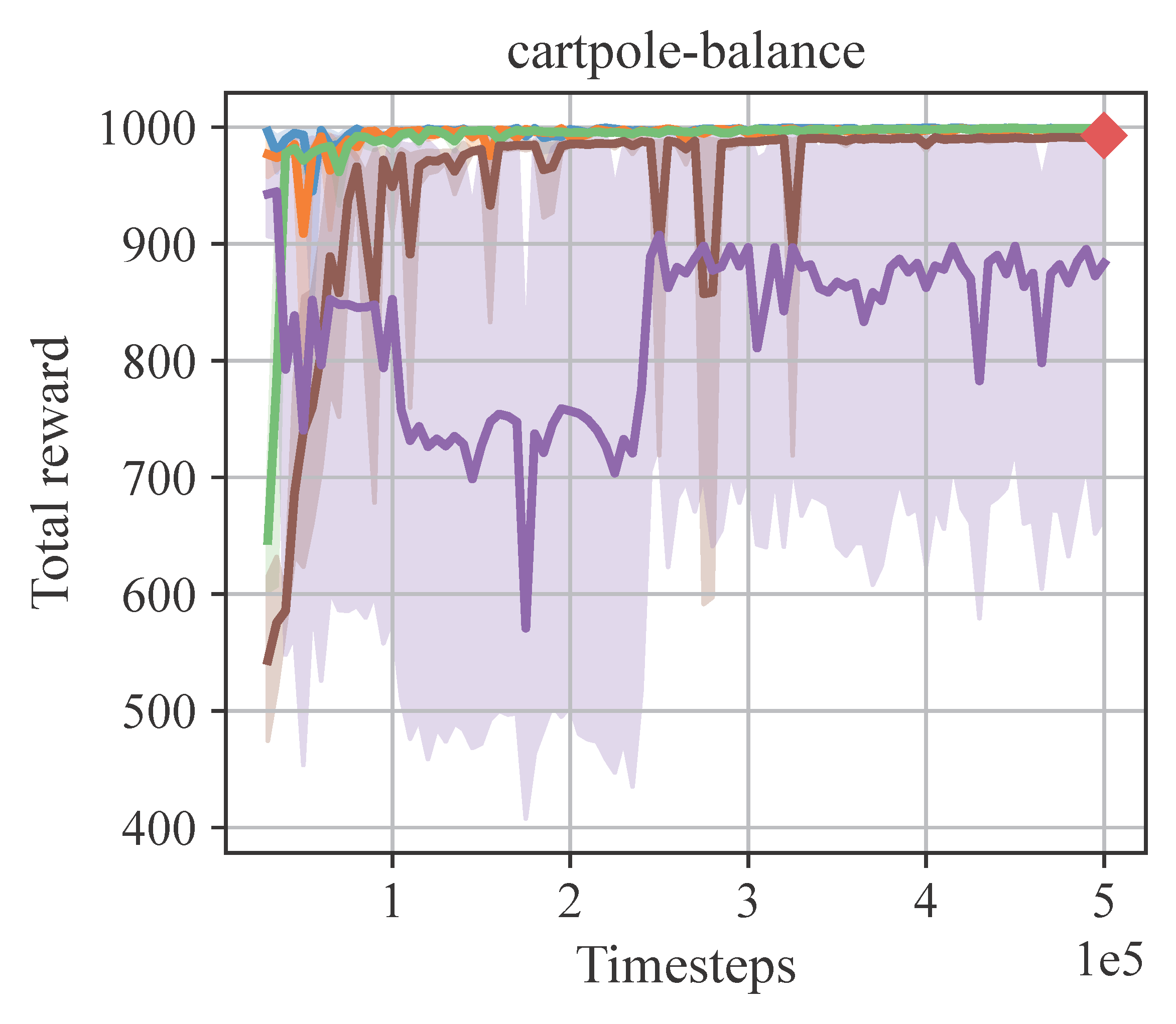}
\includegraphics[width=0.246\textwidth]{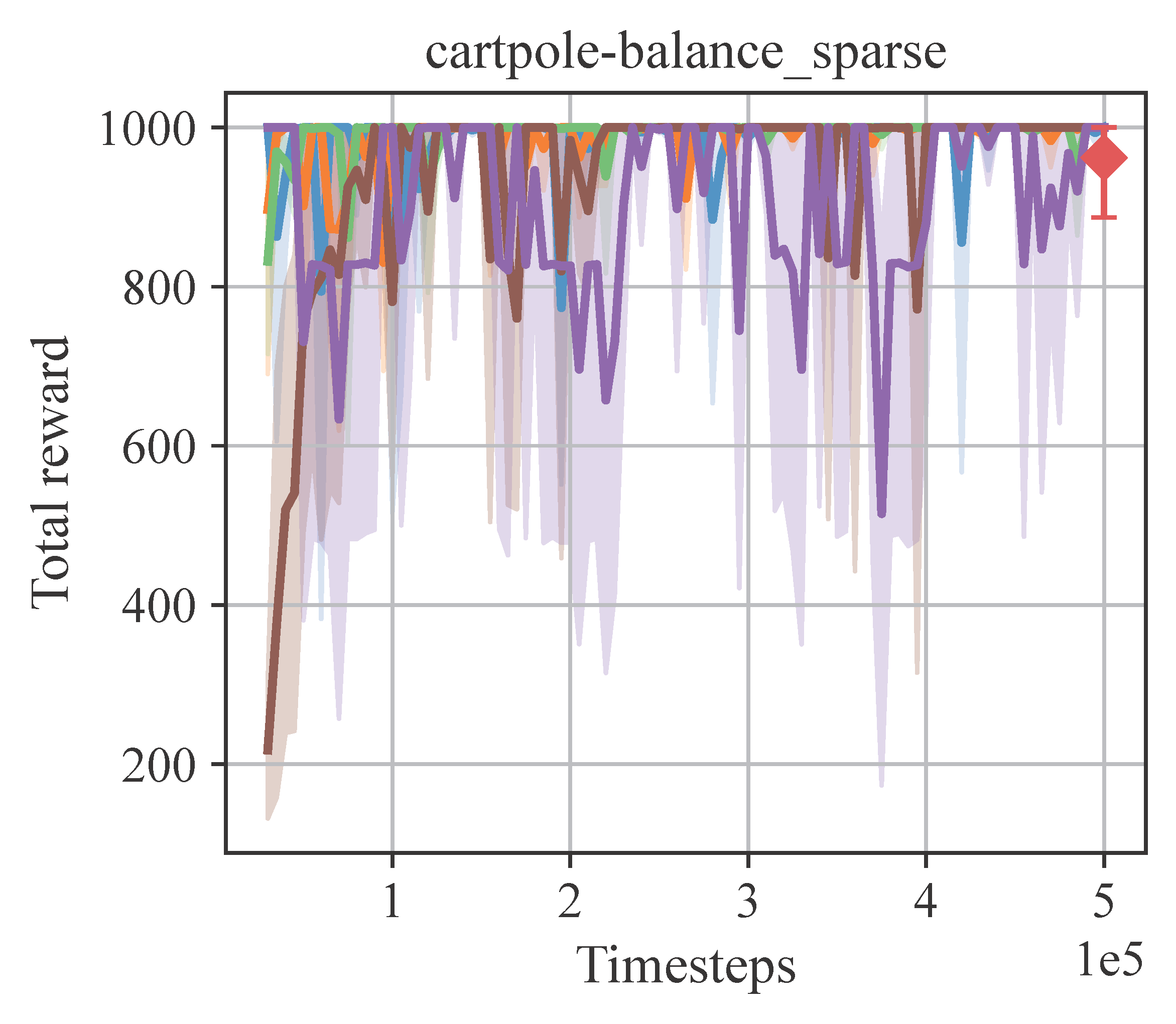}

\includegraphics[width=0.246\textwidth]{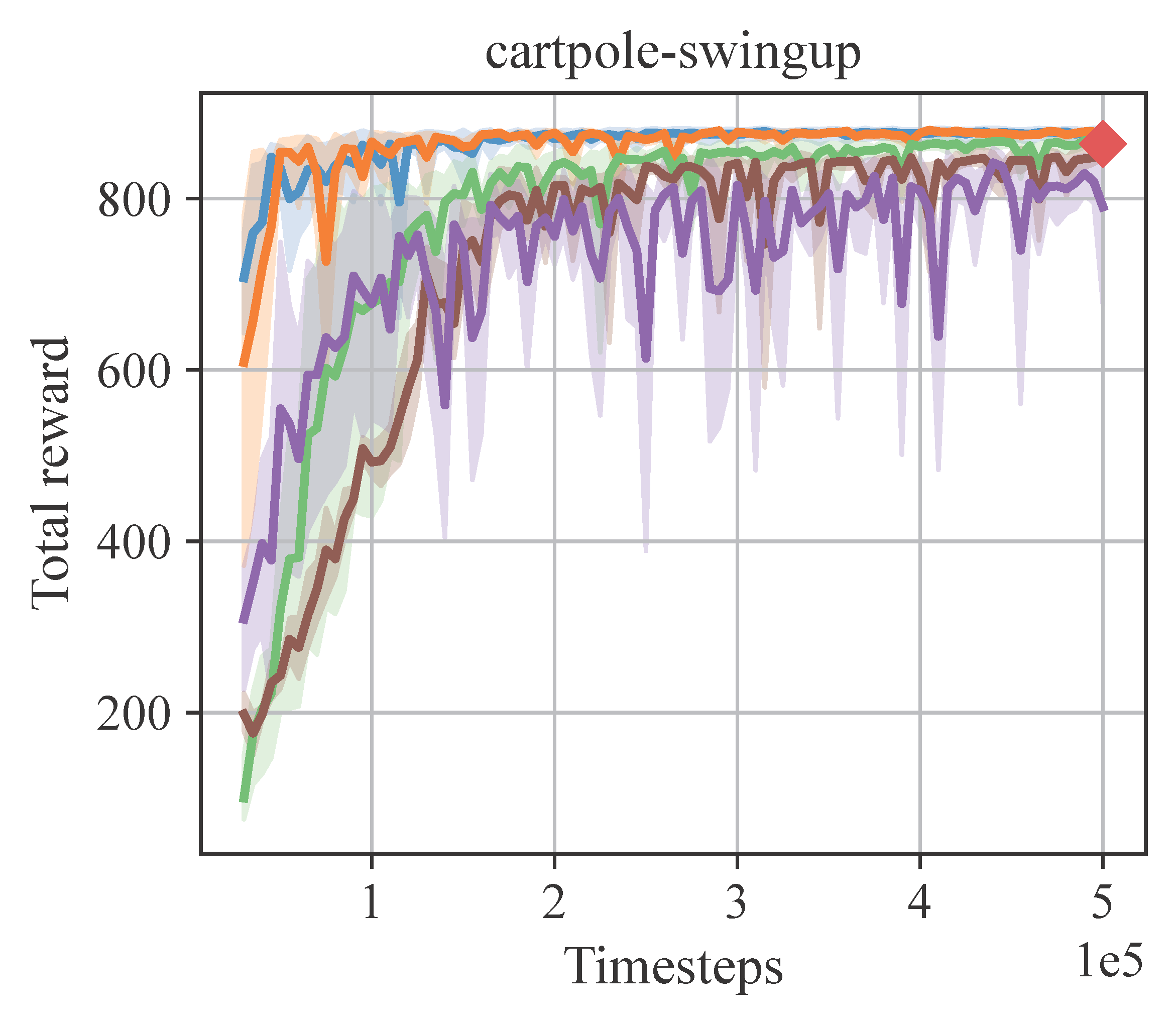}
\includegraphics[width=0.246\textwidth]{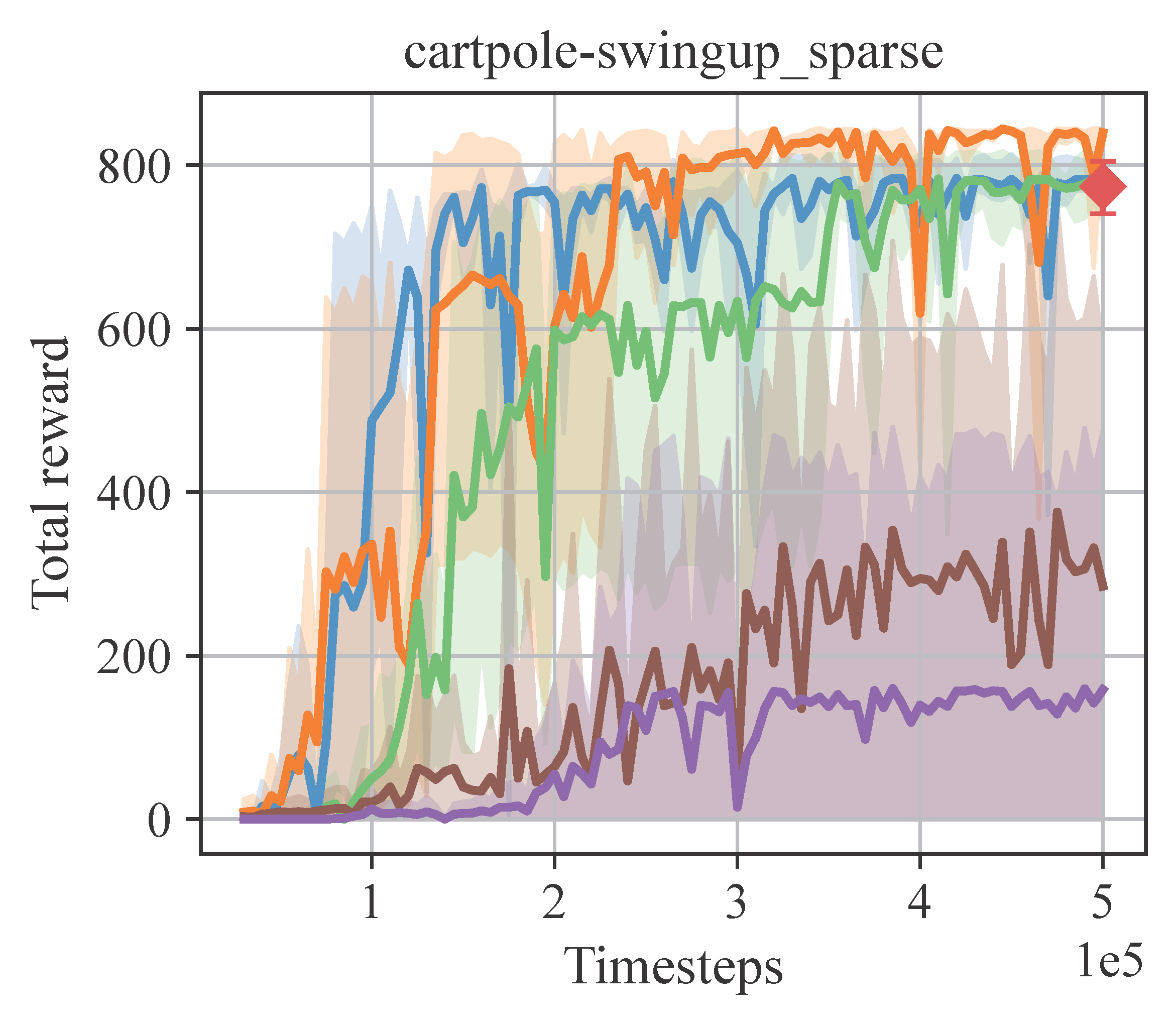}
\includegraphics[width=0.246\textwidth]{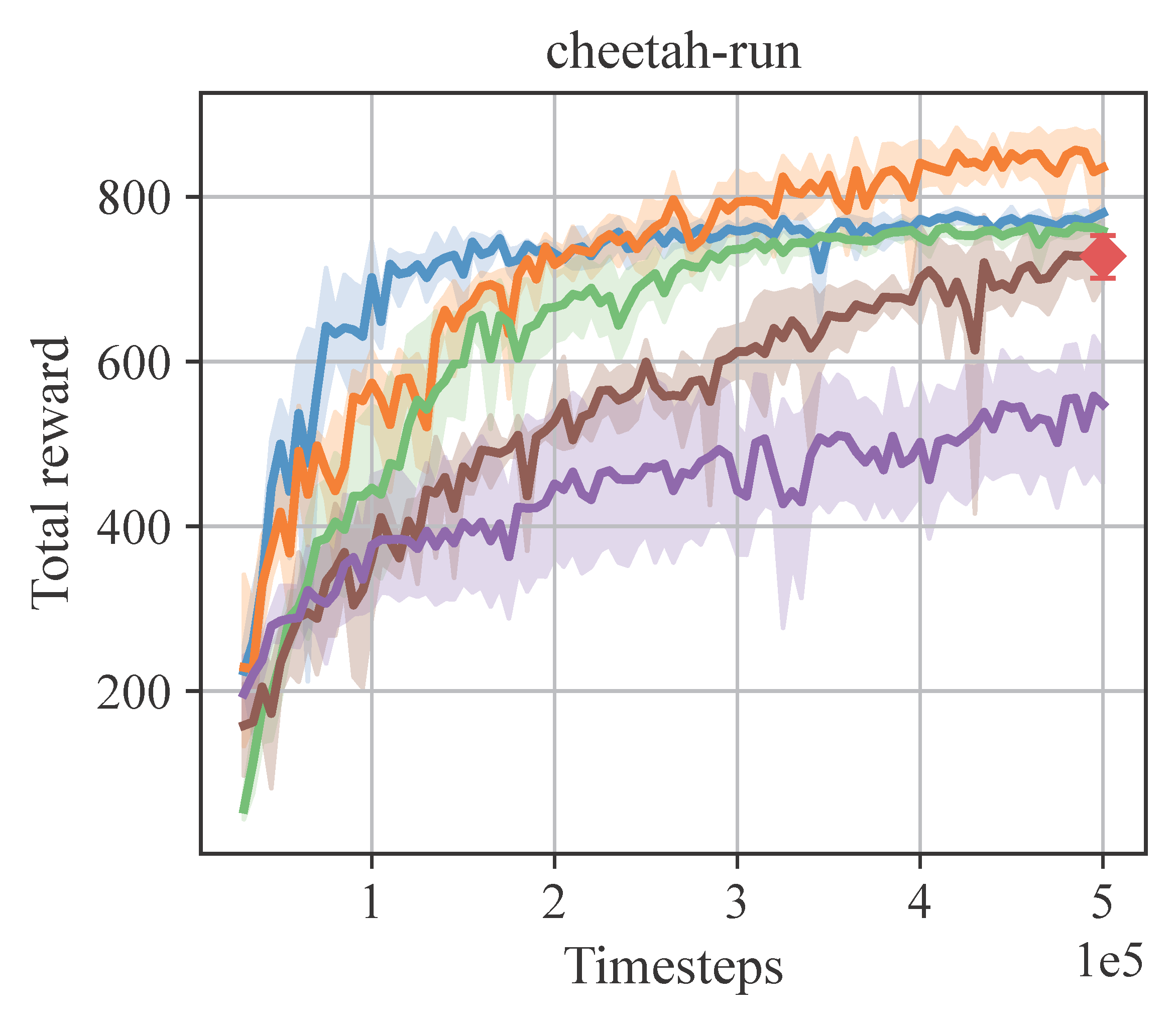}
\includegraphics[width=0.246\textwidth]{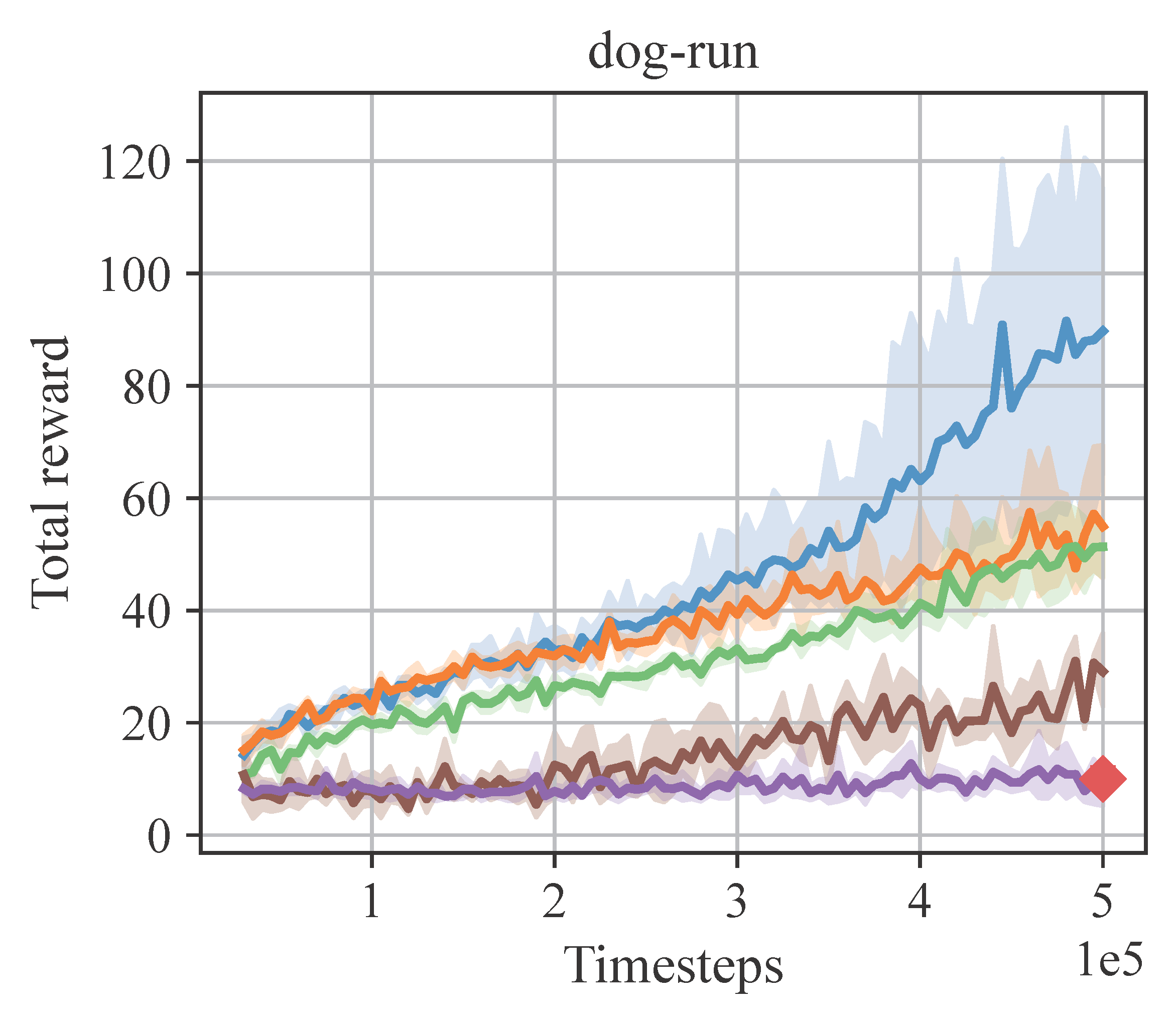}

\includegraphics[width=0.246\textwidth]{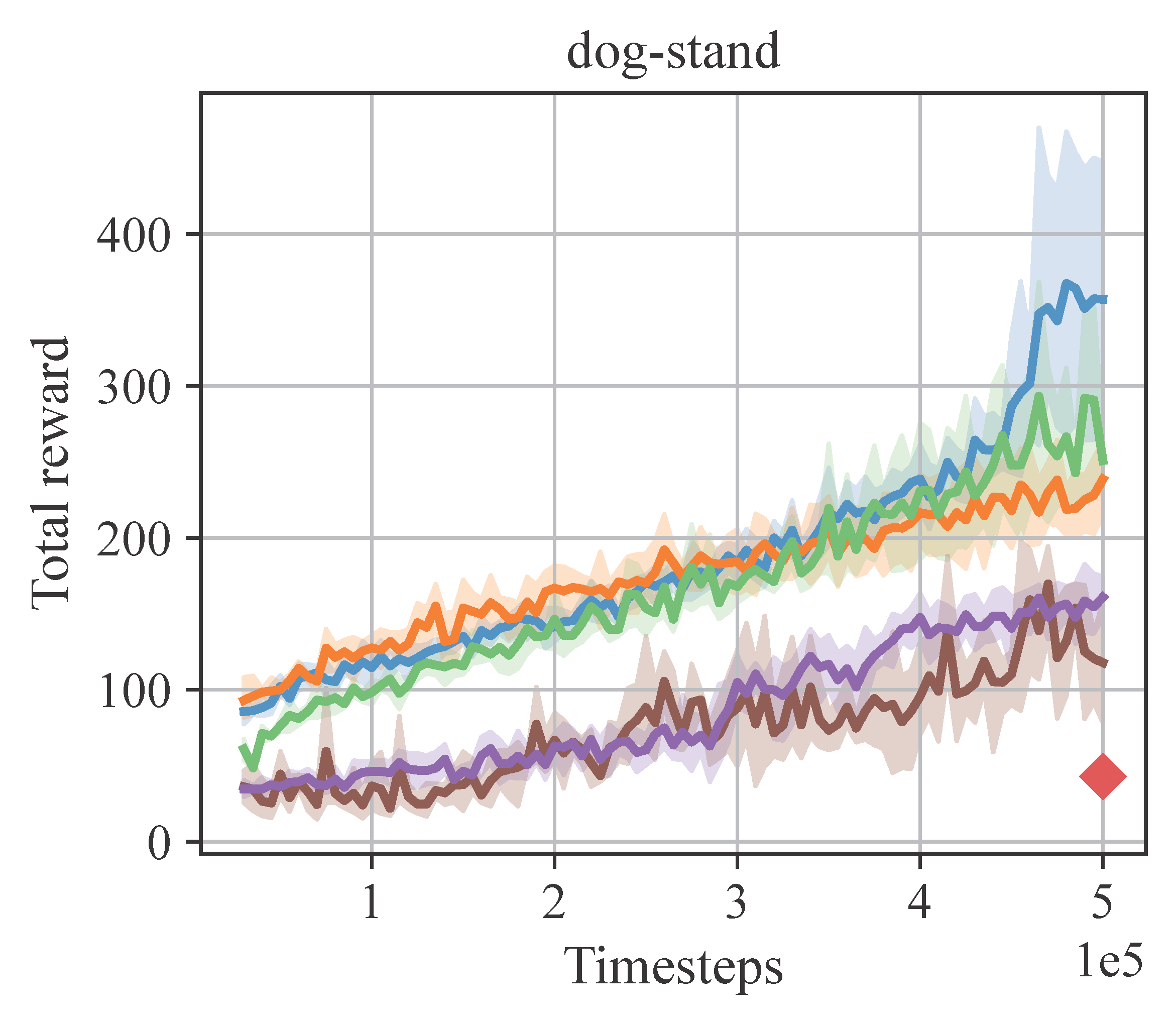}
\includegraphics[width=0.246\textwidth]{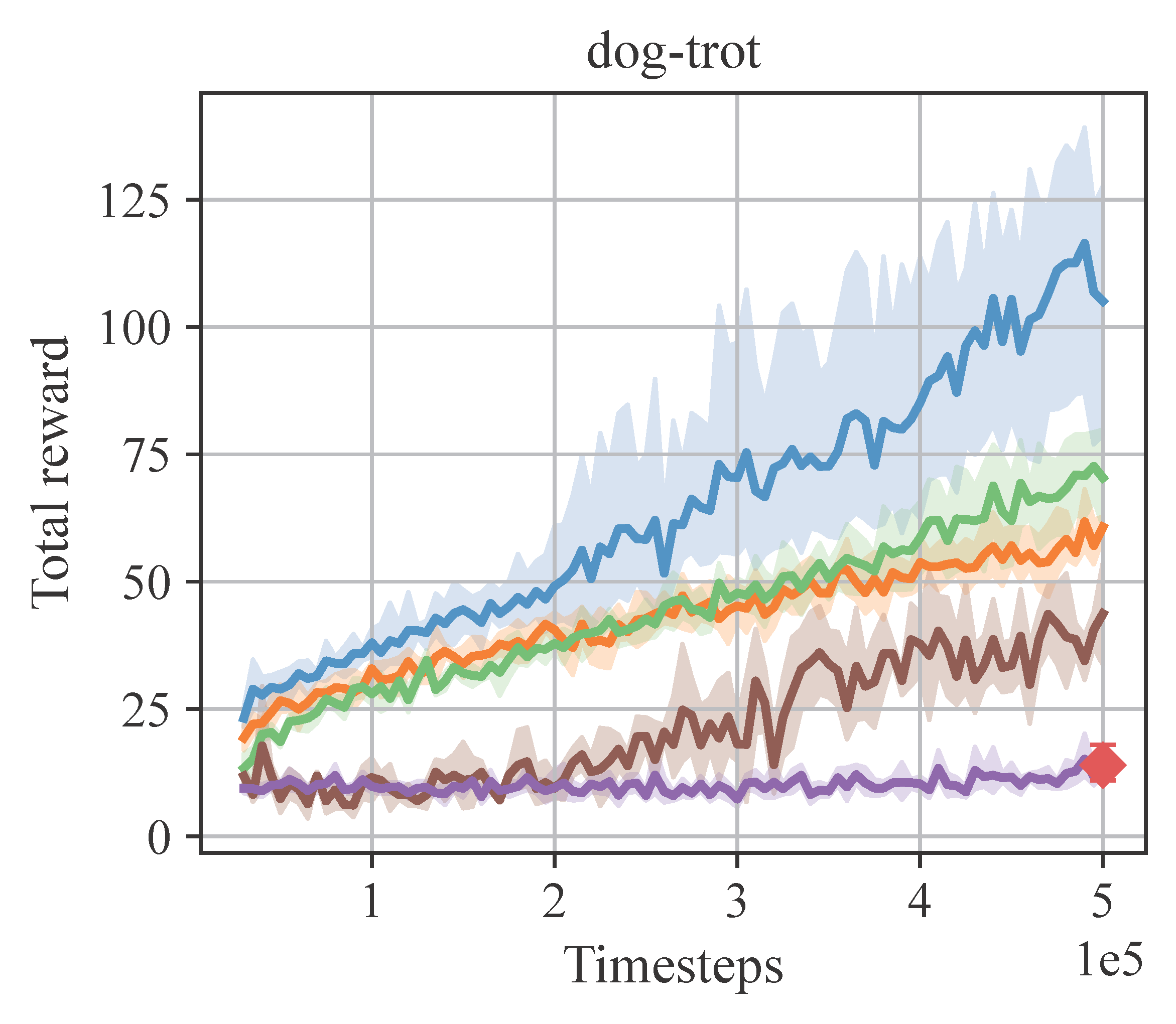}
\includegraphics[width=0.246\textwidth]{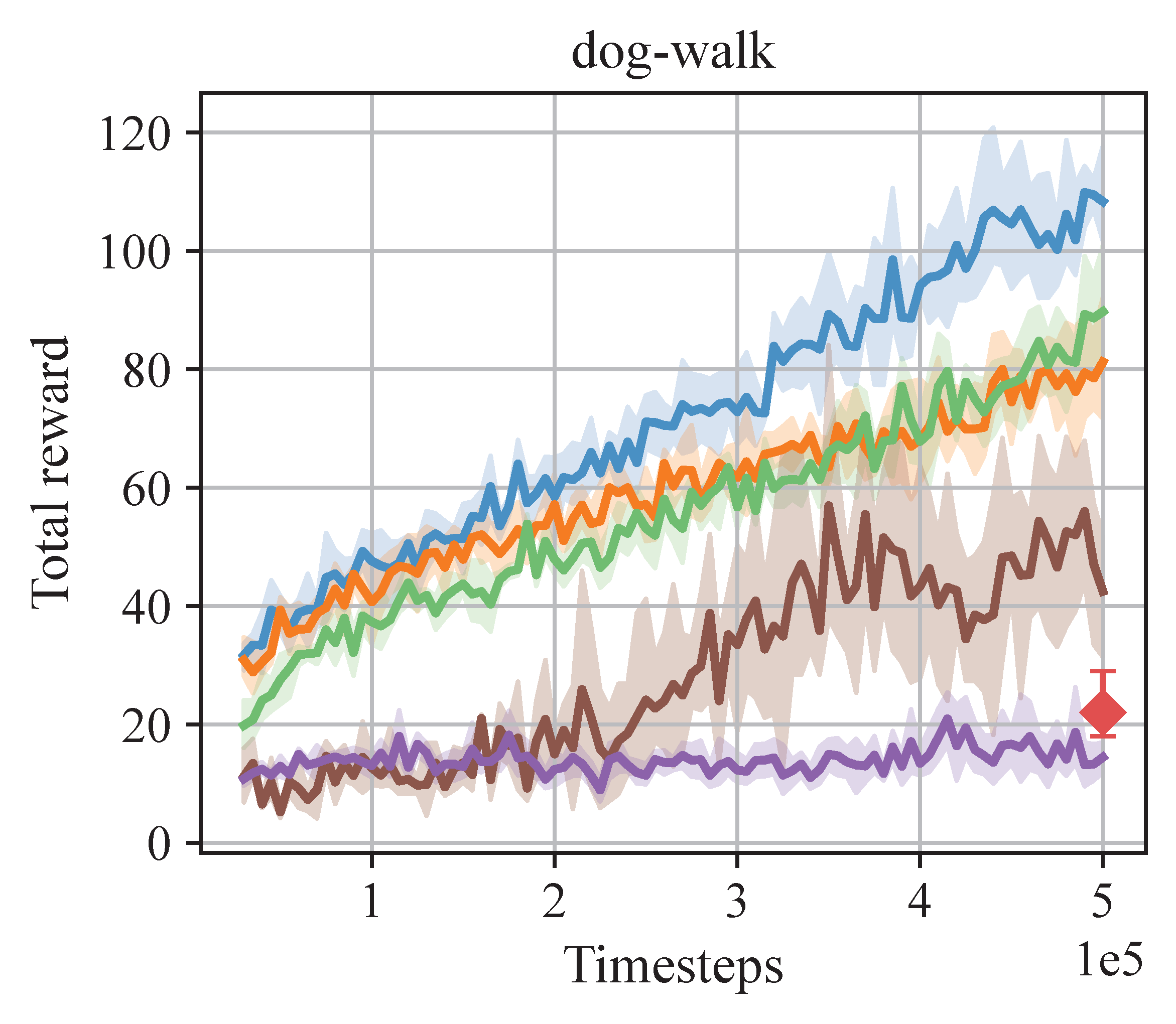}
\includegraphics[width=0.246\textwidth]{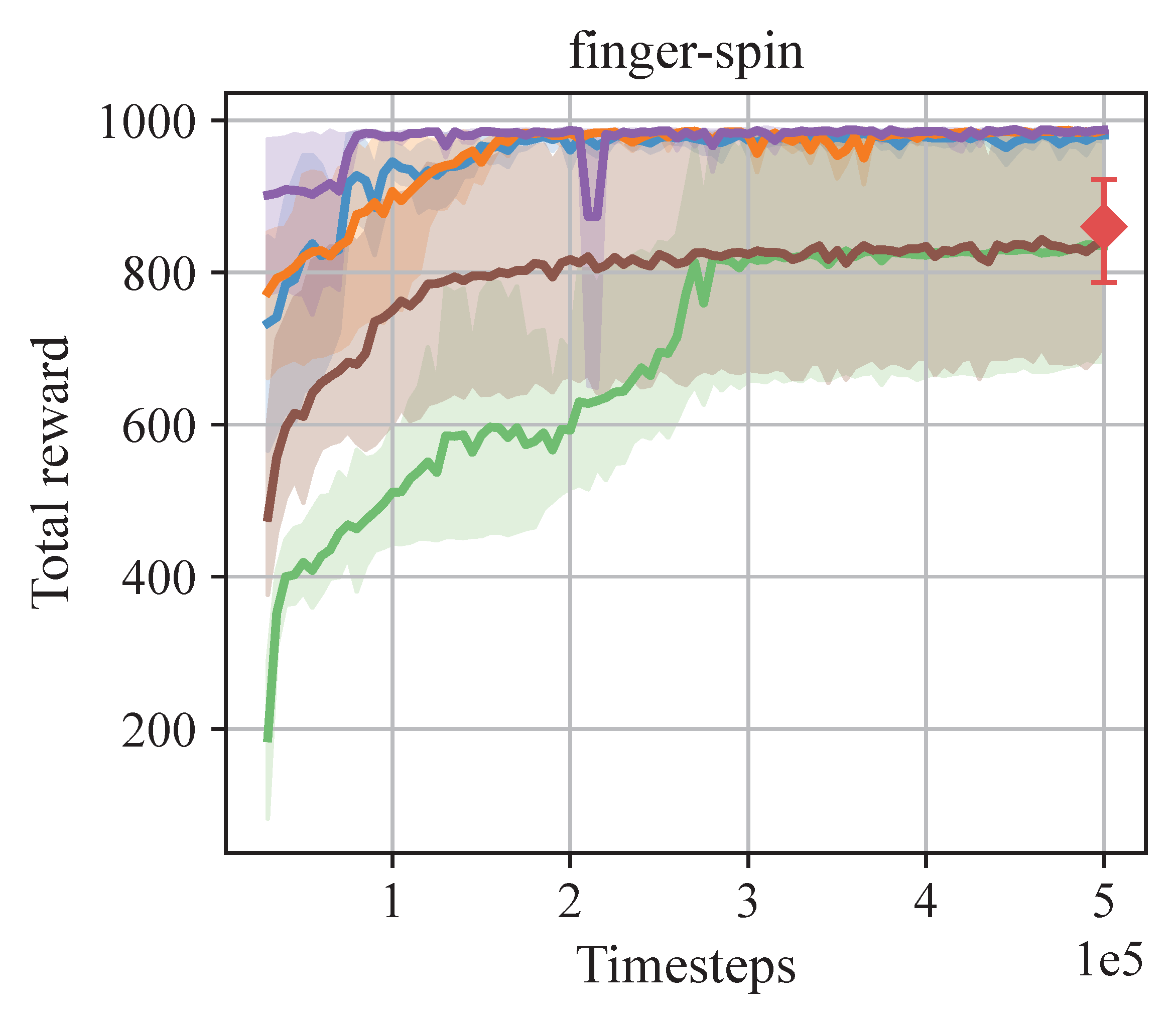}

\includegraphics[width=0.246\textwidth]{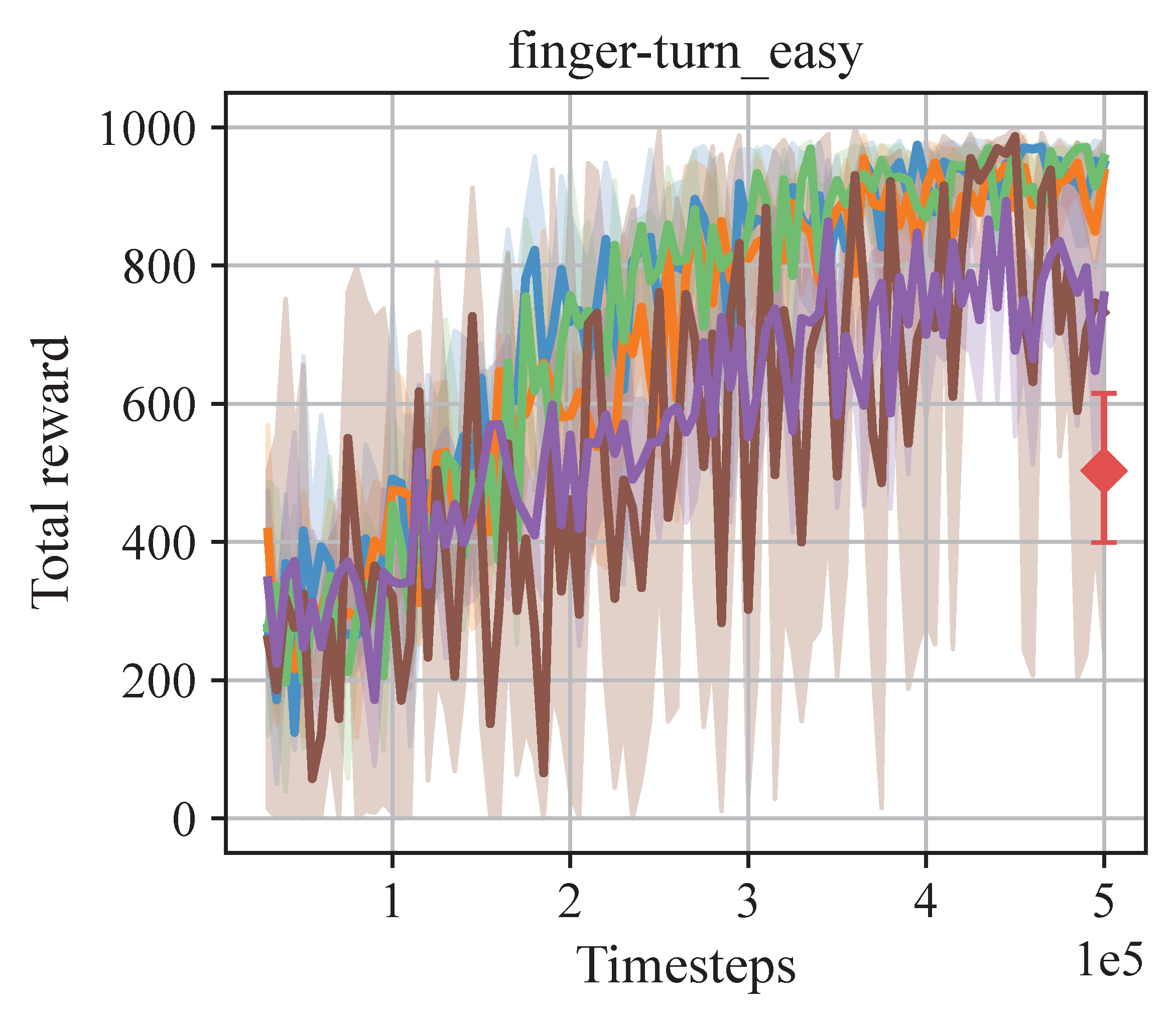}
\includegraphics[width=0.246\textwidth]{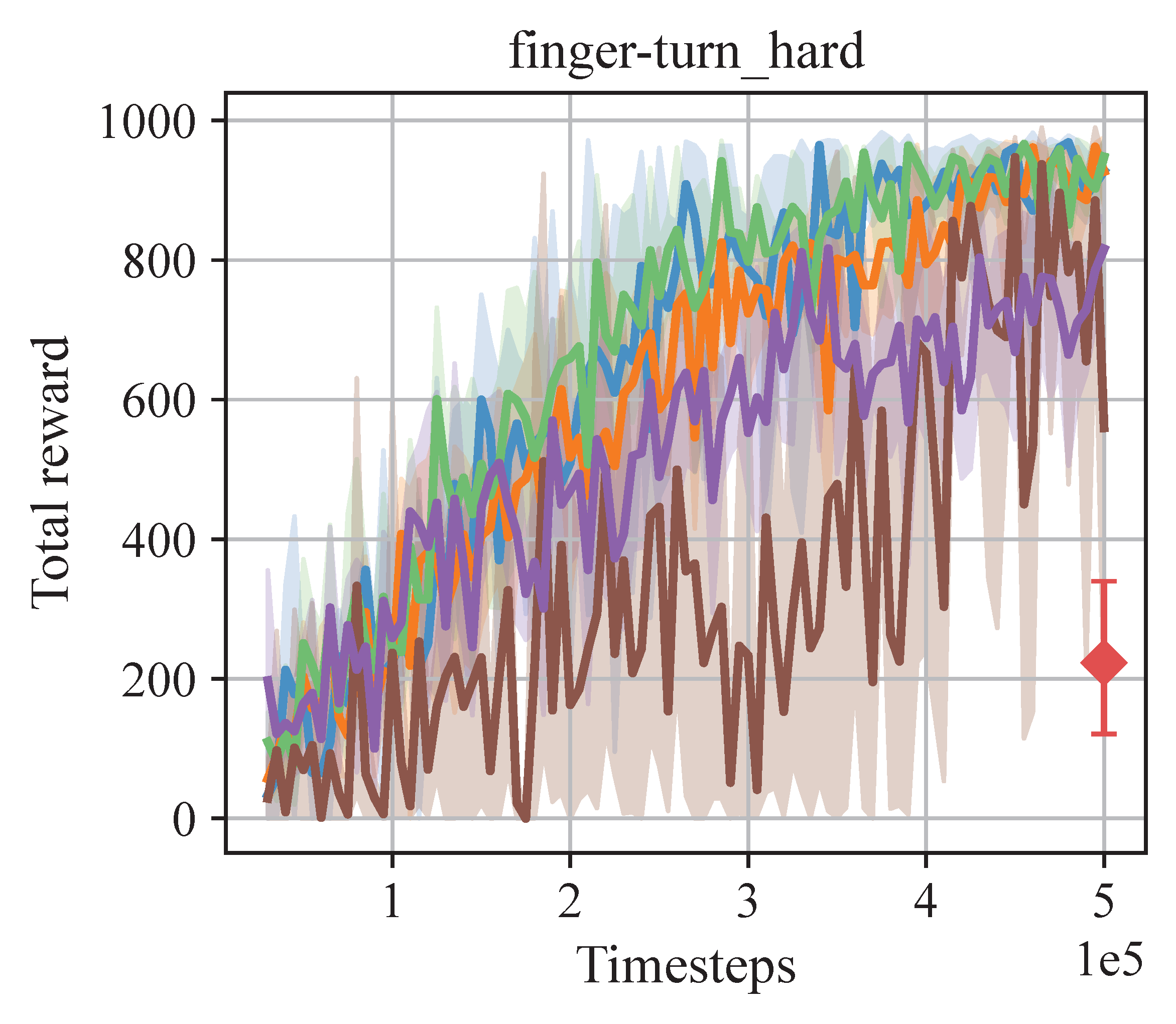}
\includegraphics[width=0.246\textwidth]{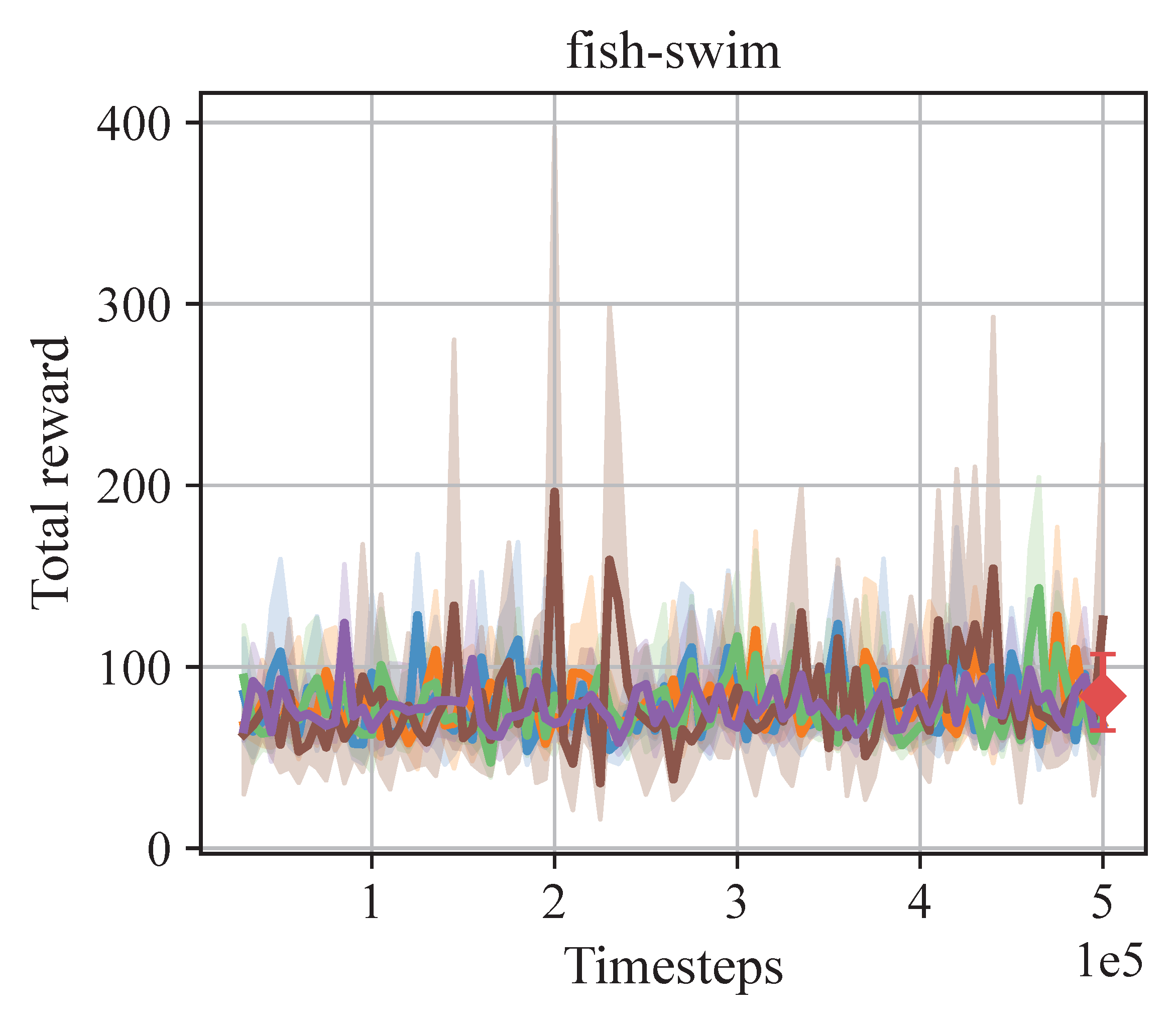}
\includegraphics[width=0.246\textwidth]{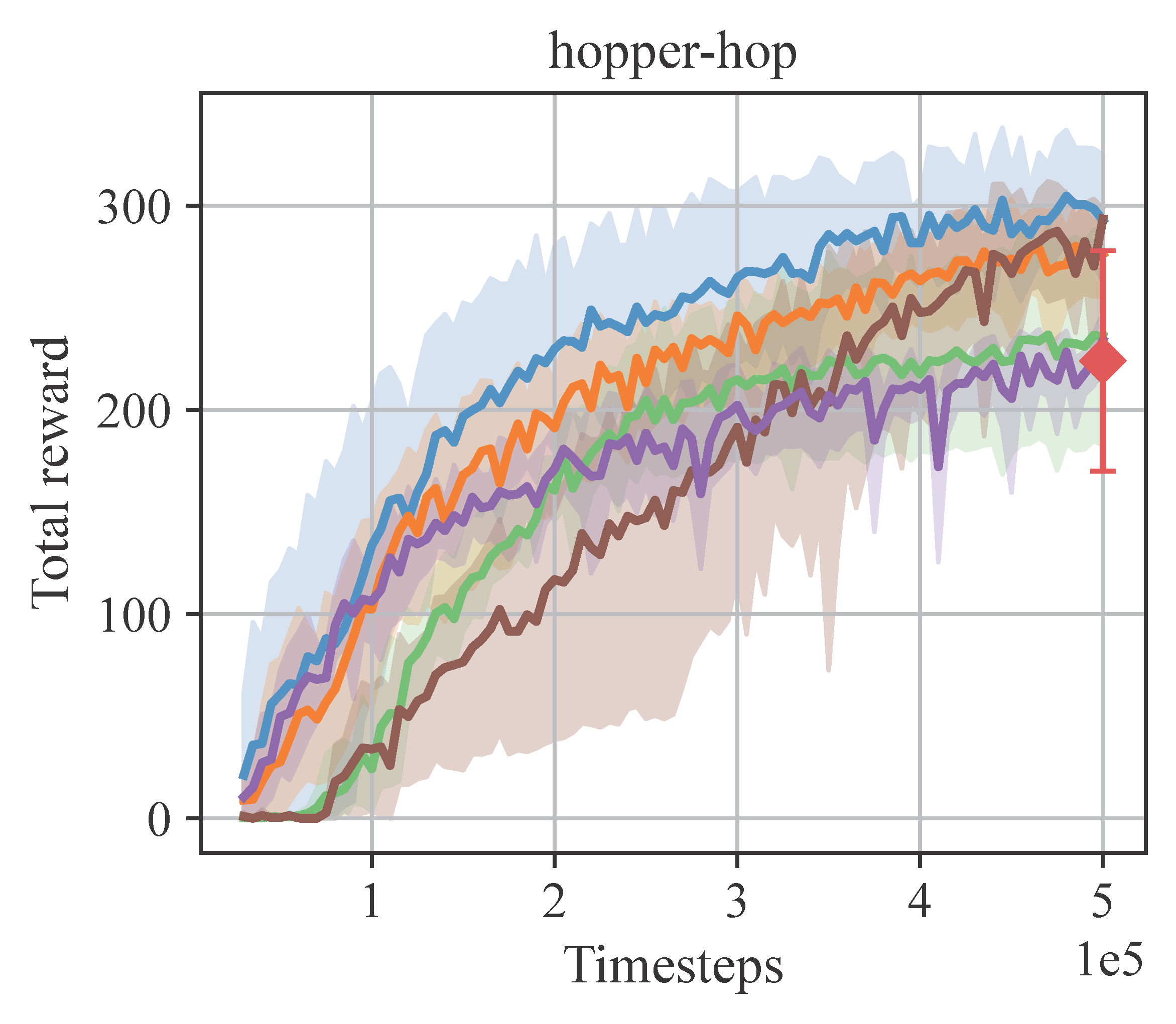}
\includegraphics[width=0.7\textwidth]{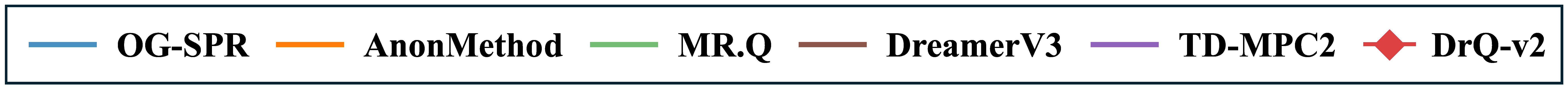}
\caption{
Per-task learning curves on the DMControl benchmark for OG-SPR and baselines (Part 1). Solid lines indicate average performance over 5 seeds, and shaded areas indicate the 95\% bootstrap confidence interval.
}
\label{per_task1}
\end{figure*}

\begin{figure*}[t]
\centering
\includegraphics[width=0.246\textwidth]{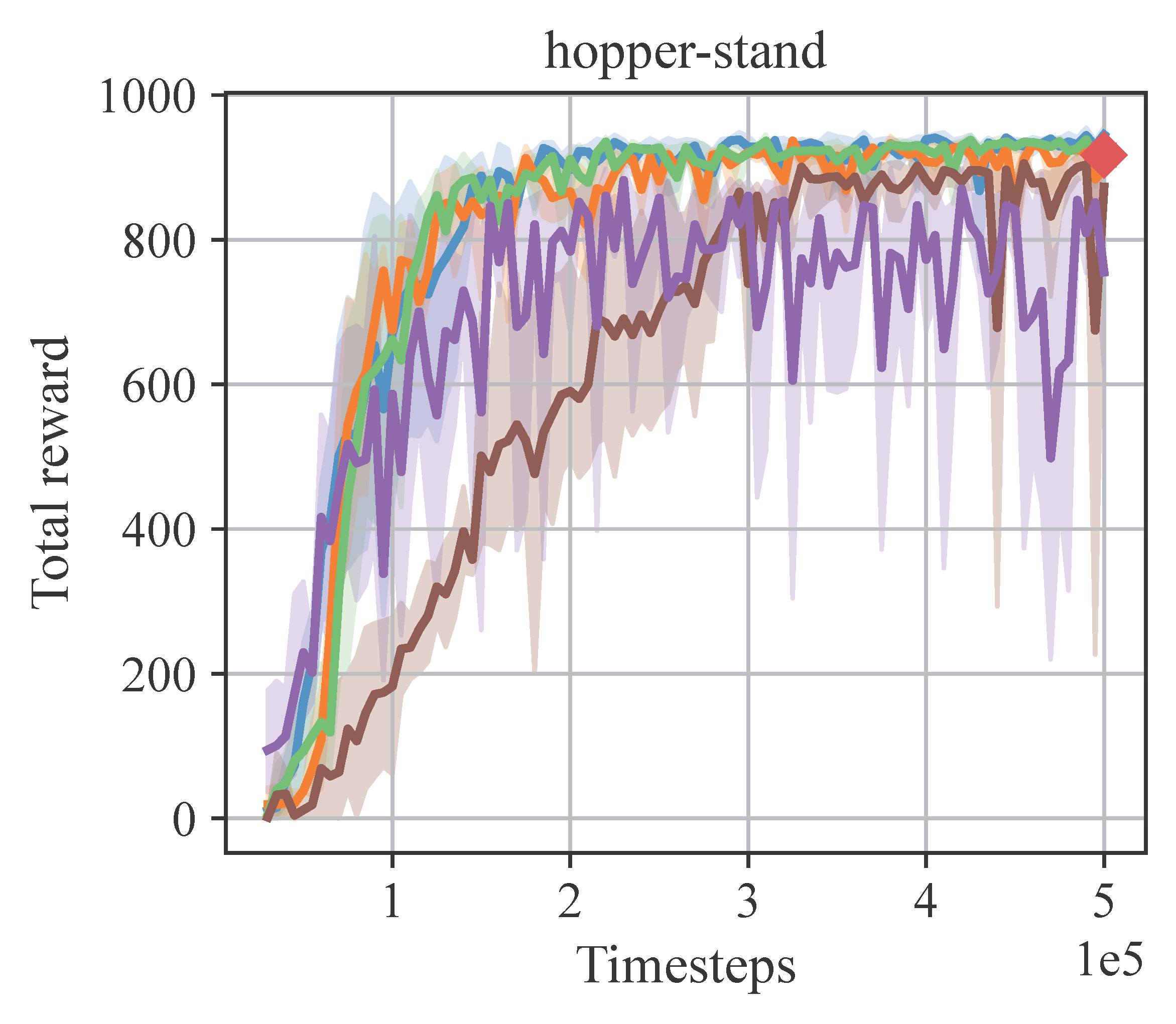}
\includegraphics[width=0.246\textwidth]{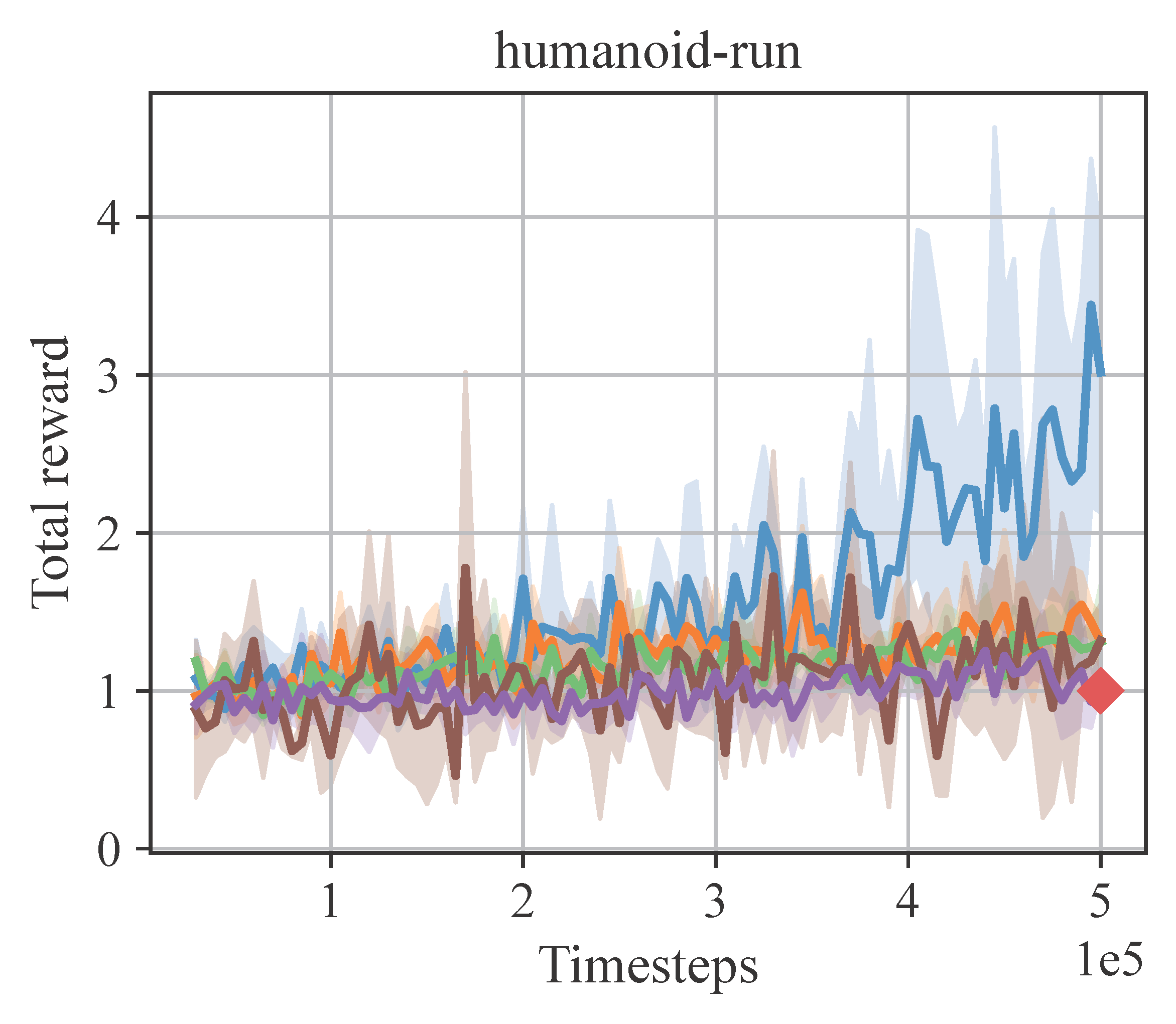}
\includegraphics[width=0.246\textwidth]{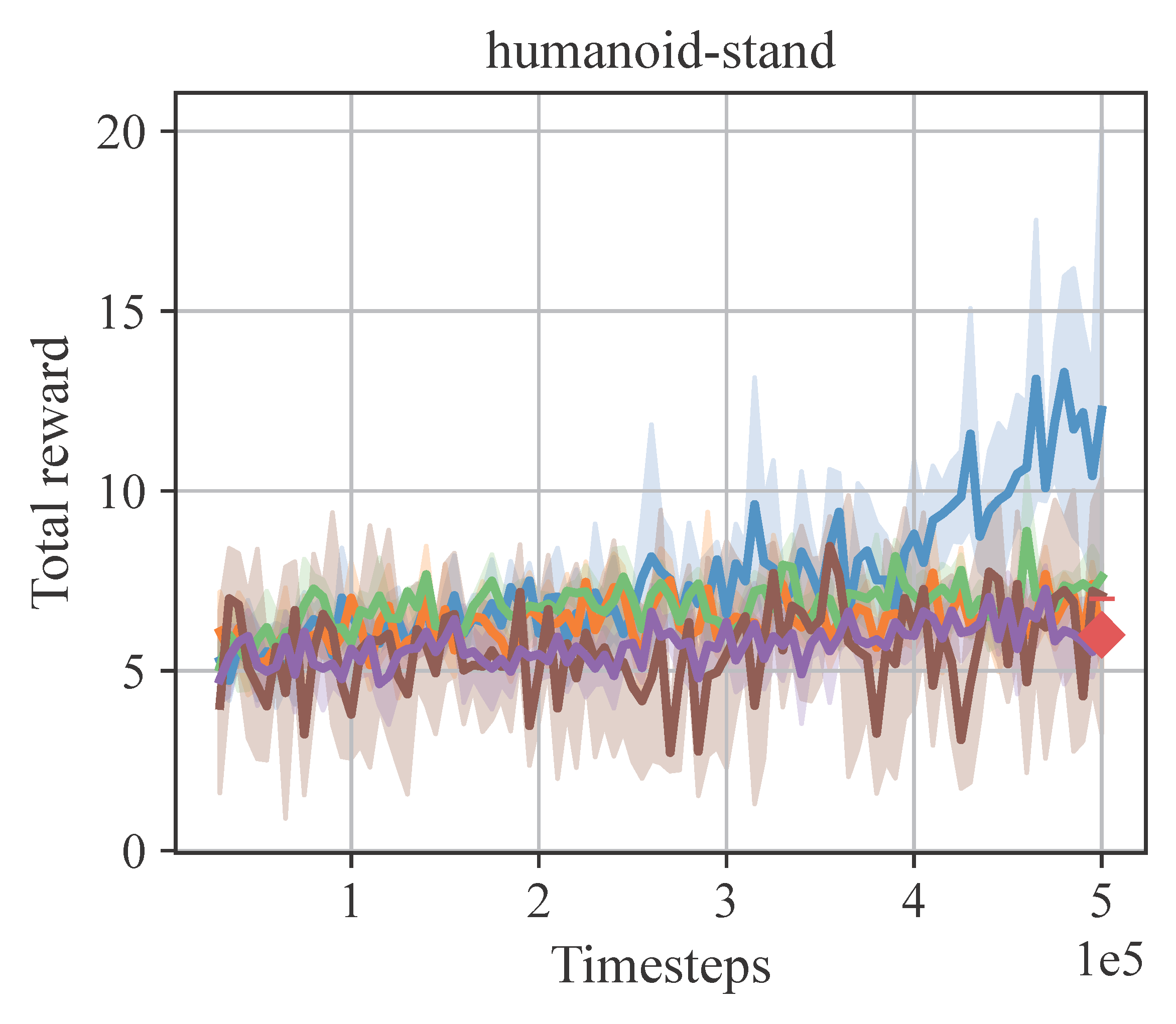}
\includegraphics[width=0.246\textwidth]{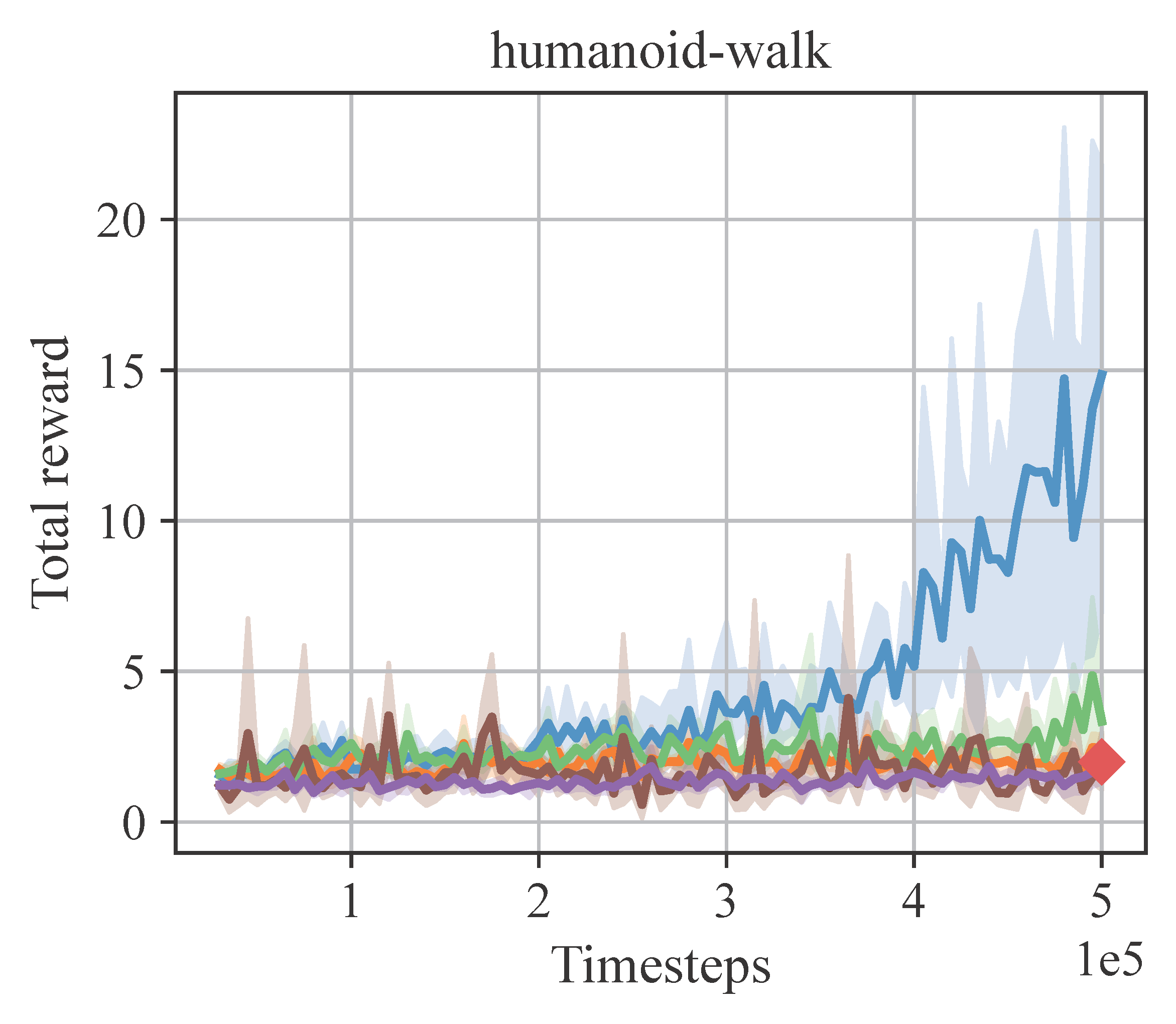}

\includegraphics[width=0.246\textwidth]{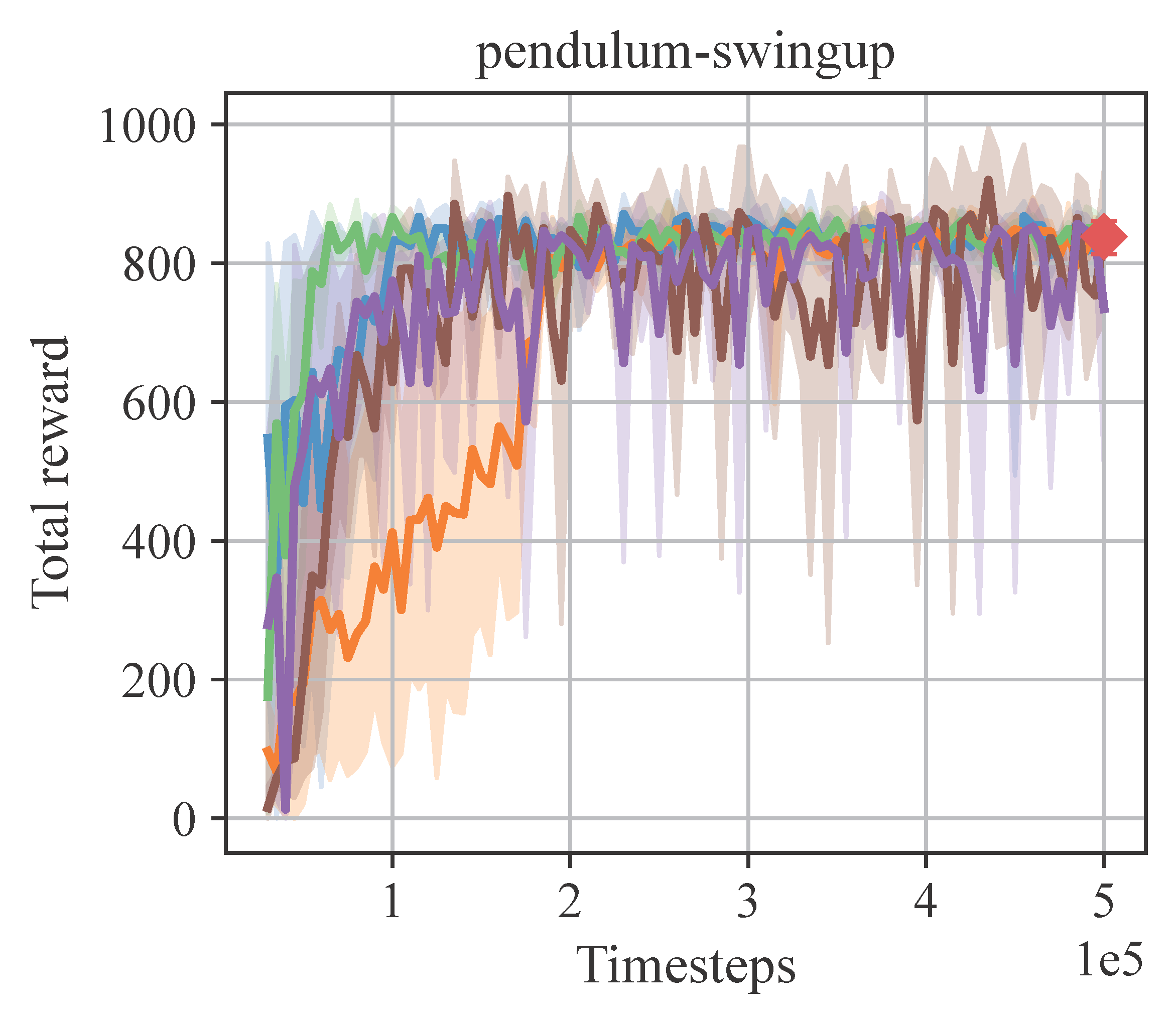}
\includegraphics[width=0.246\textwidth]{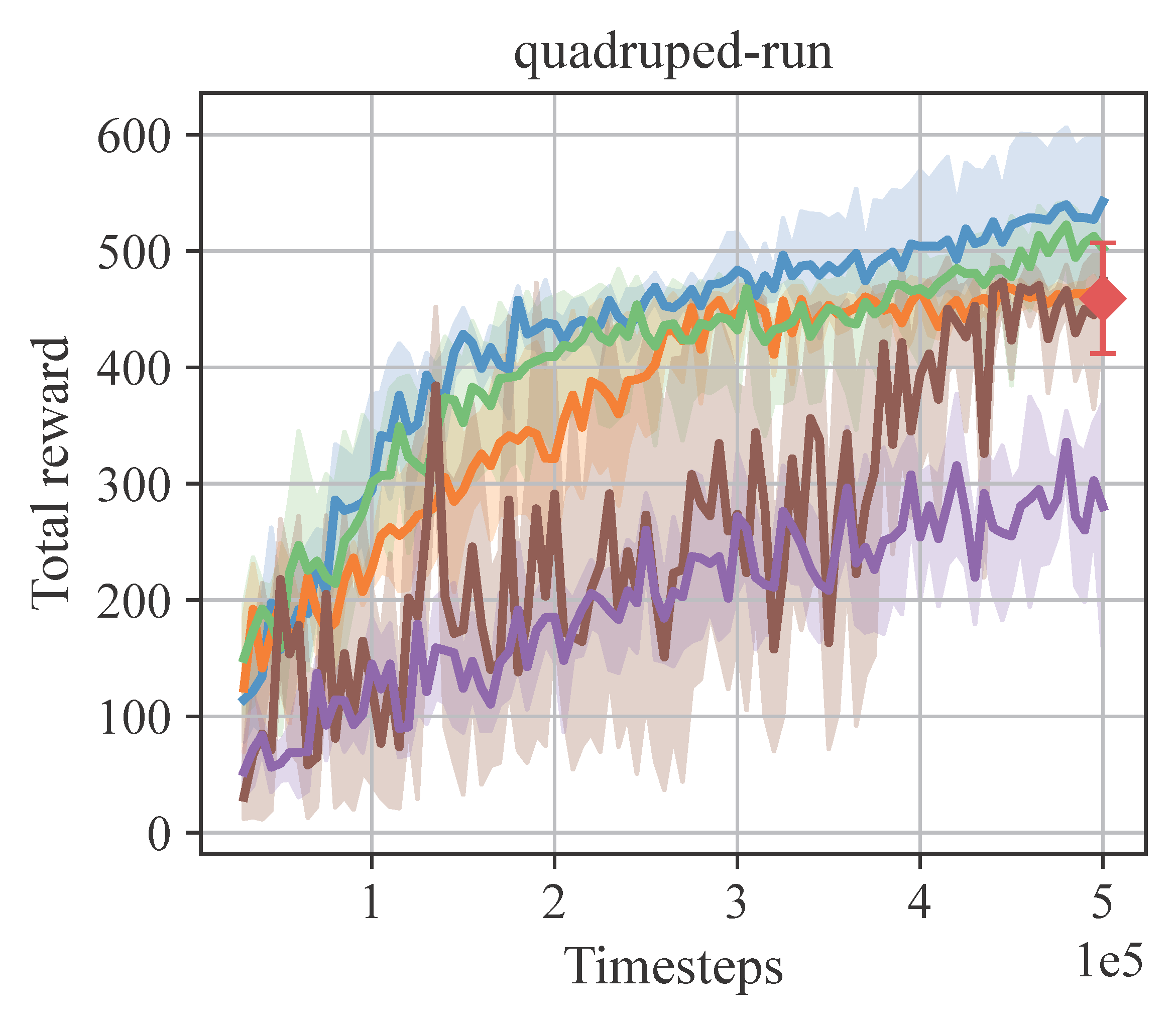}
\includegraphics[width=0.246\textwidth]{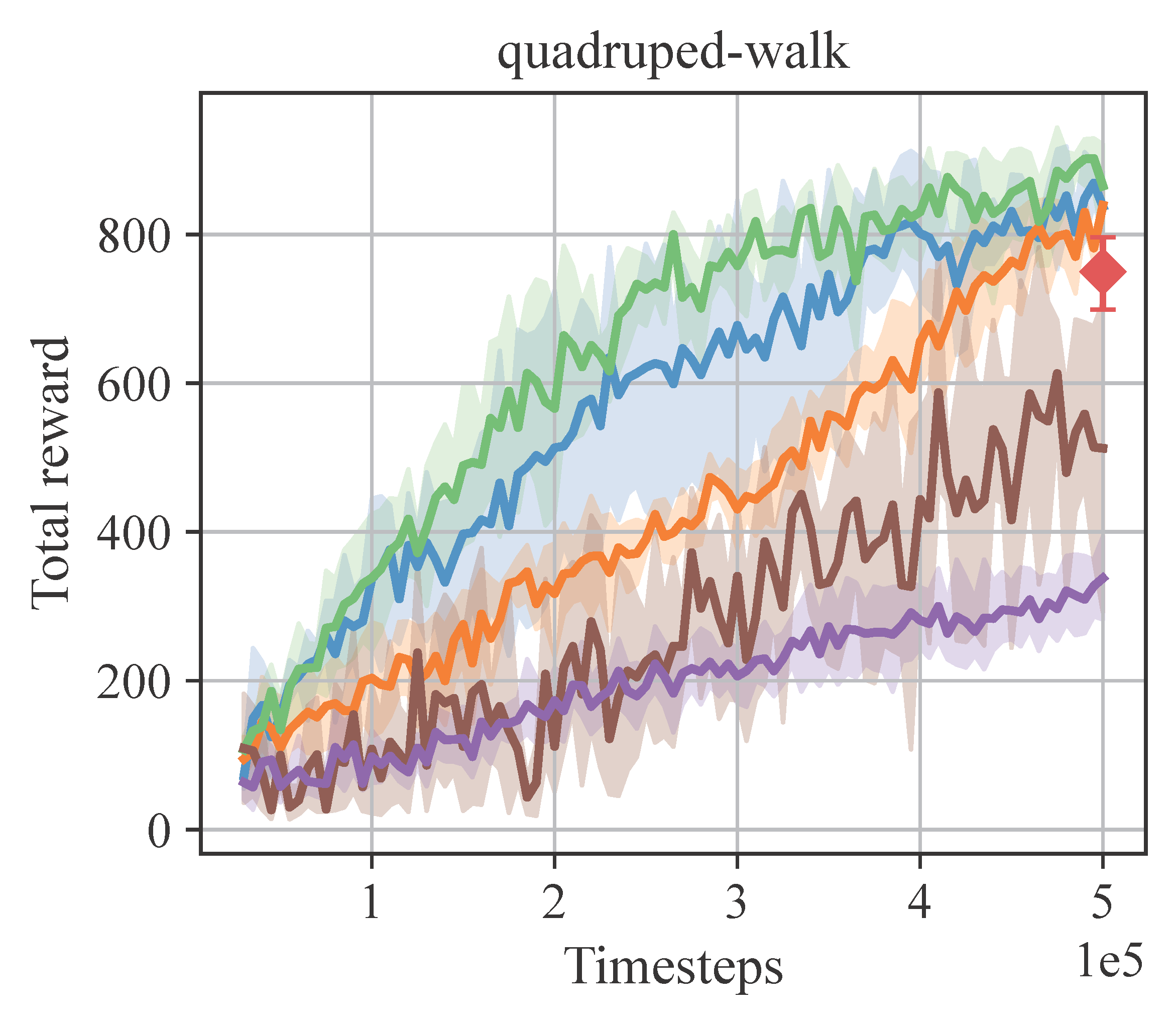}
\includegraphics[width=0.246\textwidth]{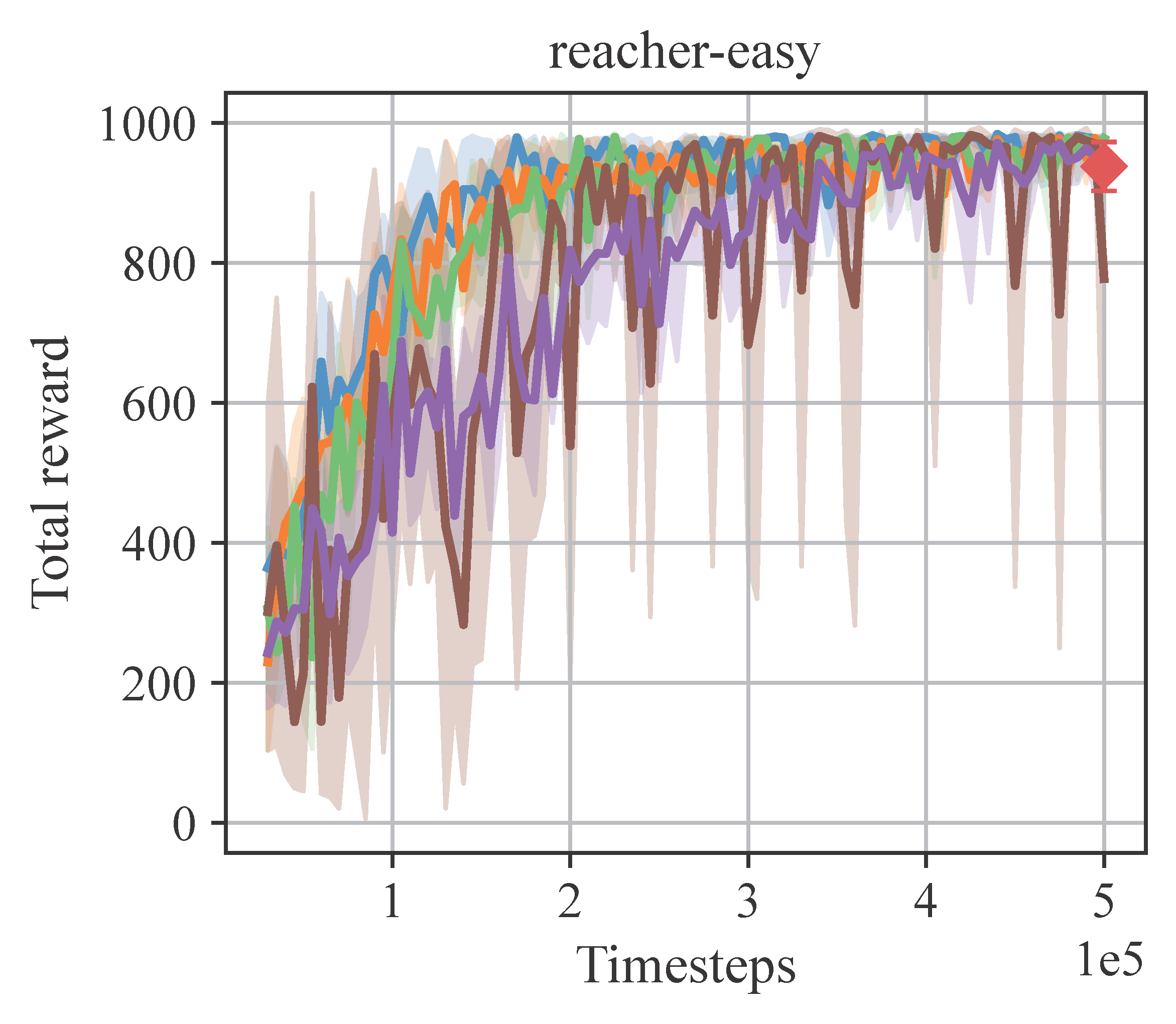}

\includegraphics[width=0.246\textwidth]{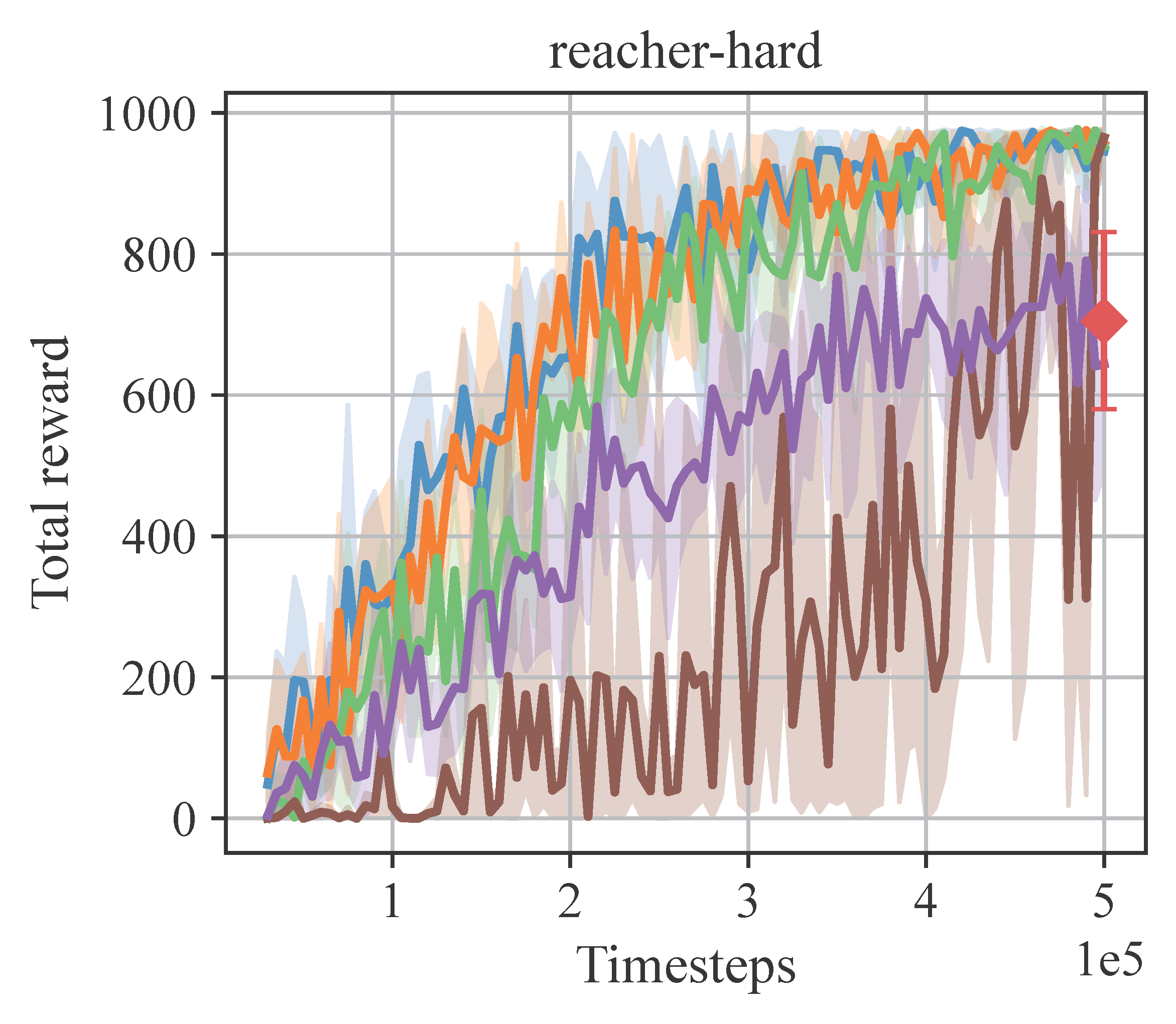}
\includegraphics[width=0.246\textwidth]{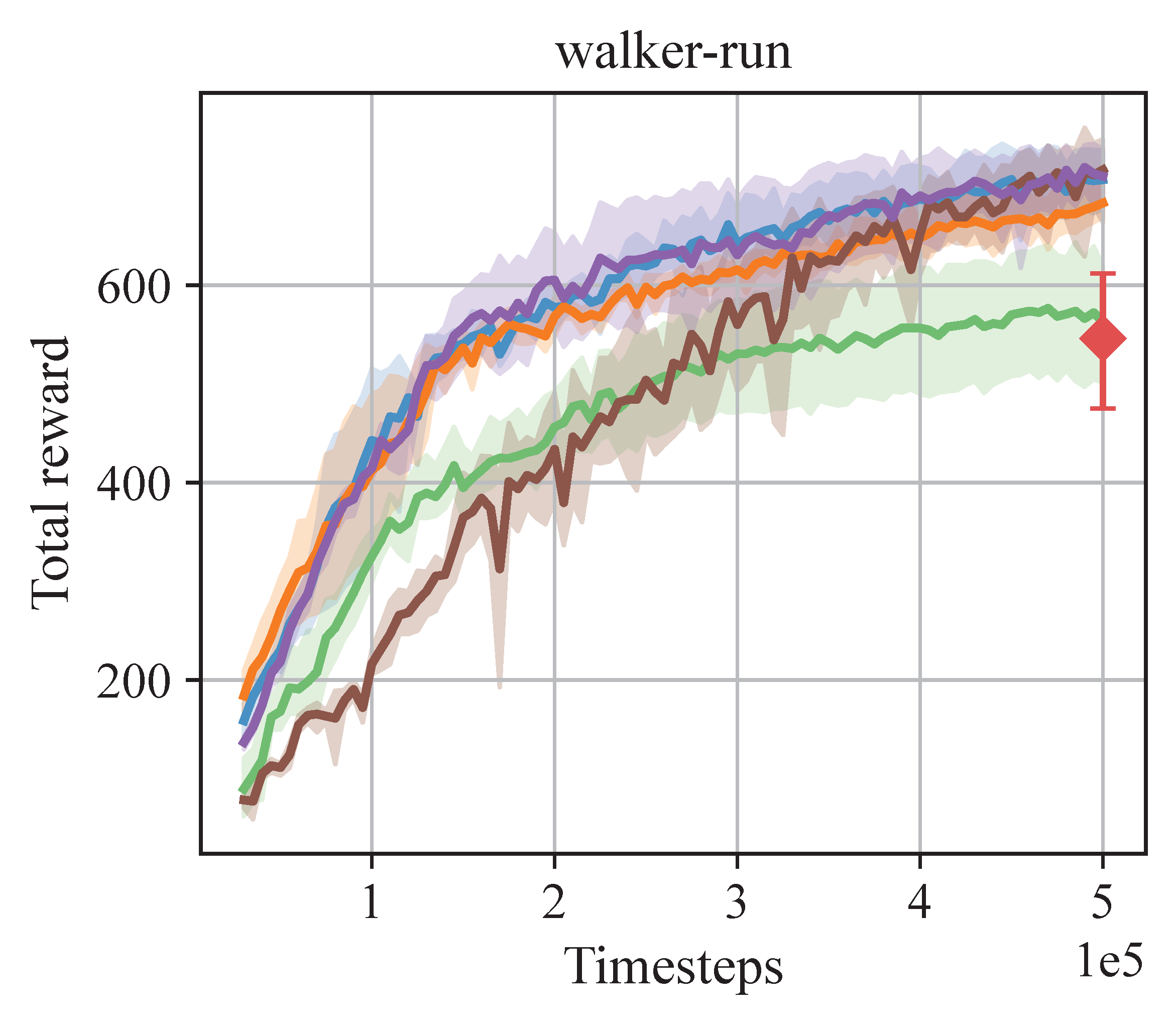}
\includegraphics[width=0.246\textwidth]{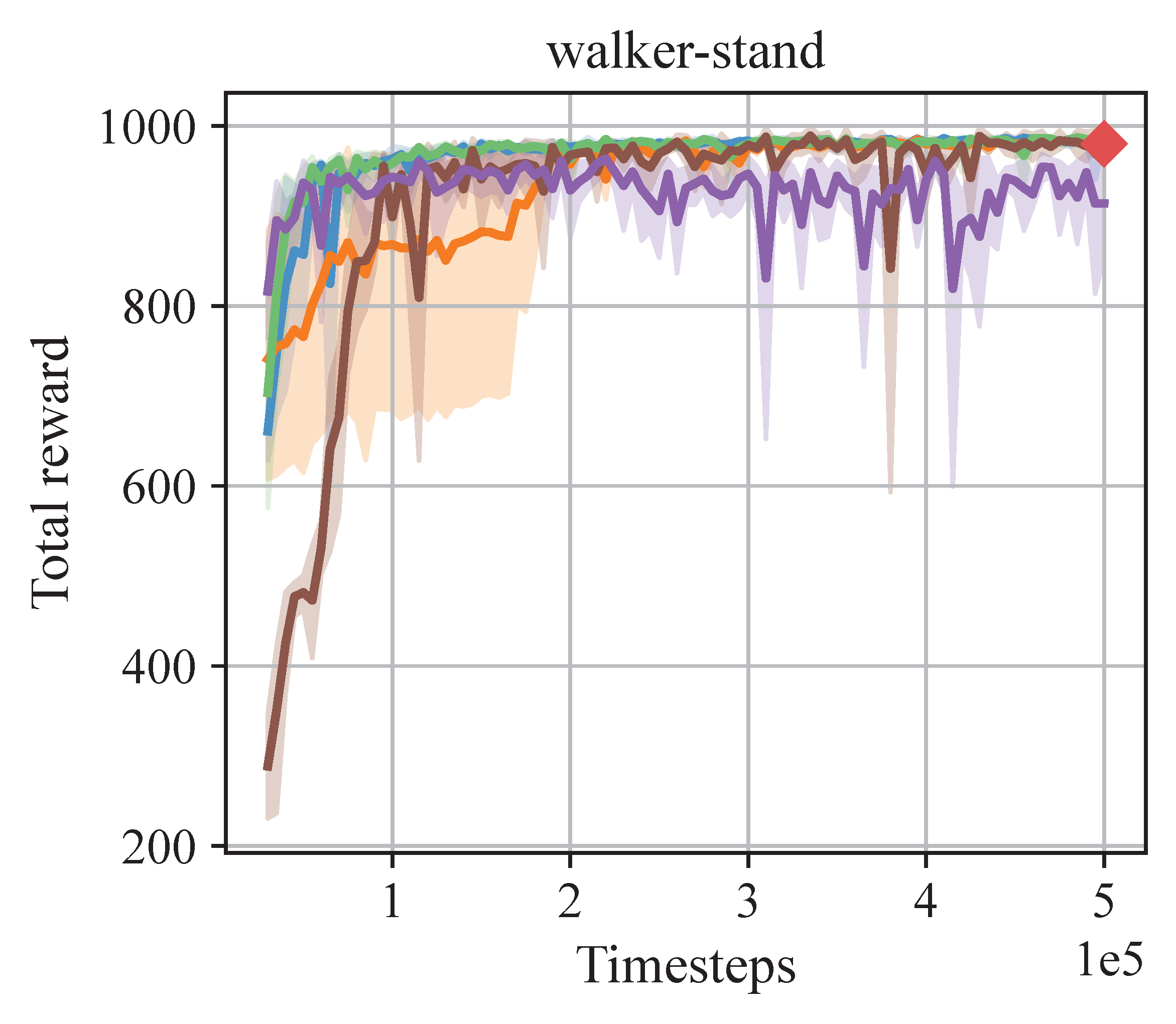}
\includegraphics[width=0.246\textwidth]{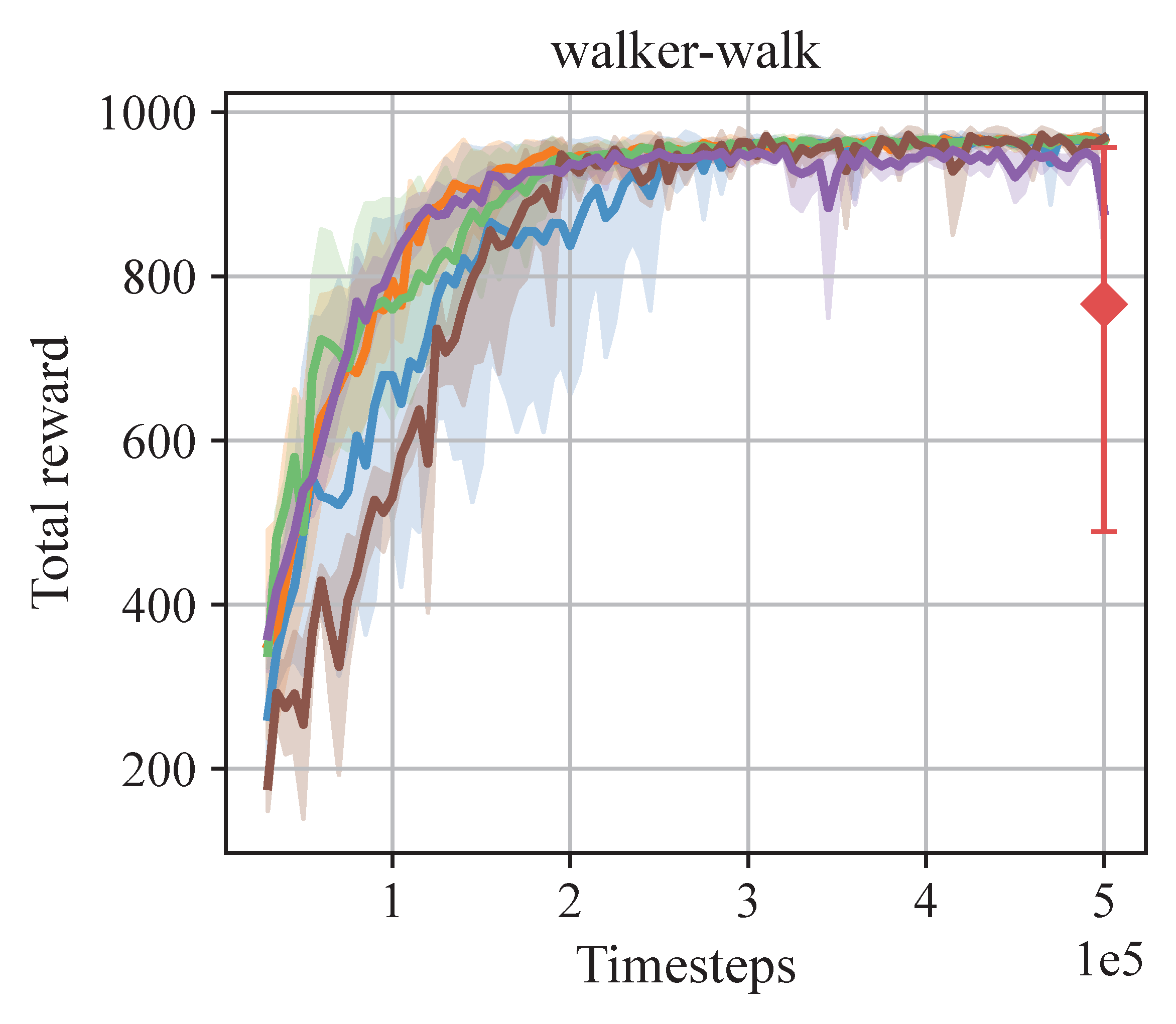}

\includegraphics[width=0.7\textwidth]{Per_tasks/bars.png}
\caption{
Per-task learning curves on the DMControl benchmark for OG-SPR and baselines (Part 2). Solid lines indicate average performance over 5 seeds, and shaded areas indicate the 95\% bootstrap confidence interval.
}
\label{per_task2}
\end{figure*}

\begin{figure*}[t]
\centering
\includegraphics[width=0.33\textwidth]{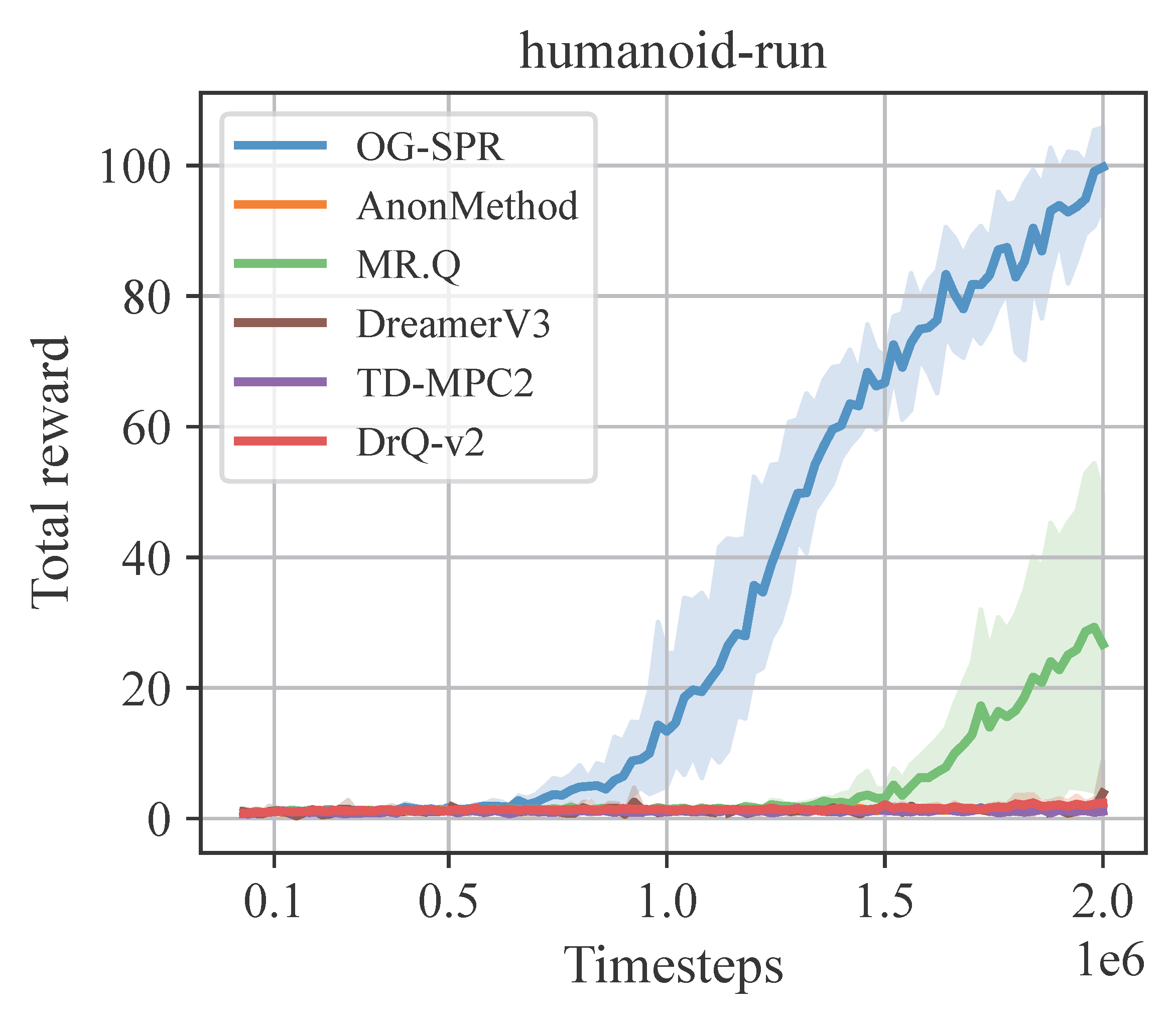}
\includegraphics[width=0.33\textwidth]{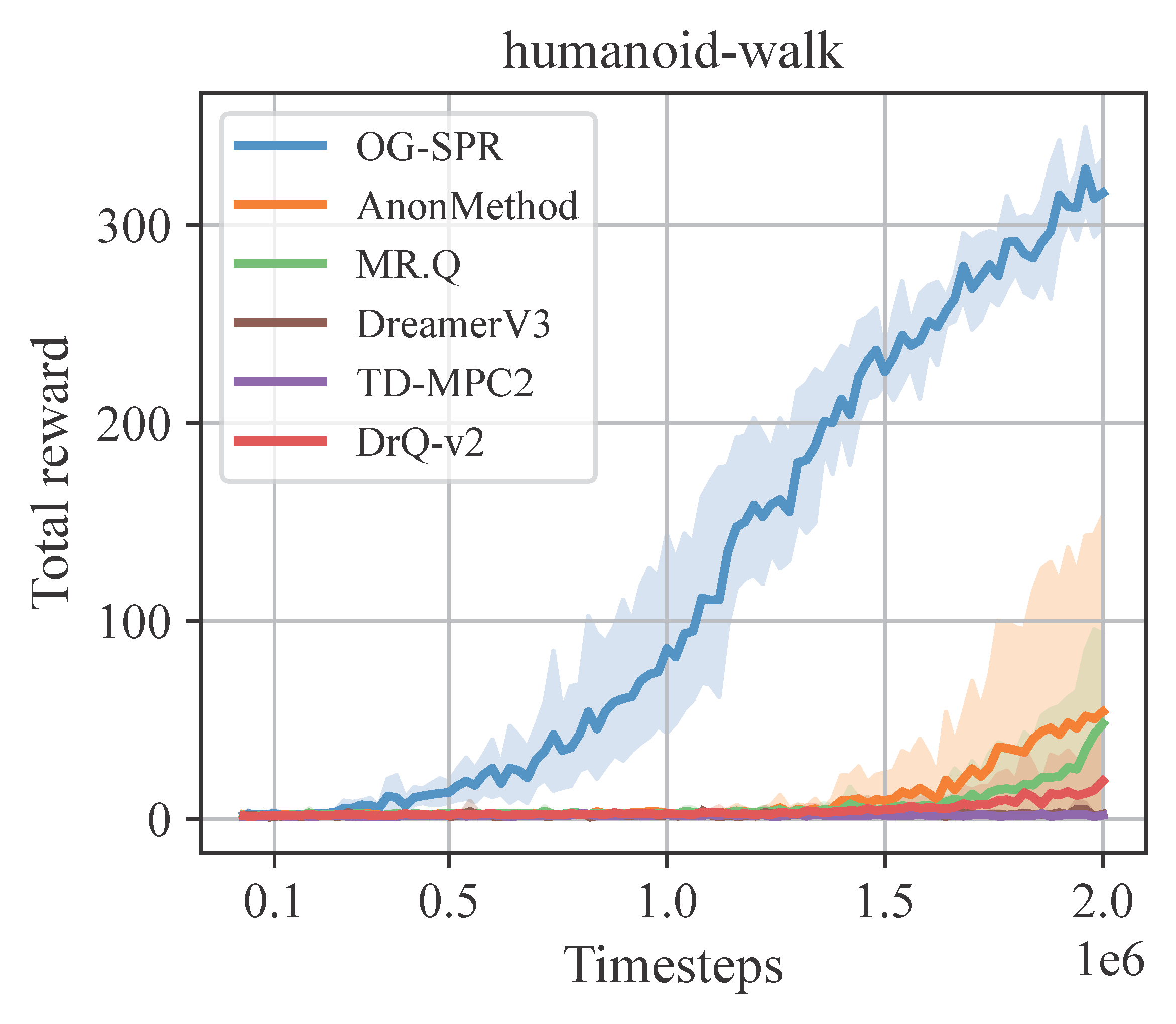}
\includegraphics[width=0.33\textwidth]{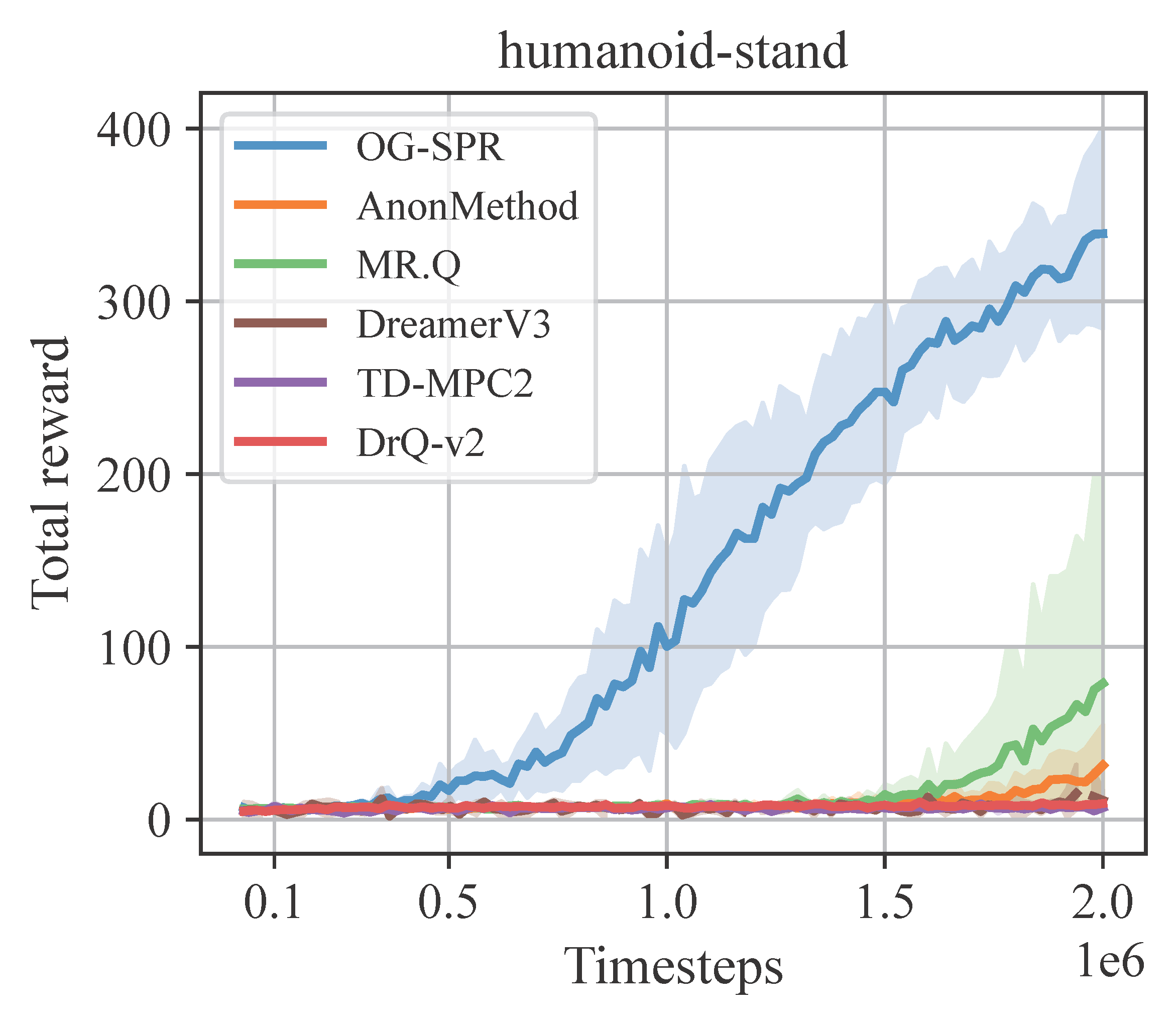}
\caption{
Per-task learning curves on the \textit{humanoid} tasks from the DMControl benchmark for OG-SPR and baselines under the 2M-environment-step training budget. 
Solid lines indicate mean performance over 5 seeds, and shaded areas indicate 95\% bootstrap confidence intervals.
}
\label{per_task3}
\end{figure*}

\begin{table*}[t]
    \setlength{\tabcolsep}{1mm}
    \small
    \centering
    \begin{tabular}{lcccccc}
        \toprule
        \textbf{Task} & \textbf{OG-SPR - OP} & \textbf{OG-SPR - SP} & \textbf{OG-SPR - SVP} &
        \textbf{Shared Adapter} & \textbf{No Adapter} & \textbf{Adapter 2 Only} \\
        \midrule
        acrobot-swingup & 
        455 {{[412, 511]}} & 
        268 {{[221, 277]}} & 
        391 {{[360, 431]}} & 
        412 {{[343, 460]}} & 
        318 {{[250, 365]}} &
        419 {{[414, 426]}} \\
        
        ball\_in\_cup-catch & 
        979 {{[977, 980]}} & 
        977 {{[975, 980]}} & 
        978 {{[975, 981]}} & 
        975 {{[970, 981]}} & 
        922 {{[825, 973]}} &
        973 {{[969, 976]}} \\
        
        cartpole-balance & 
        998 {{[996, 999]}} & 
        998 {{[996, 999]}} & 
        999 {{[997, 999]}} & 
        999 {{[999, 1000]}} & 
        999 {{[999, 1000]}} &
        998 {{[998, 999]}} \\
        
        \makecell[l]{cartpole-\\balance\_sparse} & 
        1000 {{[1000, 1000]}} & 
        1000 {{[1000, 1000]}} & 
        1000 {{[1000, 1000]}} & 
        1000 {{[1000, 1000]}} & 
        999 {{[998, 1000]}} &
        1000 {{[1000, 1000]}} \\
        
        cartpole-swingup & 
        870 {{[866, 878]}} & 
        881 {{[880, 882]}} & 
        877 {{[870, 881]}} & 
        872 {{[862, 882]}} &  
        873 {{[867, 879]}} & 
        880 {{[878, 883]}} \\

        \makecell[l]{cartpole-\\swingup\_sparse} & 
        840 {{[831, 847]}} & 
        841 {{[822, 856]}} & 
        821 {{[789, 845]}} & 
        675 {{[360, 835]}} & 
        814 {{[778, 838]}} & 
        781 {{[671, 844]}} \\
        
        cheetah-run & 
        684 {{[558, 760]}} & 
        854 {{[823, 881]}} & 
        682 {{[575, 739]}} & 
        700 {{[585, 761]}} & 
        726 {{[698, 754]}} & 
        791 {{[756, 843]}} \\
        
        dog-run & 
        71 {{[52, 90]}} & 
        62 {{[50, 80]}} & 
        78 {{[58, 99]}} & 
        84 {{[77, 91]}} &   
        78 {{[57, 100]}} & 
        57 {{[49, 67]}} \\
        
        dog-stand & 
        324 {{[276, 380]}} & 
        240 {{[212, 271]}} & 
        431 {{[329, 534]}} & 
        398 {{[295, 503]}} &  
        401 {{[276, 527]}} & 
        284 {{[232, 356]}} \\
        
        dog-trot & 
        79 {{[70, 88]}} & 
        62 {{[59, 66]}} & 
        83 {{[76, 88]}} & 
        100 {{[78, 123]}} &  
        76 {{[65, 86]}} & 
        69 {{[63, 75]}} \\

        dog-walk & 
        103 {{[100, 106]}} & 
        81 {{[73, 90]}} & 
        114 {{[99, 129]}} & 
        101 {{[88, 114]}} &
        114 {{[102, 131]}} & 
        98 {{[85, 111]}} \\
        
        finger-spin & 
        981 {{[979, 982]}} & 
        984 {{[983, 986]}} & 
        983 {{[977, 986]}} & 
        983 {{[981, 987]}} & 
        982 {{[978, 987]}} & 
        984 {{[979, 988]}} \\
        
        finger-turn\_easy & 
        924 {{[880, 968]}} & 
        912 {{[887, 948]}} & 
        875 {{[749, 969]}} & 
        874 {{[870, 881]}} &  
        945 {{[892, 976]}} & 
        948 {{[900, 974]}} \\
        
        finger-turn\_hard & 
        943 {{[895, 971]}} & 
        920 {{[882, 959]}} & 
        960 {{[952, 969]}} & 
        940 {{[902, 962]}} &  
        747 {{[590, 858]}} & 
        728 {{[507, 916]}} \\
        
        fish-swim & 
        87 {{[53, 120]}} & 
        66 {{[64, 71]}} & 
        101 {{[66, 141]}} & 
        78 {{[56, 115]}} &  
        91 {{[48, 134]}} & 
        64 {{[57, 70]}} \\

        hopper-hop & 
        275 {{[260, 298]}} & 
        287 {{[260, 310]}} & 
        297 {{[272, 321]}} & 
        280 {{[274, 287]}} & 
        301 {{[275, 334]}} & 
        287 {{[263, 316]}} \\
        
        hopper-stand & 
        930 {{[913, 941]}} & 
        903 {{[866, 939]}} & 
        933 {{[928, 939]}} & 
        919 {{[895, 936]}} & 
        912 {{[858, 948]}} & 
        905 {{[864, 929]}} \\
        
        humanoid-run & 
        3 {{[2, 4]}} & 
        1 {{[1, 1]}} & 
        3 {{[2, 4]}} & 
        1 {{[1, 2]}} &  
        2 {{[1, 2]}} & 
        3 {{[2, 4]}} \\
        
        humanoid-stand & 
        9 {{[6, 12]}} & 
        7 {{[5, 8]}} & 
        12 {{[8, 17]}} & 
        11 {{[6, 16]}} &  
        12 {{[8, 16]}} & 
        8 {{[7, 8]}} \\
        
        humanoid-walk & 
        12 {{[7, 17]}} & 
        2 {{[2, 3]}} & 
        7 {{[3, 12]}} & 
        15 {{[7, 23]}} &  
        3 {{[2, 3]}} & 
        5 {{[3, 7]}} \\

        pendulum-swingup & 
        828 {{[785, 889]}} & 
        844 {{[822, 859]}} & 
        836 {{[793, 882]}} & 
        830 {{[812, 852]}} &
        852 {{[837, 868]}} &
        762 {{[593, 852]}} \\
        
        quadruped-run & 
        545 {{[469, 620]}} & 
        481 {{[461, 502]}} & 
        520 {{[478, 602]}} & 
        519 {{[424, 614]}} & 
        414 {{[360, 464]}} & 
        441 {{[410, 473]}} \\
        
        quadruped-walk & 
        679 {{[458, 900]}} & 
        795 {{[752, 849]}} & 
        724 {{[508, 917]}} & 
        833 {{[779, 902]}} &
        561 {{[436, 753]}} & 
        840 {{[737, 918]}} \\
        
        reacher-easy & 
        975 {{[974, 977]}} & 
        978 {{[975, 981]}} & 
        957 {{[908, 985]}} & 
        979 {{[978, 981]}} & 
        978 {{[974, 983]}} & 
        977 {{[976, 978]}} \\
        
        reacher-hard & 
        949 {{[905, 974]}} & 
        939 {{[894, 964]}} & 
        898 {{[869, 947]}} & 
        970 {{[967, 973]}} &  
        972 {{[971, 975]}} & 
        948 {{[900, 974]}} \\

        walker-run & 
        734 {{[721, 746]}} & 
        685 {{[651, 710]}} & 
        646 {{[550, 719]}} & 
        610 {{[562, 656]}} &  
        690 {{[631, 749]}} & 
        674 {{[648, 705]}} \\
        
        walker-stand & 
        983 {{[981, 985]}} & 
        984 {{[979, 990]}} & 
        984 {{[982, 987]}} & 
        979 {{[970, 985]}} &   
        982 {{[981, 984]}} & 
        986 {{[983, 989]}} \\
        
        walker-walk & 
        965 {{[961, 969]}} & 
        972 {{[961, 982]}} & 
        968 {{[966, 972]}} & 
        965 {{[961, 969]}} & 
        957 {{[944, 965]}} & 
        964 {{[957, 971]}} \\

        \bottomrule
    \end{tabular}
    \caption{
    Final raw episode returns of the ablation variants on DMControl after 500k environment steps, corresponding to 1M frames under action repeat of 2. Results are averaged over 5 seeds. The {[bracketed values]} represent a 95\% bootstrap confidence interval. Due to table-width constraints, the full OG-SPR results are not included in this table and are instead reported in Table~\ref{tab:full_dmc}.
    }
    \label{tab:ab_results}
\end{table*}

\begin{table*}[t]
    \setlength{\tabcolsep}{1mm}
    \small
    \centering
    \begin{tabular}{lcccccc}
        \toprule
        \textbf{Task} & \textbf{Human} & \textbf{Random} & \textbf{SPR} &
         \textbf{MR.Q} & \textbf{AnonMethod} & \textbf{OG-SPR} \\
        \midrule
        Alien & 7127.7 & 227.8 &
        903 {{[747, 1060]}} & 
        880 {{[821, 948]}} &
        1158 {{[973, 1366]}} &
        947 [853, 1047] \\
        
        Amidar &  1719.5 & 5.8 &
        203 {{[183, 228]}} & 
        174 {{[154, 196]}} &
        155 {{[121, 190]}} &
        201 [155, 247] \\
        
        Assault &  742.0 & 222.4 &
        699 {{[662, 733]}} & 
        604 {{[587, 625]}} &
        687 {{[659, 714]}} &
        708 [672, 743] \\
        
        Asterix &  8503.3 & 210.0 &
        922 {{[871, 974]}} & 
        1110 {{[1058, 1161]}} &
        1286 {{[1124, 1447]}} &
        1284 [1126, 1509] \\
        
        BankHeist &  753.1 & 14.2 &
        108 {{[32, 242]}} & 
        24 {{[21, 27]}} & 
        41 {{[27, 56]}} &
        55 [39, 71] \\

        BattleZone &  37187.5 & 2360.0 &
        9028 {{[7130, 11218]}} & 
        6539 {{[5122, 7982]}} & 
        4886 {{[2718, 7894]}} & 
        4716 [4346, 5130] \\
        
        Boxing &  12.1 & 0.1 &
        41 {{[30, 53]}} & 
        72 {{[69, 76]}} & 
        76 {{[72, 80]}} &
        60 [47, 73] \\
        
        Breakout &  30.5 & 1.7 &
        14 {{[12, 15]}} & 
        19 {{[17, 21]}} & 
        17 {{[15, 19]}} &
        22 [20, 23] \\
        
        ChopperCommand &  7387.8 & 811.0 &
        1295 {{[1125, 1481]}} & 
        1381 {{[1282, 1494]}} & 
        1741 {{[1613, 1869]}} &
        2041 [1905, 2211] \\
        
        CrazyClimber &  35829.4 & 10780.5 &
        23109 {{[21408, 24950]}} & 
        57113 {{[53805, 60048]}} & 
        69018 {{[61868, 75604]}} &
        68404 [59739, 79846] \\

        DemonAttack &  1971.0 & 152.1 &
        1502 {{[1375, 1638]}} & 
        834 {{[755, 917]}} & 
        1952 {{[1449, 2464]}} &
        1614 [1407, 1851]\\

        Freeway &  29.6 & 0.0 &
        24 {{[18, 29]}} & 
        19 {{[10, 29]}} & 
        16 {{[6, 25]}} &
        6 [1, 9] \\
        
        Frostbite &  4334.7 & 65.2 &
        2938 {{[2732, 3114]}} & 
        2488 {{[1905, 2937]}} & 
        1982 {{[1285, 2601]}} &
        1286 [259, 2452] \\
        
        Gopher &  2412.5 & 257.6 &
        550 {{[448, 649]}} & 
        746 {{[661, 826]}} & 
        810 {{[713, 905]}} & 
        698 [594, 783] \\
        
        Hero &  30826.4 & 1027.0 &
        7956 {{[7587, 8314]}} & 
        8129 {{[7216, 9403]}} & 
        5487 {{[4613, 6360]}} &
        6976 [4846, 8725] \\
        
        Jamesbond &  302.8 & 29.0 &
        405 {{[368, 446]}} &
        411 {{[374, 447]}} & 
        376 {{[334, 410]}} &
        434 [382, 498] \\

        Kangaroo &  3035.0 & 52.0 &
        4282 {{[2849, 5552]}} & 
        1345 {{[659, 2586]}} & 
        422 {{[276, 563]}} &
        2634 [944, 5846] \\
        
        Krull &  2665.5 & 1598.0 &
        3923 {{[3629, 4308]}} & 
        7571 {{[7116, 8056]}} & 
        7570 {{[7226, 7860]}} &
        5467 [5146, 5915] \\
        
        KungFuMaster &  22736.3 & 258.5 &
        15946 {{[13995, 17624]}} & 
        16099 {{[13545, 18704]}} & 
        14670 {{[12813, 16451]}} &
        9729 [9343, 10336]  \\
        
        MsPacman &  6951.6 & 307.3 &
        1529 {{[1402, 1715]}} & 
        1740 {{[1478, 2011]}} & 
        1353 {{[1264, 1444]}} &
        1578 [1236, 1971]  \\
        
        Pong &  14.6 & -20.7 &
        -6 {{[-10, 0]}} & 
        13 {{[9, 17]}} & 
        6 {{[0, 13]}} &
        10 [5, 16]  \\

        PrivateEye &  69571.3 & 24.9 &
        80 {{[59, 100]}} & 
        2 {{[0, 5]}} &
        68 {{[38, 95]}} & 
        94 [83, 100]\\
        
        Qbert &  13455.0 & 163.9 &
        3240 {{[2534, 4029]}} & 
        2397 {{[1455, 3467]}} & 
        766 {{[624, 955]}} &
        2740 [1507, 3974] \\
        
        RoadRunner &  7845.0 & 11.5 &
        13353 {{[9767, 16281]}} & 
        13898 {{[11547, 16591]}} & 
        15285 {{[11914, 18792]}} & 
        11274 [8647, 15859] \\
        
        Seaquest &  42054.7 & 68.4 &
        568 {{[531, 603]}} & 
        591 {{[566, 607]}} & 
        1040 {{[901, 1179]}} &
        736 [541, 895]  \\
        
        UpNDown &  11693.2 & 533.4 &
        7266 {{[5169, 10190]}} & 
        5974 {{[5126, 6778]}} & 
        8508 {{[5620, 13591]}} &
        7545 [5378, 10857]   \\
        \bottomrule
    \end{tabular}
    \caption{
    Final raw game scores on Atari100k after 100k environment steps, corresponding to 400k frames under an action repeat of 4. Results are averaged over 5 seeds. The {{[bracketed values]}} represent a 95\% bootstrap confidence interval.
    }
    \label{main_result_atari}
\end{table*}

{
\small
\bibliography{aaai2027}}